\documentclass[11pt,a4paper]{article}
\usepackage[hyperref]{emnlp2020}
\usepackage{times}
\usepackage{latexsym}

\newcommand{\lm}{LM~}

\usepackage{microtype}
\usepackage{amsmath}
\usepackage{graphics} 
\usepackage{graphicx} 
\usepackage{multirow}
\usepackage{subcaption}
\usepackage{lettrine}

\usepackage{enumitem} 
\setlist[itemize]{noitemsep, topsep=0pt}
\setlist[enumerate]{noitemsep, topsep=0pt}

\usepackage{url}
\aclfinalcopy 
\def\aclpaperid{2563} 

\title{Revisiting Neural Language Modelling with Syllables} 

\author{Arturo Oncevay${}^\dagger{}^\ddagger$ \quad Kervy Rivas Rojas${}^\ddagger$ \\ 
  ${}^\dagger$ School of Informatics, University of Edinburgh, Scotland \\
  ${}^\ddagger$ Artificial Intelligence Research Group, Pontificia Universidad Cat\'olica del Per\'u, Peru \\
  \texttt{a.oncevay@ed.ac.uk,k.rivas@pucp.edu.pe} 
  }

\date{}

\begin{document}
\maketitle
\begin{abstract}
Language modelling is regularly analysed at word, subword or character units, but syllables are seldom used. Syllables provide shorter sequences than characters, they can be extracted with rules, and their segmentation typically requires less specialised effort than identifying morphemes.
We reconsider syllables for an open-vocabulary generation task in 20 languages. 
We use rule-based syllabification methods for five languages and address the rest with a hyphenation tool, which behaviour as syllable proxy is validated. With a comparable perplexity, we show that syllables outperform characters, annotated morphemes and unsupervised subwords. Finally, we also study the overlapping of syllables concerning other subword pieces and discuss some limitations and opportunities. 
\end{abstract}

\setlength{\belowdisplayskip}{2pt} \setlength{\belowdisplayshortskip}{2pt}
\setlength{\abovedisplayskip}{2pt} \setlength{\abovedisplayshortskip}{2pt}

\section{Introduction}
In language modelling (LM), we learn distributions over sequences of words, subwords or characters, where the latter two can allow an open-vocabulary generation~\cite{Sutskever:2011:GTR:3104482.3104610}. We rely on subword segmentation as a widespread approach to generate rare subword units \cite{sennrich-etal-2016-neural}. However, the lack of a representative corpus, in terms of the word vocabulary, constrains the unsupervised segmentation (e.g. with scarce monolingual texts \cite{joshi-etal-2020-state}). 
As an alternative, we could use character-level modelling, since it also has access to subword information \cite{kim-etal-2016-character}, but we face long-term dependency issues and require longer training time to converge. 

In this context, 
we focus on syllables, which are based on speech units: ``A syl-la-ble con-tains a sin-gle vow-el u-nit''. These are more linguistically-based units than characters, and behave as a mapping function to reduce the length of the sequence with a larger ``alphabet'' or syllabary. Their extraction can be rule-based and corpus-independent, but data-driven methods or hyphenation using dictionaries can approximate them as well (see \S\ref{sec:analysis-seg}).  

Previous work on syllable-aware neural \lm failed to beat characters in a closed-vocabulary generation at word-level \cite{assylbekov-etal-2017-syllable};  
however, we propose to assess syllables under three new settings. 
First, we analysed an open-vocabulary scenario with syllables by disregarding additional functions in the input layer (e.g. convolutional filters to hierarchically compose the representations \cite{botha-blunsom-2014-compositional}). Second, we extended the scope from 6 to 20 languages 
to cover different levels of orthographic depth, which is the degree of grapheme-phoneme correspondence \cite{borgwaldt-etal-2005-onset} and a factor that can increase complexity to syllabification. English is a language with deep orthography (weak correspondence) whereas Finnish is transparent \cite{ziegler-etal-2010-orthographic}. Third, we distinguished rule-based syllabification with hyphenation tools, but also validated their proximity for LM. 

Therefore, we revisit \lm for open-vocabulary generation with syllables using pure recurrent neural networks~\cite{merity2018regularizing} for a more diverse set of languages, and compare their performance against characters and other subword units. 

\section{Open-vocabulary language modelling with a comparable perplexity}

\paragraph{Language modelling} Given an input of generic sequence units (such as words, subwords or characters), denoted as $\mathbf{s}=s_1,s_2,...s_n$, a language model computes the probability of $\mathbf{s}$ as:
\begin{equation}
p(\mathbf{s})=\prod_{i=1}^{|\mathbf{s}|} p\left(s_{i} | s_{1}, s_{2}, \ldots, s_{i-1}\right)
\end{equation}

We then can use a recurrent neural network or RNN (e.g. a LSTM variation \cite{merity2018regularizing}), trained at each time step $t$, for calculating the probability of the input $s_{t+1}$, given $s_t$:
\begin{equation}
p\left(s_{t+1}|s_{\leq t}\right)=g\left(\operatorname{RNN}\left(\mathbf{w}_{t}, \mathbf{h}_{t-1}\right)\right)
\end{equation}

where $w_t$ is the learned embedding of the sequence unit $s_t$, $h_{t-1}$ is the hidden state of the RNN for the previous time step, and $g$ is a softmax function for the vocabulary space of the segmentation.

We thereafter compute the loss function $\mathcal{L}_{\mathrm{LM}}(\mathbf{s})$ for the neural \lm as the cross entropy of the model for the sequence $\mathbf{s}$ in a time step t:
\begin{equation}
\mathcal{L}_{\mathrm{LM}}(\mathbf{s})=-\sum_{j=1}^{|V|} b_{t, j} \times \log \left(p({s}_{t, j}\right))
\end{equation}

where $|V|$ is the size of vocabulary, $ p({s}_{t, j})$ is the probability distribution over the vocabulary at each time-step $t$ and $b_{t, j}$ is an indicator if $j$ is the true token for the sequence $\mathbf{s}$.

\paragraph{Character-level perplexity} For a fair comparison across all granularities, we evaluate all results with character-level perplexity: 
\begin{equation}
\operatorname{ppl}^{c} = \exp{(\mathcal{L}_{\mathrm{LM}}(\mathbf{s})\cdot\frac{|\mathbf{s}^{\text{seg}}|+1}{|\mathbf{s}^{c}|+1})}
\end{equation}

where $\mathcal{L}_{\mathrm{LM}}(\mathbf{s})$ is the cross entropy of a string $\mathbf{s}$ computed by the neural \lm, and $|\mathbf{s}^{\text{seg}}|$ and $|\mathbf{s}^{c}|$ refer to the length of $\mathbf{s}$ in the chosen segmentation and character-level units, respectively \cite{miekel2019charppl}. The extra unit considers the end of the sequence.

\paragraph{Open-vocabulary output} We generate the same input unit (e.g. characters or syllables) as an open-vocabulary \lm task, where there is no prediction of an ``unknown'' or out-of-vocabulary word-level token \cite{Sutskever:2011:GTR:3104482.3104610}. 
We thereby differ from previous work \cite{assylbekov-etal-2017-syllable}, and refrain from composing the syllable representations into words to evaluate only word-level perplexity. 

\section{Experimental setup}

\paragraph{Languages and datasets}
Corpora are listed in Table \ref{tab:data}. We do not use the English Penn Treebank~\cite{marcus-etal-1993-building} as in \citet{assylbekov-etal-2017-syllable}, given that it is not suitable for open-vocabulary assessment. We then chose \mbox{WikiText-2-raw~\cite[en\textsubscript{wt2};][]{merity2016pointer}}, which contains around two million word-level tokens extracted from Wikipedia articles in English. Furthermore, 
we employ 20 Universal Dependencies \cite[UD;][]{nivre-etal-2020-universal} treebanks, similarly to \citet{blevins-zettlemoyer-2019-better}.\footnote{The languages are chosen given the availability of an open-source syllabification or hyphenation tool. We prefer to use the UD treebanks, instead of other well-known datasets for language modelling (e.g. Multilingual Wikipedia Corpus \cite{kawakami-etal-2017-learning}), because they provide morphological annotation used for the study.} 


\begin{table}
    \centering
    \setlength\tabcolsep{3pt}
    \resizebox{\linewidth}{!}{%
    \begin{tabular}{cl|rrr|rrr}
   &        &  \multicolumn{3}{c|}{\# tokens} & \multicolumn{3}{c}{\# types} \\ \hline
L. & Corpus & Word & Syl & Char & Word & Syl & Char \\ \hline 
bg & UD-BTB & 125k & 386k & 710k & 25,067 & 6,035 & 132 \\ 
ca & UD-AnCora & 436k & 1,123k & 2,341k & 31,226 & 12,211 & 108 \\ 
cs & UD-PDT & 1,158k & 3,546k & 6,868k & 127,211 & 26,451 & 136 \\ 
da & UD-DDT & 81k & 215k & 442k & 16,300 & 7,362 & 96 \\ 
de & UD-GSD & 260k & 735k & 1,637k & 49,561 & 16,549 & 238 \\ 
en* & UD-EWT & 210k & 488k & 1,061k & 20,291 & 13,962 & 100 \\ 
en* & wikitext-2-raw & 2,089k & 4,894k & 10,902k & 33,279 & 20,173 & 398 \\ 
es* & UD-GSD & 376k & 1,060k & 2,043k & 46,623 & 13,676 & 292 \\ 
fi* & UD-TDT & 165k & 595k & 1,224k & 49,045 & 8,168 & 234 \\ 
fr & UD-GSD & 360k & 837k & 1,959k & 42,520 & 24,418 & 293 \\ 
it & UD-SET & 263k & 762k & 1,504k & 28,685 & 7,990 & 121 \\ 
hr & UD-ISDT & 154k & 484k & 930k & 33,365 & 7,374 & 106 \\ 
pt & UD-LVTB & 192k & 551k & 1,040k & 27,130 & 8,267 & 111 \\ 
lv & UD-Alpino & 113k & 349k & 690k & 28,677 & 7,014 & 116 \\ 
nl & UD-LFG & 187k & 488k & 1,074k & 26,704 & 10,553 & 96 \\ 
pl & UD-Bosque & 102k & 293k & 589k & 28,864 & 7,978 & 104 \\ 
ro & UD-RRT & 183k & 549k & 1,056k & 31,402 & 8,763 & 139 \\ 
ru* & UD-SynTagRus & 867k & 2,707k & 5,411k & 108,833 & 16,961 & 153 \\ 
sk & UD-SNK & 80k & 232k & 437k & 21,137 & 6,444 & 122 \\ 
tr* & UD-IMST & 38k & 126k & 242k & 13,966 & 2,595 & 84 \\ 
uk & UD-IU & 88k & 289k & 501k & 26,567 & 5,505 & 174 \\    \hline
    \end{tabular}
    }
    \caption{Total number of tokens and token types in the training set for words, syllables and characters (other segmentation data is in Appendix \ref{app:datasets}). We highlight (*) the languages with extracted rule-based syllables, and we report the statistics with hyphenation for the rest.}
    \label{tab:data}
\end{table}

\paragraph{Syllable segmentation} For splitting syllables in different languages, we used rule-based syllabification tools for English, Spanish, Russian, Finnish and Turkish, and a dictionary-based hyphenation tool for all of them except for Finnish and Turkish. All the tools are listed in Appendix \ref{app:segmentation}. 

\paragraph{Segmentation baselines} Besides the annotated morphemes in the UD treebanks, we consider Polyglot (\url{polyglot-nlp.com}), which includes models for unsupervised morpheme segmentation trained with Morfessor \cite{virpioja-2013-morfessor}. Moreover, we employ an unsupervised subword segmentation baseline of Byte Pair Encoding~\cite[BPE;][]{sennrich-etal-2016-neural}\footnote{We use: \url{https://github.com/huggingface/tokenizers}} with different vocabulary sizes from 2,500 to 10,000 tokens, with 2,500 steps. We also fix the parameter to the syllabary size%
. Appendix \ref{app:segmentation} includes details about the segmentation format.


\paragraph{Model and training} Following other open-vocabulary \lm studies~\cite{mielke-eisner-2019-spell, mielke-etal-2019-kind}, we use a low-compute version of an LSTM neural network, named Average SGD Weight-Dropped~\cite{merity2018regularizing}, developed in PyTorch\footnote{\url{https://github.com/salesforce/awd-lstm-lm}}. Appendix \ref{app:model} includes more details.  

\section{Segmentation analysis}
\label{sec:analysis-seg}

\paragraph{Syllabification versus Hyphenation} For English, Spanish and Russian, we confirmed that hyphenation is a reasonable proxy for syllabification in LM. Table \ref{tab:syl-vs-hyphen} shows that the overlapping of hyphenation pieces concerning rule-based syllables is more than 70\%. 
Furthermore, we note that the perplexity gap is not significant. The small sample is still representative because English has a deep orthography in contrast with Spanish or Russian, and Russian use a different script (Cyrillic instead of Latin). In terms of morphology, all of them are fusional, but highly agglutinative languages like Turkish or Finnish are not feasible candidates for developing a Hunspell-like dictionary, which is the source of most hyphenation tools.

\begin{table}[h!]
    \centering

\resizebox{0.8\linewidth}{!}{%
\begin{tabular}{l|cc|r}
                      & $(H \cap S)/S$ & $(S \cap H)/H$ & $\Delta\operatorname{ppl}^c$\textsubscript{S-H} \\ \hline
en\textsubscript{wt2} & 70.07\%        & 73.84\%            & -0.0001  \\ 
en\textsubscript{UD}  & 78.13\%        & 74.68\%        & -0.0186  \\
es                    & 76.37\%        & 61.95\%        & 0.0373   \\ 
ru                    & 73.58\%        & 56.13\%        & 0.0901   \\ \hline                                         
\end{tabular}
}
\caption{Token types overlapping of syllabification (S) versus hyphenation (H) and their perplexity gap.}
\label{tab:syl-vs-hyphen}
\end{table}

\begin{figure}[h!]
    \centering
\includegraphics[width=\linewidth,clip]{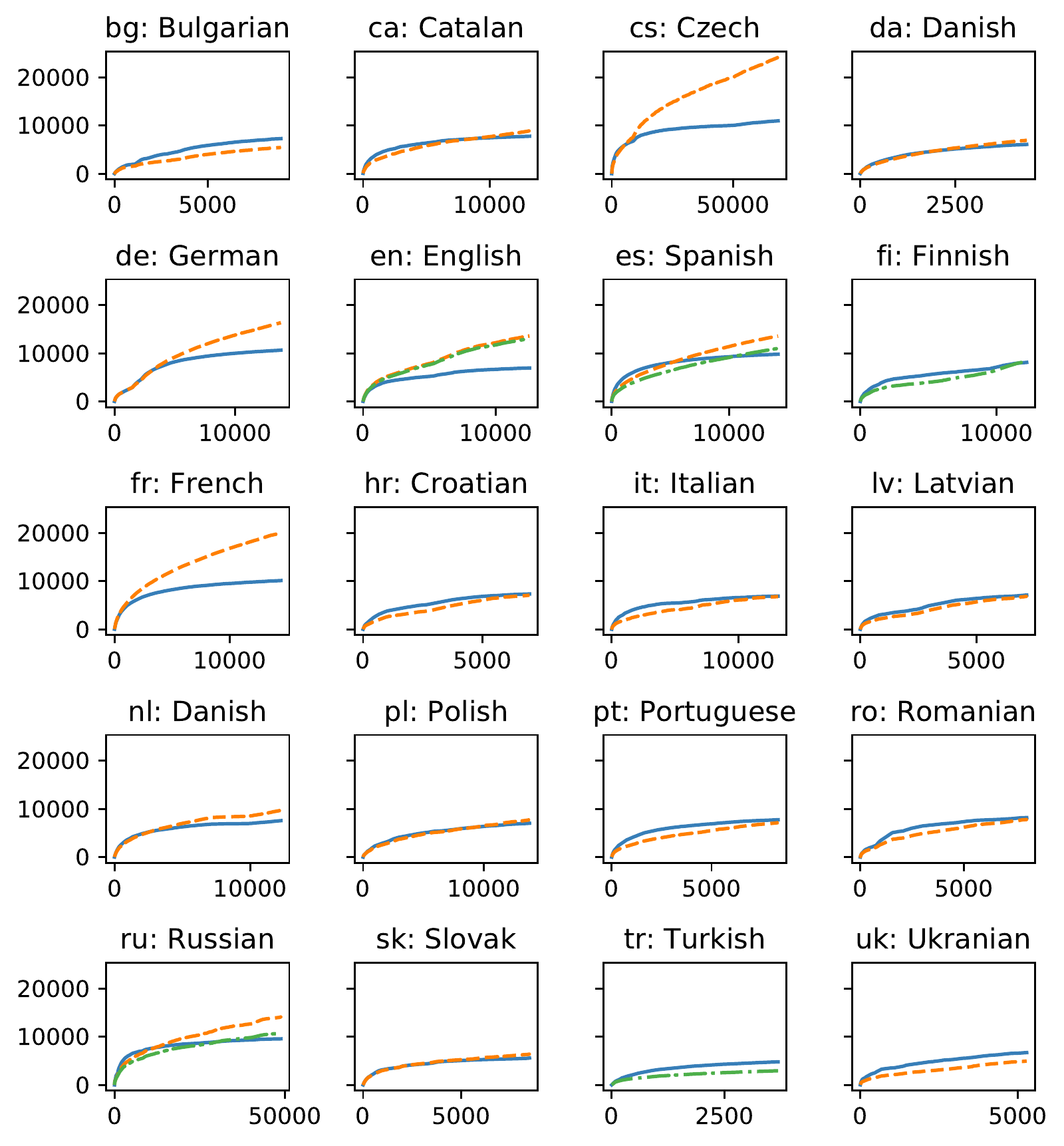}
\includegraphics[width=0.8\linewidth,clip]{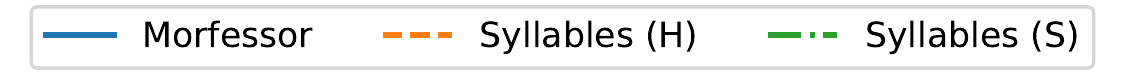}
\caption{Vocabulary growth: accumulative number of token types (y-axis) per samples in the training data (x-axis) for all the UD datasets. Y-axis range is fixed.}
\label{fig:vocab-size}
\end{figure}

\paragraph{Vocabulary growth} In Figure \ref{fig:vocab-size}, we show the vocabulary growth rate of unsupervised morphemes (with Morfessor)\footnote{We do not show the growth of the annotated morpheme pieces, because they approximate the word-level vocabulary given the presence of lemmas.} in contrast with syllables extracted by rules (S) or hyphenation (H). We do not observe a significant difference between syllabification and hyphenation, which reinforce their functional proximity for our study. We also observe that in all the UD datasets we do not have more than 10k Morfessor-based pieces, which is the reason for the upper boundary on our BPE baselines. However, for Czech, German, English and French, we observe that the syllabary size significantly surpass the number of Morfessor-based pieces. Potential explanations are the orthographic depth and the degree of syllabic complexity for those languages \cite{borleffs-etal-2017-measuring}: a low letter-sound correspondence and the difficulty to determine the syllable boundaries can induce a larger syllabary.

\begin{figure*}[t!]
\begin{center}
\centering

\begin{subfigure}[t]{0.31\linewidth}
\caption{Overlapping w.r.t. Morph.}
\label{fig:over-morph}
\includegraphics[width=\linewidth,clip]{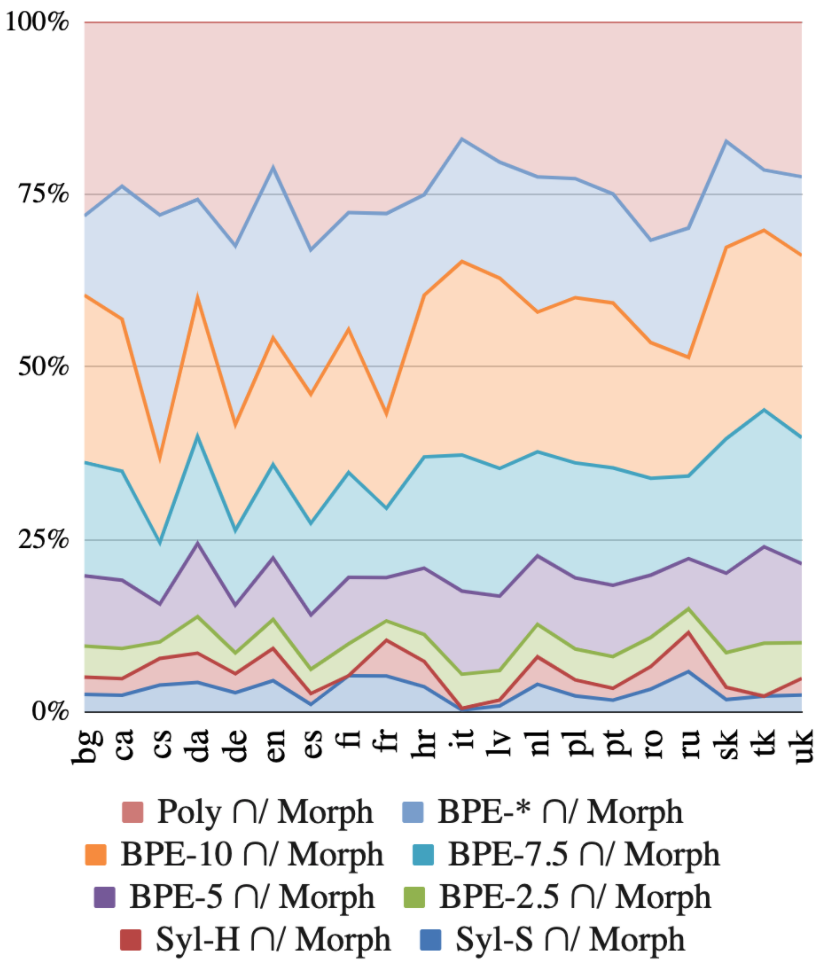}
\end{subfigure}
\begin{subfigure}[t]{0.31\linewidth}
\caption{Overlapping w.r.t. Poly.}
\label{fig:over-poly}
\includegraphics[width=\linewidth,clip]{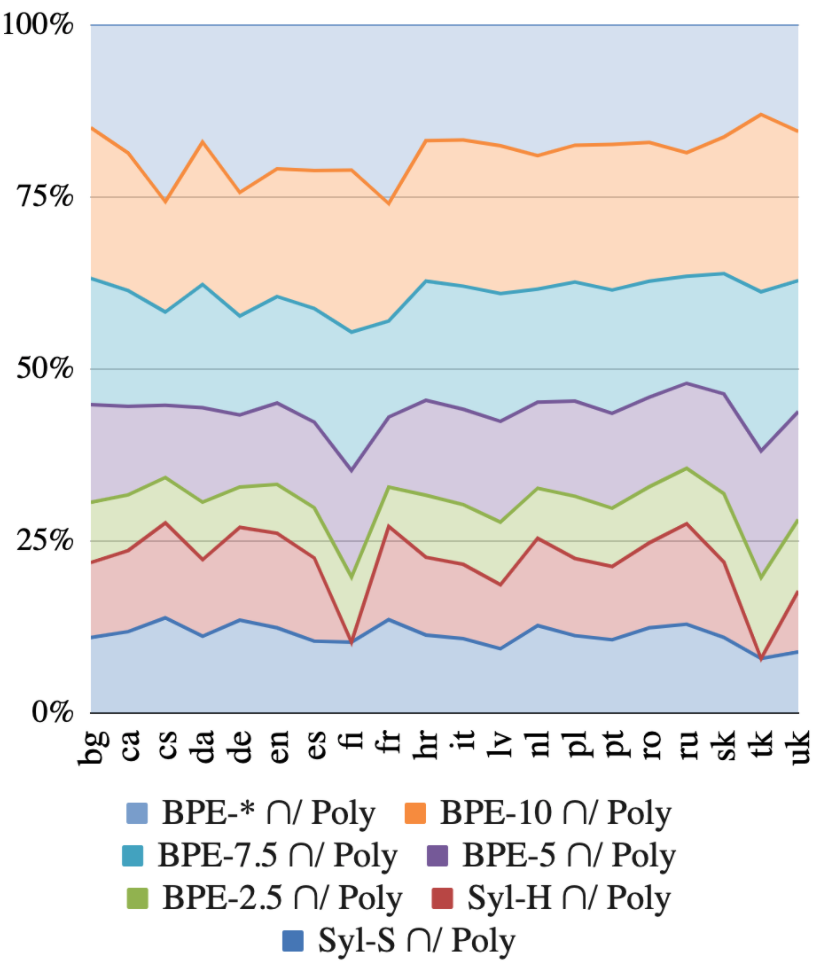}
\end{subfigure}
\begin{subfigure}[t]{0.32\linewidth}
\caption{Overlapping w.r.t. BPE}
\label{fig:over-bpe}
\includegraphics[width=\linewidth,clip]{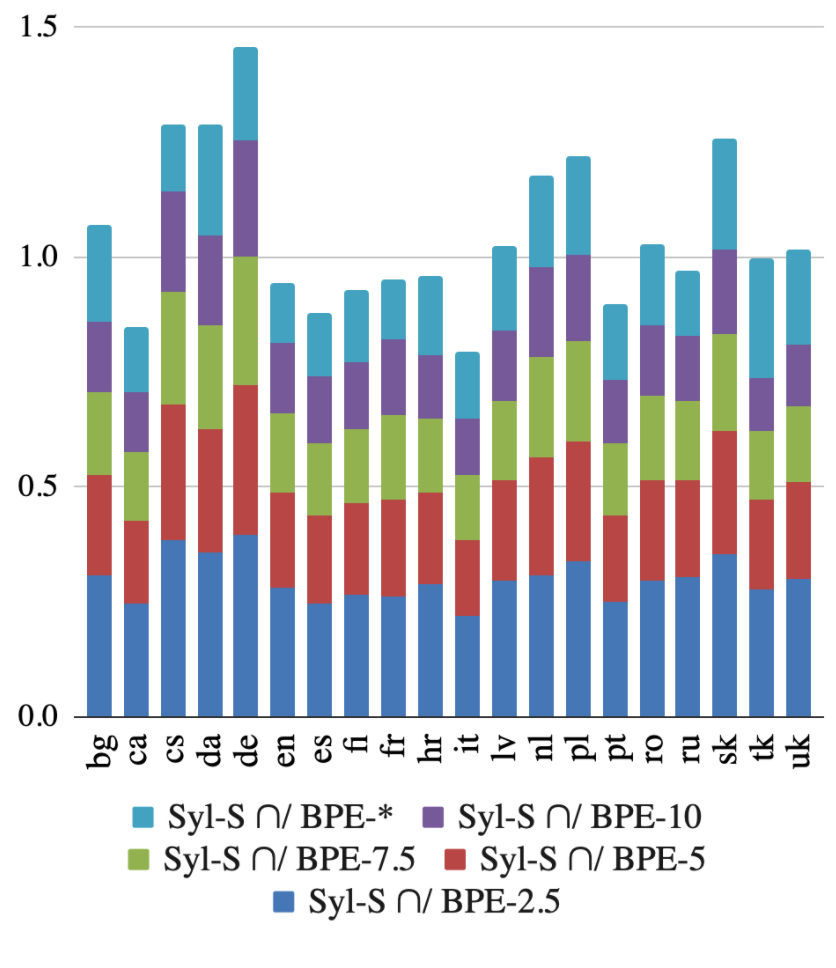}
\end{subfigure}

\caption{Overlapping analysis of syllables (Syl-S: Rules, Syl-H: Hyphenation) with respect to annotated morphemes (Morph.), unsupervised morphemes (with Morfessor on Poly.) and BPE with different vocabulary sizes in thousands (* indicates the same vocabulary size as the syllabary). The operation $A \cap / B$ is equivalent to $(A \cap B)/B$, where $A$ and $B$ are number of token types (unique subword pieces). Overlapping ratios in plots (a-b) are normalised from 0 to 100\%, whereas in (c) is only stacked.}
\label{fig:convergence}
\end{center}
\end{figure*}

\paragraph{Overlapping of syllables with subwords} In Figure \ref{fig:over-morph}, we show the stacked area plot of the overlapping ratio of syllables, Morfessor-based pieces and different BPE settings concerning the annotated morphemes. We note that the proportion of syllable overlapping is relatively low in contrast with unsupervised morphemes or BPE with larger vocabulary size. The outcome is expected, as the annotated morpheme contains lemmas or full words with long sequences of characters, which are adopted by BPE with more merge-operations. 

Analogously, Figure \ref{fig:over-poly} shows the overlapping of syllables and BPE concerning unsupervised morphemes. We observe that the intersection ratio of syllables has increased, and approximate the values of BPE with larger vocabulary size. 

In both scenarios, it is worthy to note that the BPE setting with the largest overlapping with annotated or unsupervised morphemes has a vocabulary size that equates the syllabary. Future work can assess whether the number of unique syllables can support the tuning of the BPE vocabulary size.

Finally, in Figure \ref{fig:over-bpe}, we observe that syllables overlap the most with BPE pieces that uses a small vocabulary size of 2,500. From 5,000 to 10,000, the overlapping ratio shows a downtrend for most of the datasets. However, in 9 out of 20 datasets, the largest BPE setting (fixed with the syllabary size) shows an increase of the overlapping.

\section{Open-vocabulary LM results}

Table \ref{tab:perplexity} shows the $\operatorname{ppl}^c$ values for the different levels of segmentation we considered in the study, where we did not tune the neural \lm for an specific setting. We observe that syllables always result in better perplexities that other granularities, even for deep orthography languages such as English or French. 
The results obtained by the BPE baselines are relatively poor as well, and they could not beat characters in any dataset. 
We could search for an optimal parameter for the BPE algorithm; however, the advantage of the syllables is that we do not need to tune an hyper-parameter to extract a different set of subword pieces.


As a significant outcome, we note that syllables did not fail to beat characters, at least in an open-vocabulary \lm task, which extends the results provided by \citet{assylbekov-etal-2017-syllable}. Moreover, in Figure \ref{fig:ppl-diff} in the Appendix, we can observe a strong linear relationship of the syllable type/token ratio with the gain of $\operatorname{ppl}^c$ obtained by syllables concerning characters. In other words, if our dataset possesses a rich syllabary, we are fairly approximating the amount of word-level tokens, which reduces the $\operatorname{ppl}^c$ gain. 

Beating characters implies a gain in time processing as well, given the shorter sequences of the syllable pieces. Figure \ref{fig:convergence} shows details about how many epochs each segmentation requires to converge during training, as well as how much time requires all the training. Syllable-level \lm trains faster than characters, and even when they are slower (seconds per epoch) than using BPE-pieces or unsupervised morphemes, they obtain a better $\operatorname{ppl}^c$ score. 

\begin{table}
    \centering
    \setlength\tabcolsep{3pt}
    \resizebox{\linewidth}{!}{%
    \begin{tabular}{l|ccccc}
 & Char. & Morph. & Poly. & Syl. & BPE\textsubscript{best} \\ \hline
bg & 3.56 \small{$\pm$0.03} & 4.09 \small{$\pm$0.05} & 4.69 \small{$\pm$0.01} & \textbf{2.87} \small{$\pm$0.0} & 5.19 \small{$\pm$0.01} \\
ca & 2.84 \small{$\pm$0.0} & 3.11 \small{$\pm$0.02} & 3.26 \small{$\pm$0.01} & \textbf{2.21} \small{$\pm$0.0} & 3.31 \small{$\pm$0.0} \\
cs & 3.32 \small{$\pm$0.0} & 3.11 \small{$\pm$0.01} & 4.18 \small{$\pm$0.01} & \textbf{2.66} \small{$\pm$0.0} & 4.24 \small{$\pm$0.0} \\
da & 4.25 \small{$\pm$0.01} & 4.42 \small{$\pm$0.04} & 5.6 \small{$\pm$0.0} & \textbf{3.1} \small{$\pm$0.01} & 6.21 \small{$\pm$0.03} \\
de & 3.5 \small{$\pm$0.04} & 3.36 \small{$\pm$0.08} & 3.79 \small{$\pm$0.0} & \textbf{2.48} \small{$\pm$0.0} & 3.86 \small{$\pm$0.02} \\
en\textsubscript{UD}* & 4.11 \small{$\pm$0.01} & 4.39 \small{$\pm$0.08} & 5.67 \small{$\pm$0.01} & \textbf{2.82} \small{$\pm$0.07} & 5.65 \small{$\pm$0.04} \\
en\textsubscript{wt2}* & 2.48 \small{$\pm$0.0} & - & 2.8 \small{$\pm$0.0} & \textbf{1.96} \small{$\pm$0.0} & 2.91 \small{$\pm$0.0} \\
es* & 3.16 \small{$\pm$0.01} & 3.71 \small{$\pm$0.04} & 3.95 \small{$\pm$0.01} & \textbf{2.51} \small{$\pm$0.0} & 3.98 \small{$\pm$0.0} \\
fi* & 3.77 \small{$\pm$0.01} & 4.05 \small{$\pm$0.12} & 4.76 \small{$\pm$0.01} & \textbf{3.1} \small{$\pm$0.0} & 5.27 \small{$\pm$0.01} \\
fr & 3.09 \small{$\pm$0.01} & 3.67 \small{$\pm$0.02} & 3.82 \small{$\pm$0.01} & \textbf{2.3} \small{$\pm$0.01} & 3.87 \small{$\pm$0.01} \\
hr & 3.52 \small{$\pm$0.02} & 3.92 \small{$\pm$0.01} & 4.34 \small{$\pm$0.0} & \textbf{2.8} \small{$\pm$0.0} & 4.52 \small{$\pm$0.02} \\
it & 2.8 \small{$\pm$0.0} & 3.19 \small{$\pm$0.0} & 3.43 \small{$\pm$0.01} & \textbf{2.27} \small{$\pm$0.01} & 3.61 \small{$\pm$0.0} \\
lv & 4.55 \small{$\pm$0.02} & 5.31 \small{$\pm$0.0} & 6.82 \small{$\pm$0.02} & \textbf{3.59} \small{$\pm$0.0} & 7.19 \small{$\pm$0.0} \\
nl & 3.83 \small{$\pm$0.05} & 3.69 \small{$\pm$0.1} & 4.44 \small{$\pm$0.01} & \textbf{2.76} \small{$\pm$0.01} & 4.83 \small{$\pm$0.01} \\
pl & 4.03 \small{$\pm$0.01} & 4.77 \small{$\pm$0.22} & 5.96 \small{$\pm$0.04} & \textbf{3.19} \small{$\pm$0.0} & 5.99 \small{$\pm$0.0} \\
pt & 3.31 \small{$\pm$0.01} & 3.46 \small{$\pm$0.03} & 4.03 \small{$\pm$0.01} & \textbf{2.56} \small{$\pm$0.0} & 4.24 \small{$\pm$0.01} \\
ro & 3.4 \small{$\pm$0.02} & 3.89 \small{$\pm$0.04} & 4.25 \small{$\pm$0.01} & \textbf{2.72} \small{$\pm$0.0} & 4.71 \small{$\pm$0.01} \\
ru* & 3.28 \small{$\pm$0.0} & 2.93 \small{$\pm$0.01} & 4.05 \small{$\pm$0.0} & \textbf{2.69} \small{$\pm$0.01} & 4.04 \small{$\pm$0.0} \\
sk & 6.16 \small{$\pm$0.05} & 5.1 \small{$\pm$0.07} & 7.61 \small{$\pm$0.08} & \textbf{4.62} \small{$\pm$0.01} & 10.51 \small{$\pm$0.03} \\
tr* & 4.16 \small{$\pm$0.05} & 4.86 \small{$\pm$0.05} & 6.41 \small{$\pm$0.07} & \textbf{3.66} \small{$\pm$0.03} & 6.98 \small{$\pm$0.1} \\
uk & 4.92 \small{$\pm$0.02} & 6.45 \small{$\pm$0.11} & 8.11 \small{$\pm$0.03} & \textbf{4.24} \small{$\pm$0.02} & 9.23 \small{$\pm$0.02} \\  \hline

    \end{tabular}
    }
    \caption{Character-level perplexity ($\downarrow$) in test. We show the mean and standard deviation for three runs with different seeds. ``Syl.'' presents the syllabification-based result if it is available (*), or the hyphenation otherwise.}
    \label{tab:perplexity}
\end{table}

\begin{figure}[]
\begin{center}
\centering

\begin{subfigure}[t]{0.49\linewidth}
\label{fig:wiki-ppl}
\includegraphics[width=\linewidth,clip]{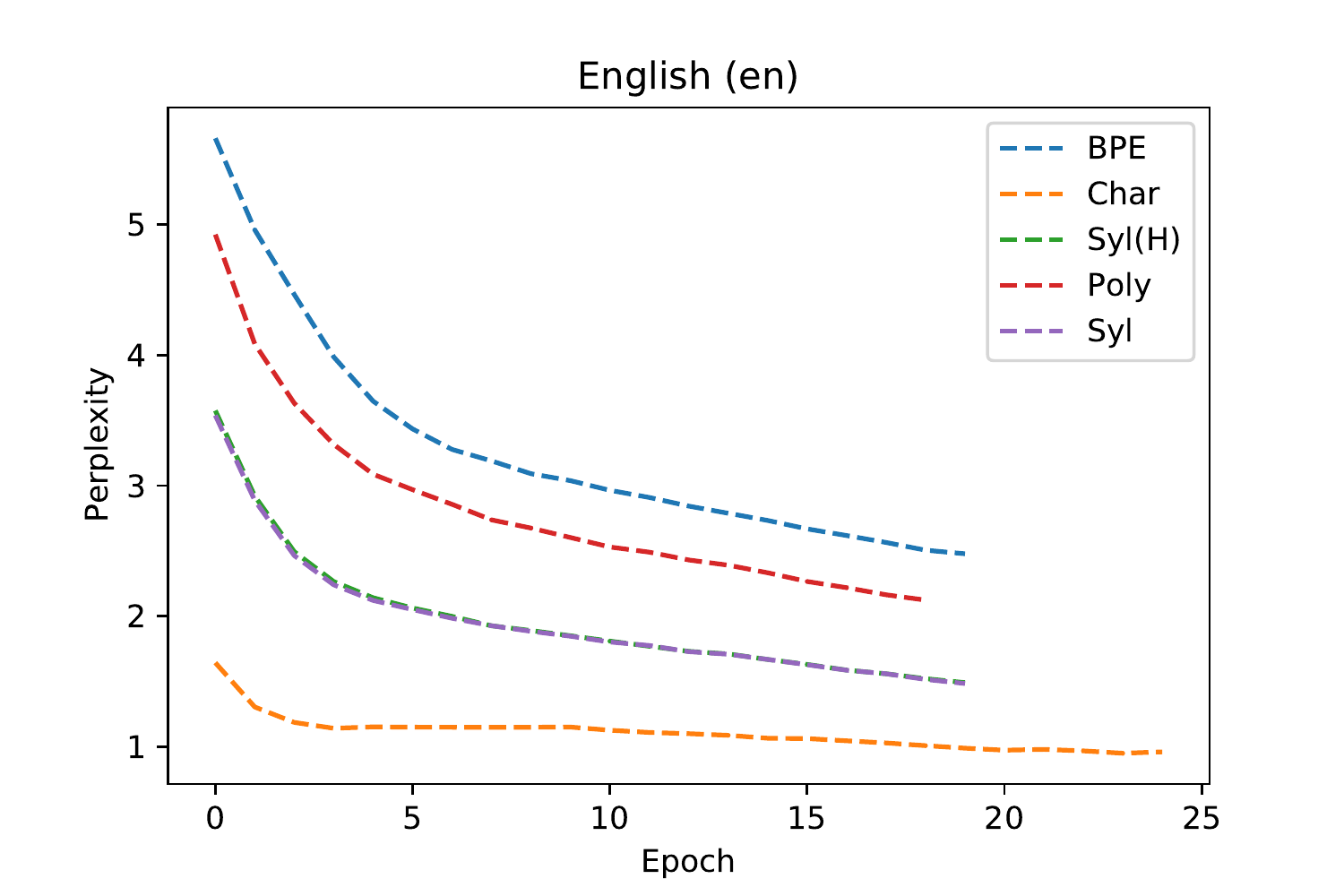}
\end{subfigure}
\begin{subfigure}[t]{0.49\linewidth}
\label{fig:wiki-time}
\includegraphics[width=\linewidth,clip]{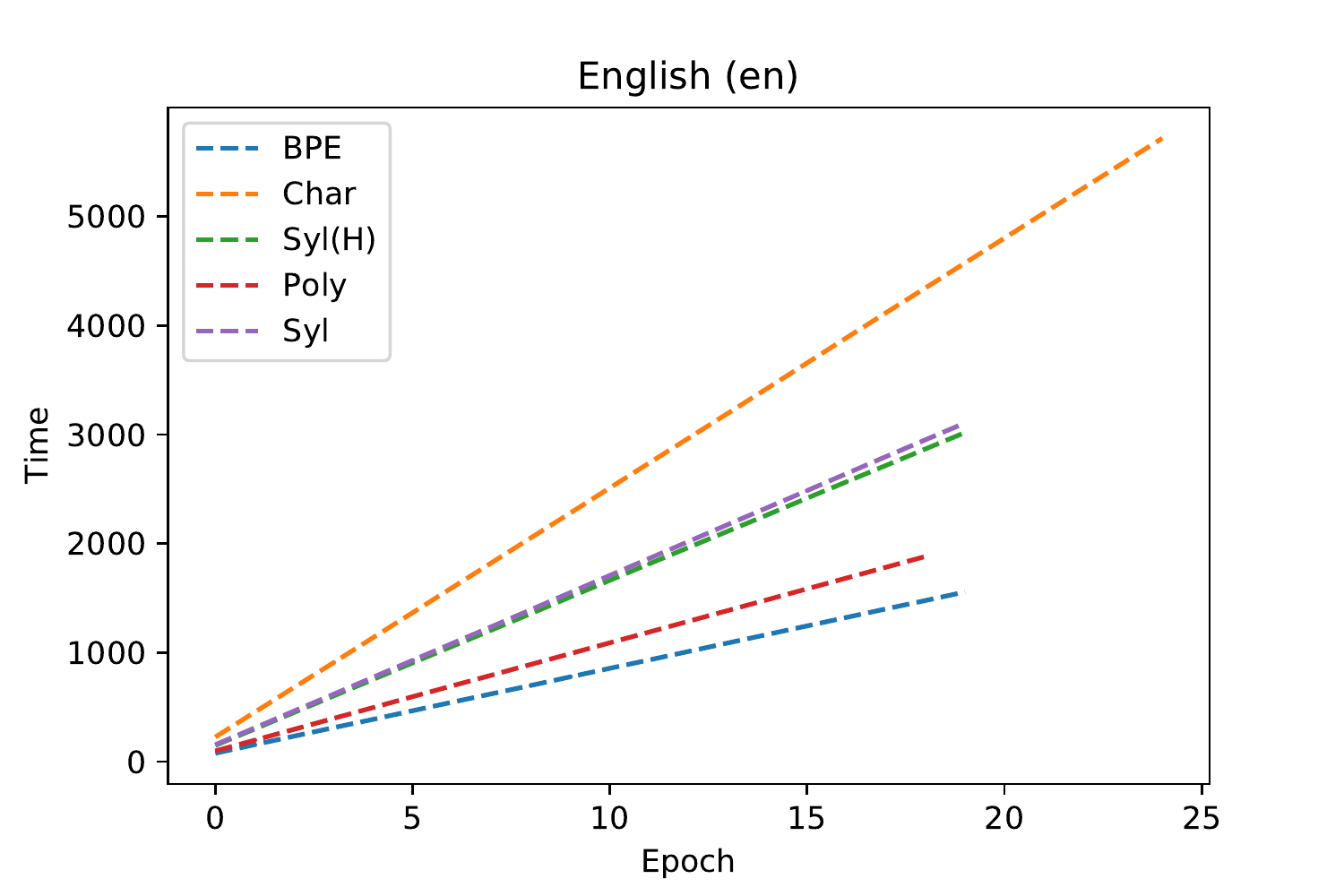}
\end{subfigure}

\caption{Left: validation perplexity for the English wikitext-2 per epoch. Right: training time (in seconds) per epoch for the same dataset. Similar plots for other datasets are included in the Appendix.} 
\label{fig:convergence}
\end{center}
\end{figure}

\section{Related work}

In subword-aware LM, \citet{vania-lopez-2017-characters} investigated if we can capture morphology using characters, character n-grams, BPE pieces or morphemes with a closed-vocabulary. Whereas for open-vocabulary generation, \citet{blevins-zettlemoyer-2019-better} incorporates morphological supervision with a multi-task objective, and \citet{kawakami-etal-2017-learning,mielke-eisner-2019-spell} have focused in improving the neural architecture to jointly use the representation of characters and words in a hybrid open-vocabulary setting.

The closest study to ours is from \citet{mikolov-etal-2012-subword}, where they performed subword-grained prediction with different settings, and used syllables as a proxy to split words with low frequency, reduce the vocabulary and compress the model size. However, they only focused on English. Besides, syllable-aware \lm was addressed by \citet{assylbekov-etal-2017-syllable} for English, German, French, Czech, Spanish and Russian, and by \citet{yu-etal-2017-syllable} for Korean. However, in both cases, the syllable units have been composed with convolutional filters into word-level representations to assess a closed-vocabulary setting. As far as we know, we propose the first syllable-level open-vocabulary \lm study that analyses up to 20 languages and compare their results against morphemes.

\section{Limitations and opportunities} 

Syllables only cannot offer a universal solution to the subword segmentation problem for all the languages, as the syllabification tools are language-dependent. Besides, the analysis should be extended to different scripts and morphological typology\footnote{Only Bulgarian, Russian and Ukranian use the Cyrillic script instead of Latin; and only Finnish and Turkish have highly agglutinative morphology instead of fusional. Polysynthetic languages can also be potential beneficiaries with syllable-level language modelling}. Furthermore, we do not encode any semantics in the syllable-vector space, with a few exceptions like in Korean \cite{choi-etal-2017-syllable}. 

Nevertheless, our results confirm that syllables are reliable for LM, and building a syllable splitter might require less effort than annotating morphemes to train a robust supervised tool.\footnote{For instance, the syllabification tool that we used for English is based on five general rules published in: \url{https://www.howmanysyllables.com/divideintosyllables}. Their implementation should take less effort than annotating a UD treebank or building a Finite-State-Transducer for morphological analysis.} 
Moreover, as we discussed at the end of \S\ref{sec:analysis-seg}, we could further assess whether the syllables can support the hyper-parameter tuning or impact the internal procedure of unsupervised segmentation methods like BPE or Morfessor, which are corpus-dependent. 


\section{Conclusion}
We proved that syllables are valuable for an open-vocabulary \lm task, where they behave positively even for languages with deep orthography, and they overcome character-level models that require longer time to train. Syllables do not have an embedded meaning in most of the languages; however, the required effort for their segmentation could be advantageous against morphological-aware or unsupervised-driven methods. Given our analysis, we could consider working on syllable-driven subword segmentation, neural machine translation and hybrid-\lm (with characters and words).

\section*{Acknowledgments}
\lettrine[image=true, lines=2, findent=1ex, nindent=0ex, loversize=.15]{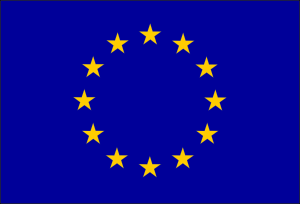}%
{T}he first author is supported by funding from the European Union's Horizon 2020 research and innovation programme under grant agreements No 825299 (GoURMET) and the EPSRC fellowship grant EP/S001271/1 (MTStretch), whereas the second author is supported by a research grant of the ``Consejo Nacional de Ciencia, Tecnolog\'ia e Innovaci\'on Tecnol\'ogica'' (CONCYTEC, Peru) under the contract 183-2018-FONDECYT-BM-IADT-MU.

We express our thanks to Shay B. Cohen, Clara Vania and Adam L\'opez, whose provided us feedback in a very early stage of the study, and to Alexandra Birch, Barry Haddow, Fernando Alva-Manchego, John E. Miller and Marco Sobrevilla-Cabezudo, whose reviewed different draft versions. Finally, we acknowledge the support of NVIDIA Corporation with the donation of a Titan Xp GPU used for the study.

\bibliography{emnlp2020}

\begin{thebibliography}{25}
\expandafter\ifx\csname natexlab\endcsname\relax\def\natexlab#1{#1}\fi

\bibitem[{Assylbekov et~al.(2017)Assylbekov, Takhanov, Myrzakhmetov, and
  Washington}]{assylbekov-etal-2017-syllable}
Zhenisbek Assylbekov, Rustem Takhanov, Bagdat Myrzakhmetov, and Jonathan~N.
  Washington. 2017.
\newblock \href {https://doi.org/10.18653/v1/D17-1199} {Syllable-aware neural
  language models: A failure to beat character-aware ones}.
\newblock In \emph{Proceedings of the 2017 Conference on Empirical Methods in
  Natural Language Processing}, pages 1866--1872, Copenhagen, Denmark.
  Association for Computational Linguistics.

\bibitem[{Blevins and Zettlemoyer(2019)}]{blevins-zettlemoyer-2019-better}
Terra Blevins and Luke Zettlemoyer. 2019.
\newblock \href {https://www.aclweb.org/anthology/P19-1156} {Better character
  language modeling through morphology}.
\newblock In \emph{Proceedings of the 57th Annual Meeting of the Association
  for Computational Linguistics}, pages 1606--1613, Florence, Italy.
  Association for Computational Linguistics.

\bibitem[{Borgwaldt et~al.(2005)Borgwaldt, Hellwig, and
  De~Groot}]{borgwaldt-etal-2005-onset}
Susanne~R Borgwaldt, Frauke~M Hellwig, and Annette~MB De~Groot. 2005.
\newblock Onset entropy matters--letter-to-phoneme mappings in seven languages.
\newblock \emph{Reading and Writing}, 18(3):211--229.

\bibitem[{Borleffs et~al.(2017)Borleffs, Maassen, Lyytinen, and
  Zwarts}]{borleffs-etal-2017-measuring}
Elisabeth Borleffs, Ben~AM Maassen, Heikki Lyytinen, and Frans Zwarts. 2017.
\newblock Measuring orthographic transparency and morphological-syllabic
  complexity in alphabetic orthographies: a narrative review.
\newblock \emph{Reading and writing}, 30(8):1617--1638.

\bibitem[{Botha and Blunsom(2014)}]{botha-blunsom-2014-compositional}
Jan~A. Botha and Phil Blunsom. 2014.
\newblock \href {http://dl.acm.org/citation.cfm?id=3044805.3045104}
  {Compositional morphology for word representations and language modelling}.
\newblock In \emph{Proceedings of the 31st International Conference on
  International Conference on Machine Learning - Volume 32}, ICML'14, pages
  II--1899--II--1907. JMLR.org.

\bibitem[{Choi et~al.(2017)Choi, Kim, Seol, and Lee}]{choi-etal-2017-syllable}
Sanghyuk Choi, Taeuk Kim, Jinseok Seol, and Sang-goo Lee. 2017.
\newblock \href {https://doi.org/10.18653/v1/W17-4105} {A syllable-based
  technique for word embeddings of {K}orean words}.
\newblock In \emph{Proceedings of the First Workshop on Subword and Character
  Level Models in {NLP}}, pages 36--40, Copenhagen, Denmark. Association for
  Computational Linguistics.

\bibitem[{Joshi et~al.(2020)Joshi, Santy, Budhiraja, Bali, and
  Choudhury}]{joshi-etal-2020-state}
Pratik Joshi, Sebastin Santy, Amar Budhiraja, Kalika Bali, and Monojit
  Choudhury. 2020.
\newblock \href {https://doi.org/10.18653/v1/2020.acl-main.560} {The state and
  fate of linguistic diversity and inclusion in the {NLP} world}.
\newblock In \emph{Proceedings of the 58th Annual Meeting of the Association
  for Computational Linguistics}, pages 6282--6293, Online. Association for
  Computational Linguistics.

\bibitem[{Kawakami et~al.(2017)Kawakami, Dyer, and
  Blunsom}]{kawakami-etal-2017-learning}
Kazuya Kawakami, Chris Dyer, and Phil Blunsom. 2017.
\newblock \href {https://doi.org/10.18653/v1/P17-1137} {Learning to create and
  reuse words in open-vocabulary neural language modeling}.
\newblock In \emph{Proceedings of the 55th Annual Meeting of the Association
  for Computational Linguistics (Volume 1: Long Papers)}, pages 1492--1502,
  Vancouver, Canada. Association for Computational Linguistics.

\bibitem[{Kim et~al.(2016)Kim, Jernite, Sontag, and
  Rush}]{kim-etal-2016-character}
Yoon Kim, Yacine Jernite, David Sontag, and Alexander~M. Rush. 2016.
\newblock \href {http://dl.acm.org/citation.cfm?id=3016100.3016285}
  {Character-aware neural language models}.
\newblock In \emph{Proceedings of the Thirtieth AAAI Conference on Artificial
  Intelligence}, AAAI'16, pages 2741--2749. AAAI Press.

\bibitem[{Kingma and Ba(2015)}]{kingma-ba-2015-adam}
Diederik~P. Kingma and Jimmy Ba. 2015.
\newblock \href {http://arxiv.org/abs/1412.6980} {Adam: {A} method for
  stochastic optimization}.
\newblock In \emph{3rd International Conference on Learning Representations,
  {ICLR} 2015, San Diego, CA, USA, May 7-9, 2015, Conference Track
  Proceedings}.

\bibitem[{Kudo and Richardson(2018)}]{kudo-richardson-2018-sentencepiece}
Taku Kudo and John Richardson. 2018.
\newblock \href {https://doi.org/10.18653/v1/D18-2012} {{S}entence{P}iece: A
  simple and language independent subword tokenizer and detokenizer for neural
  text processing}.
\newblock In \emph{Proceedings of the 2018 Conference on Empirical Methods in
  Natural Language Processing: System Demonstrations}, pages 66--71, Brussels,
  Belgium. Association for Computational Linguistics.

\bibitem[{Marcus et~al.(1993)Marcus, Santorini, and
  Marcinkiewicz}]{marcus-etal-1993-building}
Mitchell~P. Marcus, Beatrice Santorini, and Mary~Ann Marcinkiewicz. 1993.
\newblock \href {https://www.aclweb.org/anthology/J93-2004} {Building a large
  annotated corpus of {E}nglish: The {P}enn {T}reebank}.
\newblock \emph{Computational Linguistics}, 19(2):313--330.

\bibitem[{Merity et~al.(2018)Merity, Keskar, and
  Socher}]{merity2018regularizing}
Stephen Merity, Nitish~Shirish Keskar, and Richard Socher. 2018.
\newblock \href {https://openreview.net/forum?id=SyyGPP0TZ} {Regularizing and
  optimizing {LSTM} language models}.
\newblock In \emph{International Conference on Learning Representations}.

\bibitem[{Merity et~al.(2016)Merity, Xiong, Bradbury, and
  Socher}]{merity2016pointer}
Stephen Merity, Caiming Xiong, James Bradbury, and Richard Socher. 2016.
\newblock \href {https://arxiv.org/abs/1609.07843} {Pointer sentinel mixture
  models}.
\newblock \emph{arXiv preprint arXiv:1609.07843}.

\bibitem[{Mielke(2019)}]{miekel2019charppl}
Sabrina~J. Mielke. 2019.
\newblock Can you compare perplexity across different segmentations?
\newblock Available in: \url{http://sjmielke.com/comparing-perplexities.htm}.

\bibitem[{Mielke et~al.(2019)Mielke, Cotterell, Gorman, Roark, and
  Eisner}]{mielke-etal-2019-kind}
Sebastian~J. Mielke, Ryan Cotterell, Kyle Gorman, Brian Roark, and Jason
  Eisner. 2019.
\newblock \href {https://www.aclweb.org/anthology/P19-1491} {What kind of
  language is hard to language-model?}
\newblock In \emph{Proceedings of the 57th Annual Meeting of the Association
  for Computational Linguistics}, pages 4975--4989, Florence, Italy.
  Association for Computational Linguistics.

\bibitem[{Mielke and Eisner(2019)}]{mielke-eisner-2019-spell}
Sebastian~J. Mielke and Jason Eisner. 2019.
\newblock \href {https://www.aaai.org/ojs/index.php/AAAI/article/view/4660}
  {Spell once, summon anywhere: A two-level open-vocabulary language model}.
\newblock In \emph{Proceedings of the AAAI Conference on Artificial
  Intelligence}, volume~33, pages 6843--6850.

\bibitem[{Mikolov et~al.(2012)Mikolov, Sutskever, Deoras, Le, Kombrink, and
  Cernocky}]{mikolov-etal-2012-subword}
Tom{\'a}{\v{s}} Mikolov, Ilya Sutskever, Anoop Deoras, Hai-Son Le, Stefan
  Kombrink, and Jan Cernocky. 2012.
\newblock Subword language modeling with neural networks.
\newblock \emph{preprint (http://www. fit. vutbr. cz/imikolov/rnnlm/char.
  pdf)}, 8:67.

\bibitem[{Nivre et~al.(2020)Nivre, de~Marneffe, Ginter, Haji{\v{c}}, Manning,
  Pyysalo, Schuster, Tyers, and Zeman}]{nivre-etal-2020-universal}
Joakim Nivre, Marie-Catherine de~Marneffe, Filip Ginter, Jan Haji{\v{c}},
  Christopher~D. Manning, Sampo Pyysalo, Sebastian Schuster, Francis Tyers, and
  Daniel Zeman. 2020.
\newblock \href {https://www.aclweb.org/anthology/2020.lrec-1.497} {{U}niversal
  {D}ependencies v2: An evergrowing multilingual treebank collection}.
\newblock In \emph{Proceedings of the 12th Language Resources and Evaluation
  Conference}, pages 4034--4043, Marseille, France. European Language Resources
  Association.

\bibitem[{Sennrich et~al.(2016)Sennrich, Haddow, and
  Birch}]{sennrich-etal-2016-neural}
Rico Sennrich, Barry Haddow, and Alexandra Birch. 2016.
\newblock \href {https://doi.org/10.18653/v1/P16-1162} {Neural machine
  translation of rare words with subword units}.
\newblock In \emph{Proceedings of the 54th Annual Meeting of the Association
  for Computational Linguistics (Volume 1: Long Papers)}, pages 1715--1725,
  Berlin, Germany. Association for Computational Linguistics.

\bibitem[{Sutskever et~al.(2011)Sutskever, Martens, and
  Hinton}]{Sutskever:2011:GTR:3104482.3104610}
Ilya Sutskever, James Martens, and Geoffrey Hinton. 2011.
\newblock \href {http://dl.acm.org/citation.cfm?id=3104482.3104610} {Generating
  text with recurrent neural networks}.
\newblock In \emph{Proceedings of the 28th International Conference on
  International Conference on Machine Learning}, ICML'11, pages 1017--1024,
  USA. Omnipress.

\bibitem[{Vania and Lopez(2017)}]{vania-lopez-2017-characters}
Clara Vania and Adam Lopez. 2017.
\newblock \href {https://doi.org/10.18653/v1/P17-1184} {From characters to
  words to in between: Do we capture morphology?}
\newblock In \emph{Proceedings of the 55th Annual Meeting of the Association
  for Computational Linguistics (Volume 1: Long Papers)}, pages 2016--2027,
  Vancouver, Canada. Association for Computational Linguistics.

\bibitem[{Virpioja et~al.(2013)Virpioja, Smit, Grönroos, and
  Kurimo}]{virpioja-2013-morfessor}
Sami Virpioja, Peter Smit, Stig-Arne Grönroos, and Mikko Kurimo. 2013.
\newblock Morfessor 2.0: Python implementation and extensions for morfessor
  baseline.
\newblock In \emph{Aalto University publication series}. Department of Signal
  Processing and Acoustics, Aalto University.

\bibitem[{Yu et~al.(2017)Yu, Kulkarni, Lee, and Kim}]{yu-etal-2017-syllable}
Seunghak Yu, Nilesh Kulkarni, Haejun Lee, and Jihie Kim. 2017.
\newblock \href {https://doi.org/10.18653/v1/W17-4113} {Syllable-level neural
  language model for agglutinative language}.
\newblock In \emph{Proceedings of the First Workshop on Subword and Character
  Level Models in {NLP}}, pages 92--96, Copenhagen, Denmark. Association for
  Computational Linguistics.

\bibitem[{Ziegler et~al.(2010)Ziegler, Bertrand, T{\'o}th, Cs{\'e}pe, Reis,
  Fa{\'\i}sca, Saine, Lyytinen, Vaessen, and
  Blomert}]{ziegler-etal-2010-orthographic}
Johannes~C Ziegler, Daisy Bertrand, D{\'e}nes T{\'o}th, Val{\'e}ria Cs{\'e}pe,
  Alexandra Reis, Lu{\'\i}s Fa{\'\i}sca, Nina Saine, Heikki Lyytinen, Anniek
  Vaessen, and Leo Blomert. 2010.
\newblock Orthographic depth and its impact on universal predictors of reading:
  A cross-language investigation.
\newblock \emph{Psychological science}, 21(4):551--559.

\end{thebibliography}
\bibliographystyle{acl_natbib}

\appendix

\newpage
\section{Datasets}
\label{app:datasets}

Table \ref{tab:data-splits} shows the size of the training, validation and test splits for all the datasets used in the study.

\begin{table}[h!]
    \centering
    \setlength\tabcolsep{4pt}
    \resizebox{\linewidth}{!}{%
    \begin{tabular}{l|rrr|rrr|rrr}
   & \multicolumn{3}{c|}{Train} & \multicolumn{3}{c|}{Valid} & \multicolumn{3}{c}{Test} \\ \hline
&  Word & Syl & Char & Word & Syl & Char & Word & Syl & Char \\ \hline
bg & 125 & 386 & 710 & 16 & 50 & 92 & 16 & 49 & 90 \\ 
ca & 436 & 1,123 & 2,341 & 59 & 152 & 317 & 61 & 157 & 327 \\ 
cs & 1,158 & 3,546 & 6,868 & 157 & 482 & 933 & 172 & 524 & 1,012 \\ 
da & 81 & 215 & 442 & 10 & 28 & 57 & 10 & 27 & 56 \\ 
de & 260 & 735 & 1,637 & 12 & 34 & 75 & 16 & 45 & 102 \\ 
en\textsubscript{UD} & 210 & 488 & 1,061 & 26 & 61 & 133 & 26 & 61 & 132 \\ 
en\textsubscript{wt2} & 2,089 & 4,894 & 10,902 & 218 & 505 & 1,157 & 246 & 568 & 1,304 \\ 
es & 376 & 1,060 & 2,043 & 37 & 103 & 198 & 12 & 33 & 64 \\ 
fi & 165 & 595 & 1,224 & 19 & 67 & 137 & 21 & 76 & 155 \\ 
fr & 360 & 837 & 1,959 & 36 & 84 & 197 & 10 & 23 & 54 \\ 
hr & 154 & 484 & 930 & 20 & 62 & 119 & 23 & 75 & 145 \\ 
it & 263 & 762 & 1,504 & 11 & 32 & 64 & 10 & 28 & 57 \\ 
lv & 113 & 349 & 690 & 19 & 58 & 115 & 20 & 59 & 116 \\ 
nl & 187 & 488 & 1,074 & 12 & 30 & 66 & 11 & 31 & 68 \\ 
pl & 102 & 293 & 589 & 13 & 37 & 73 & 13 & 37 & 74 \\ 
pt & 192 & 551 & 1,040 & 10 & 29 & 54 & 9 & 27 & 51 \\ 
ro & 183 & 549 & 1,056 & 17 & 51 & 98 & 16 & 48 & 94 \\ 
ru & 867 & 2,707 & 5,411 & 118 & 364 & 722 & 117 & 360 & 717 \\ 
sk & 80 & 232 & 437 & 12 & 39 & 76 & 13 & 41 & 80 \\ 
tk & 38 & 126 & 242 & 10 & 33 & 63 & 10 & 33 & 64 \\ 
uk & 88 & 289 & 501 & 12 & 41 & 71 & 16 & 56 & 99 \\   \hline
    \end{tabular}
    }
    \caption{Total number of tokens (in thousands) at word, syllable and character-level for all the splits.}
    \label{tab:data-splits}
\end{table}

\section{Segmentation}
\label{app:segmentation}

\paragraph{Tools} We list the tools for rule-based syllabification and dictionary-based hyphenation:
\begin{itemize}
    \item English syllabification: Extracted from \url{https://www.howmanysyllables.com/}
    \item Spanish syllabification: \url{https://pypi.org/project/pylabeador/}
    \item Russian syllabification: \url{https://github.com/Koziev/rusyllab}
    \item Finnish syllabification: \url{https://github.com/tsnaomi/finnsyll}
    \item Turkish syllabification: \url{https://github.com/MeteHanC/turkishnlp}
    \item Hyphenation: PyPhen (\url{https://pyphen.org/}), which is based on Hunspell dictionaries.
\end{itemize} 

\paragraph{Format} For syllables, we adopt the segmentation format used by SentencePiece~\cite{kudo-richardson-2018-sentencepiece} to separate subwords: ``A @ syl la ble @ con tains @ a @ sin gle @ vow el @ u nit", where ``@'' is a special token that indicates the word boundary. We also evaluated syllables with a segmentation format like in \citet{sennrich-etal-2016-neural}: ``A syl@ la@ ble con@ tains a ...'', but we obtained lower performance in general.

\section{Model and Training}
\label{app:model}
In contrast with the default settings, we use a smaller embedding size of 500 units for faster training. Additionally, we have 3 layers of depth, 1152 of hidden layer size and a dropout of 0.15. We train for 25 epochs with a batch size of 64, a learning rate of 0.002 and Adam optimiser \cite{kingma-ba-2015-adam} with default parameters. We fit the model using the one cycle policy and an early stopping of 4. We run our experiments in a NVIDIA Titan Xp.

\section{Validation results}
\label{app:validation}

\begin{figure*}
\begin{center}
\centering

\begin{subfigure}[t]{0.24\linewidth}
\includegraphics[width=\linewidth,clip]{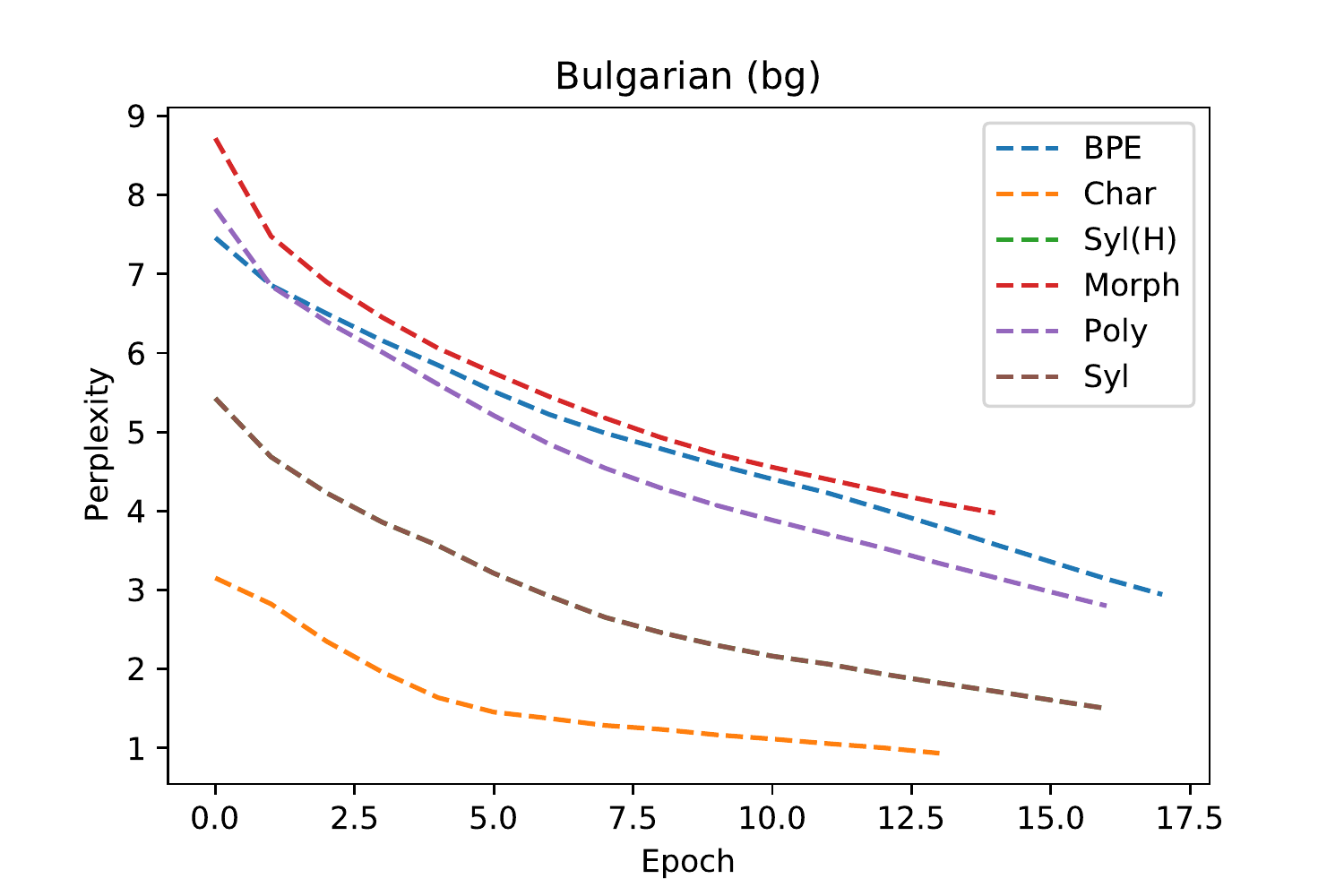}
\end{subfigure}
\begin{subfigure}[t]{0.24\linewidth}
\includegraphics[width=\linewidth,clip]{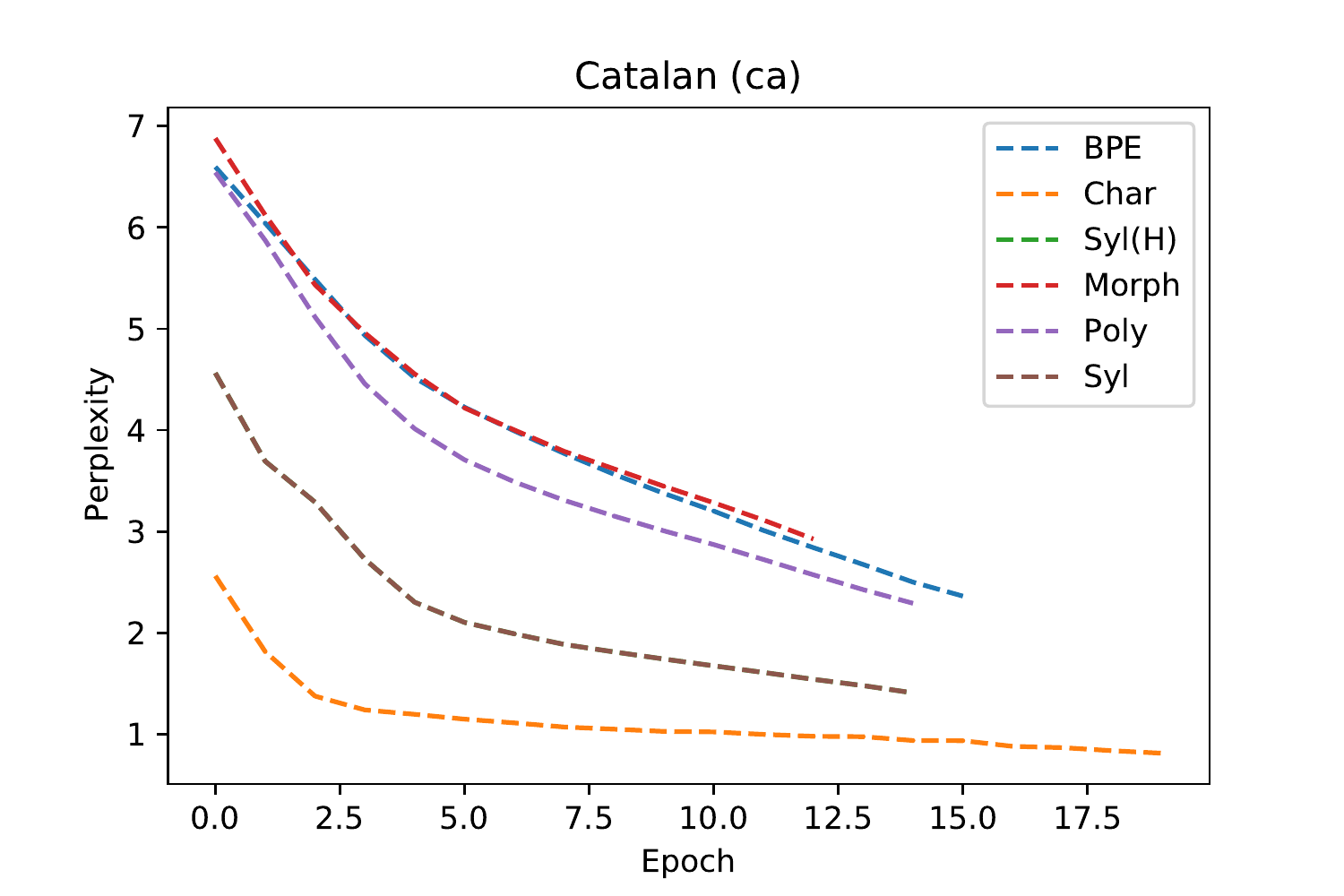}
\end{subfigure}
\begin{subfigure}[t]{0.24\linewidth}
\includegraphics[width=\linewidth,clip]{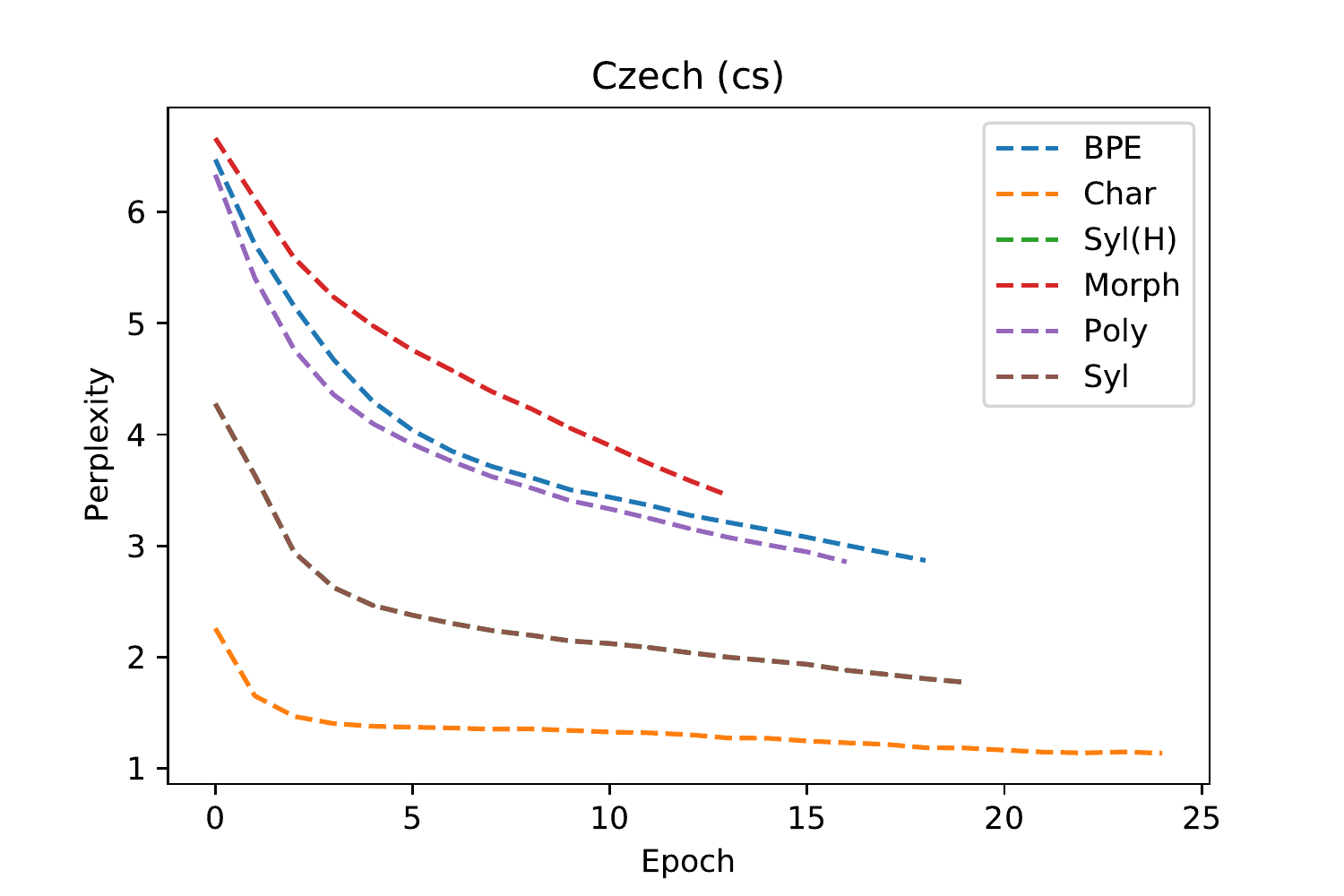}
\end{subfigure}
\begin{subfigure}[t]{0.24\linewidth}
\includegraphics[width=\linewidth,clip]{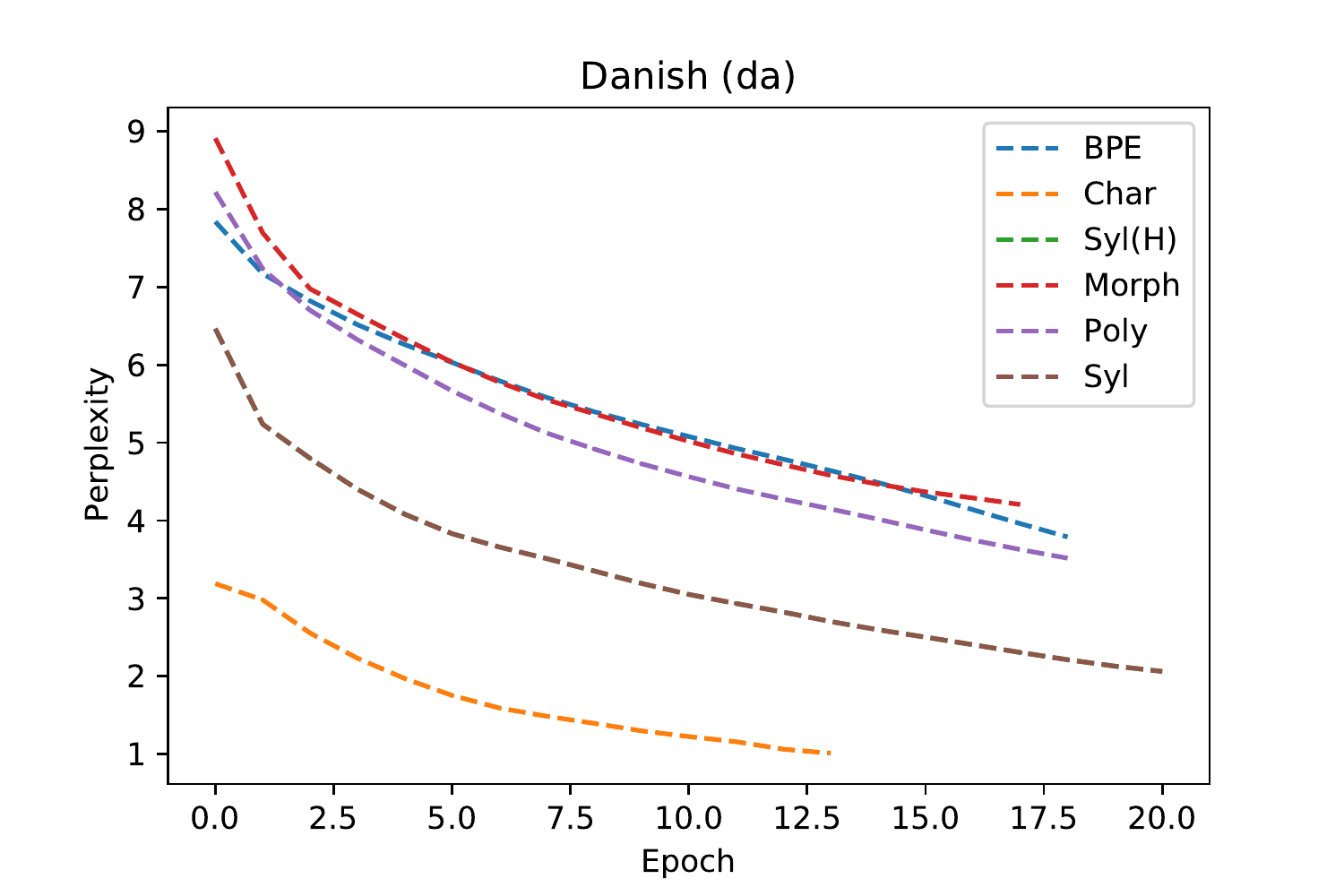}
\end{subfigure}

\begin{subfigure}[t]{0.24\linewidth}
\includegraphics[width=\linewidth,clip]{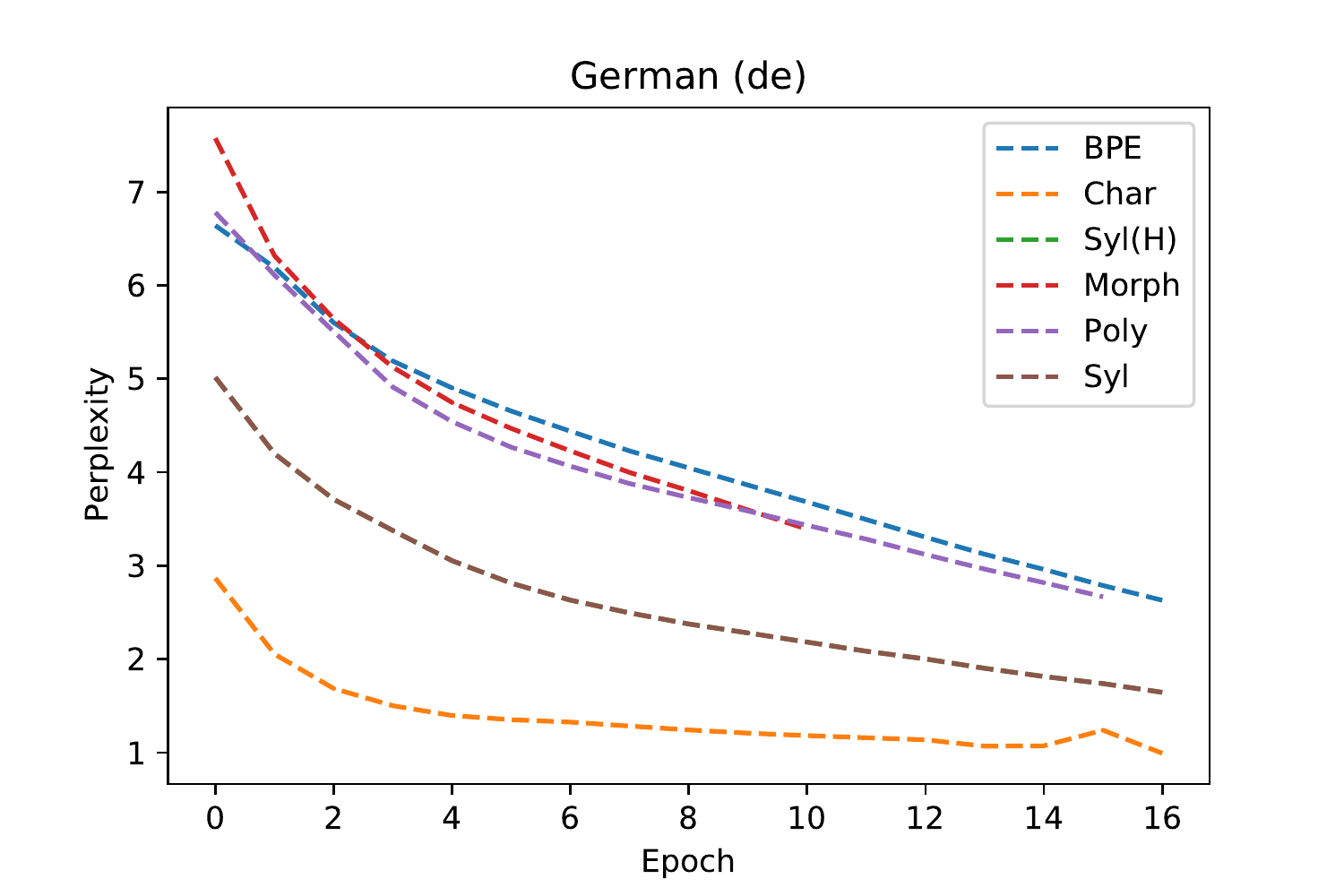}
\end{subfigure}
\begin{subfigure}[t]{0.24\linewidth}
\includegraphics[width=\linewidth,clip]{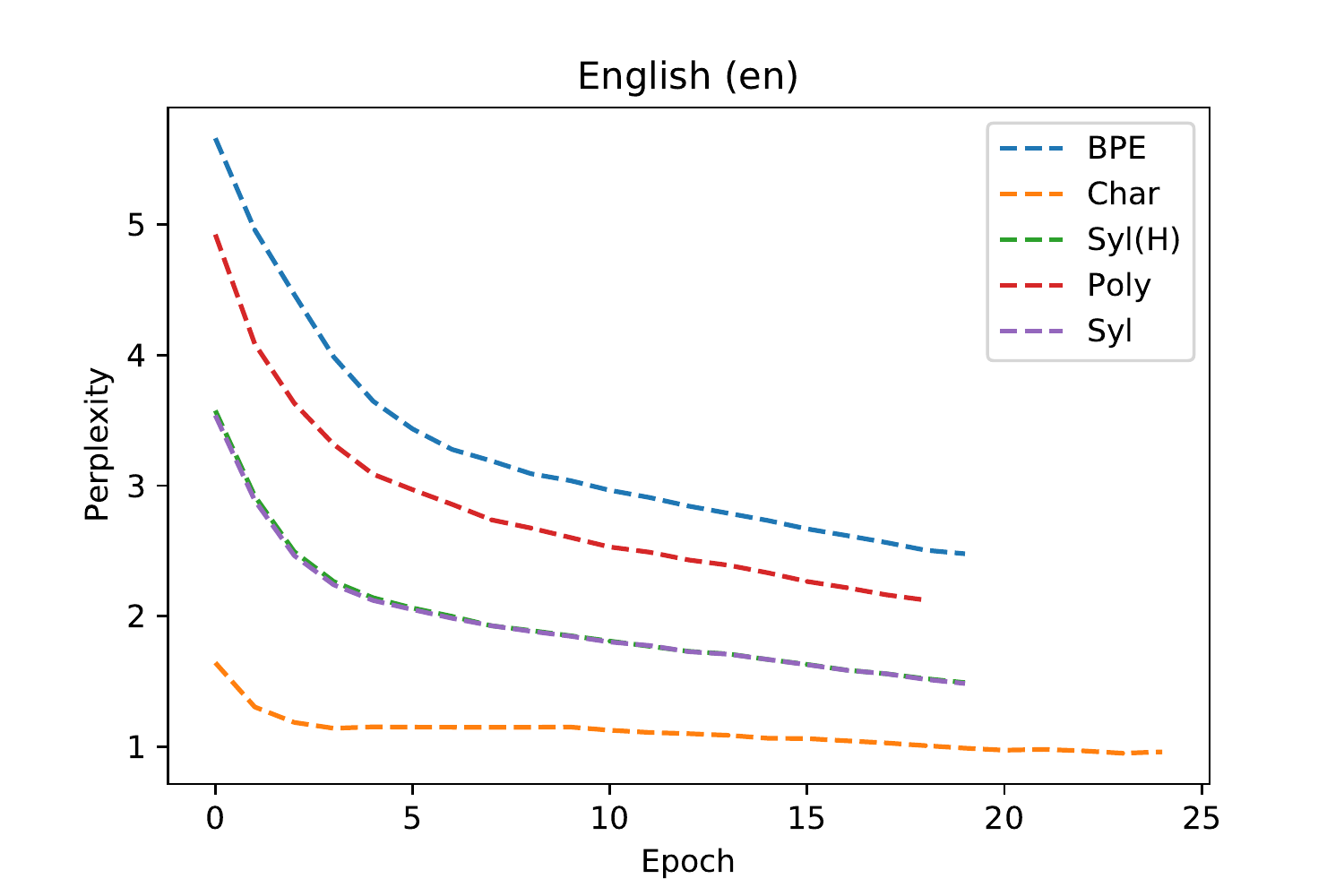}
\end{subfigure}
\begin{subfigure}[t]{0.24\linewidth}
\includegraphics[width=\linewidth,clip]{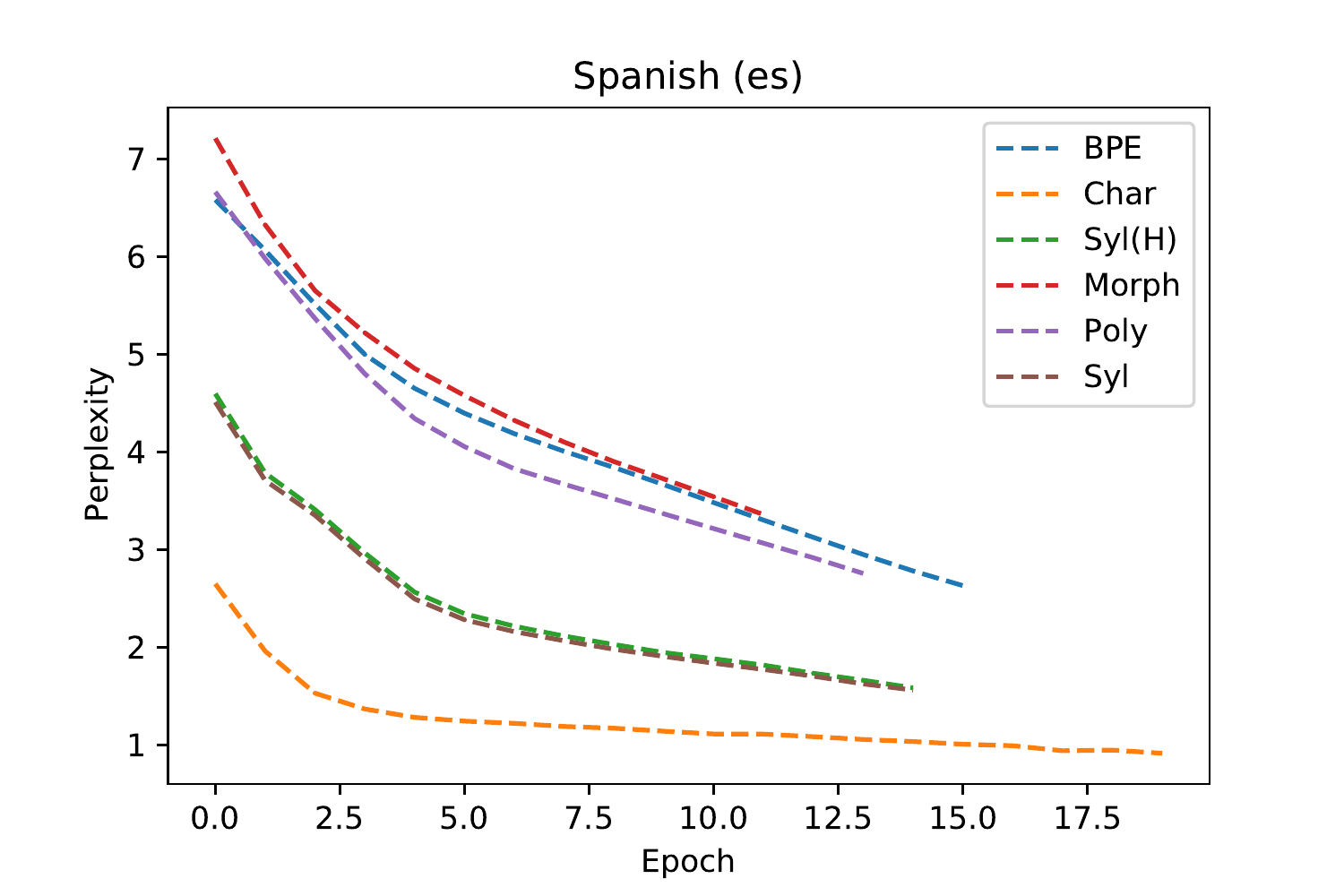}
\end{subfigure}
\begin{subfigure}[t]{0.24\linewidth}
\includegraphics[width=\linewidth,clip]{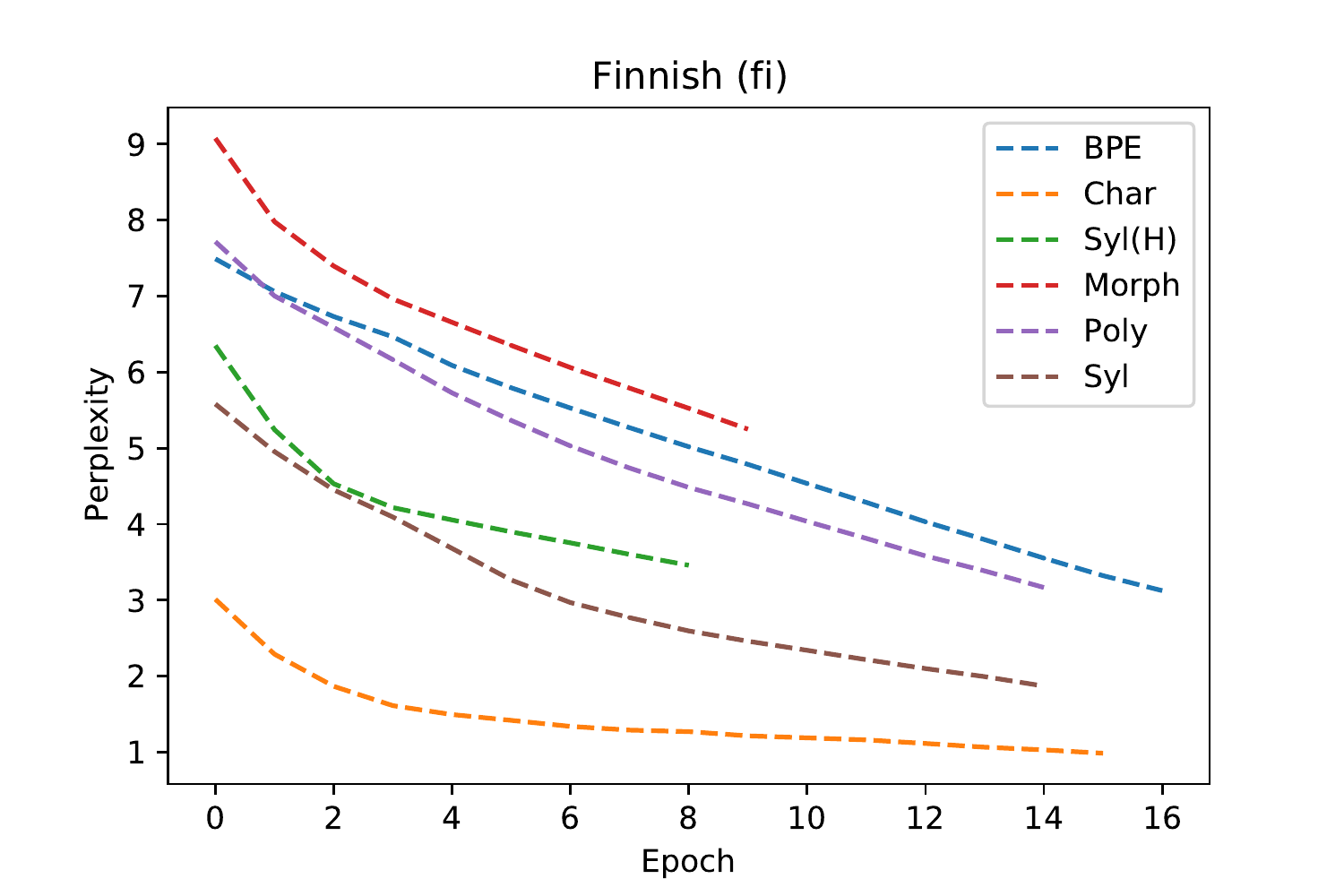}
\end{subfigure}

\begin{subfigure}[t]{0.24\linewidth}
\includegraphics[width=\linewidth,clip]{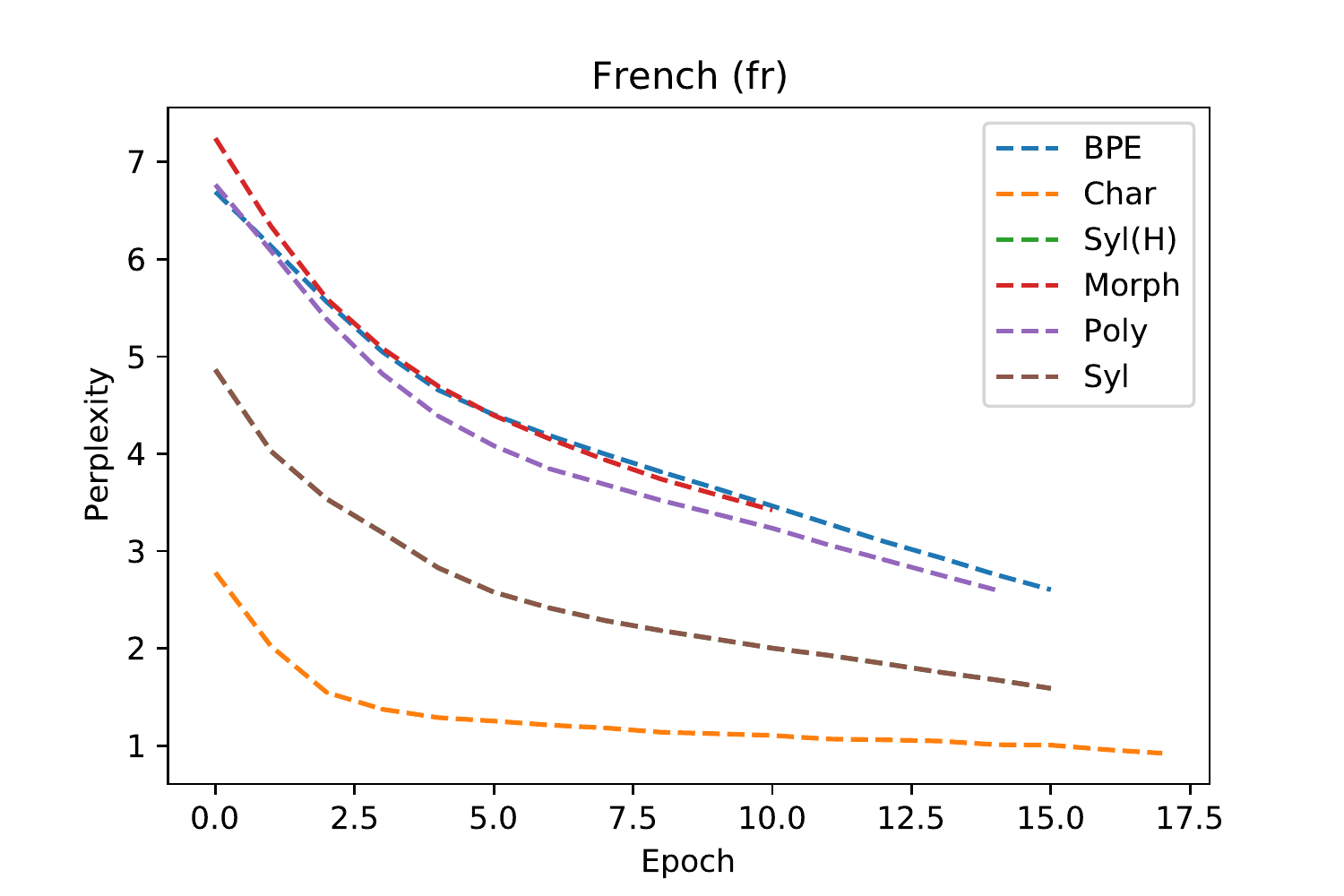}
\end{subfigure}
\begin{subfigure}[t]{0.24\linewidth}
\includegraphics[width=\linewidth,clip]{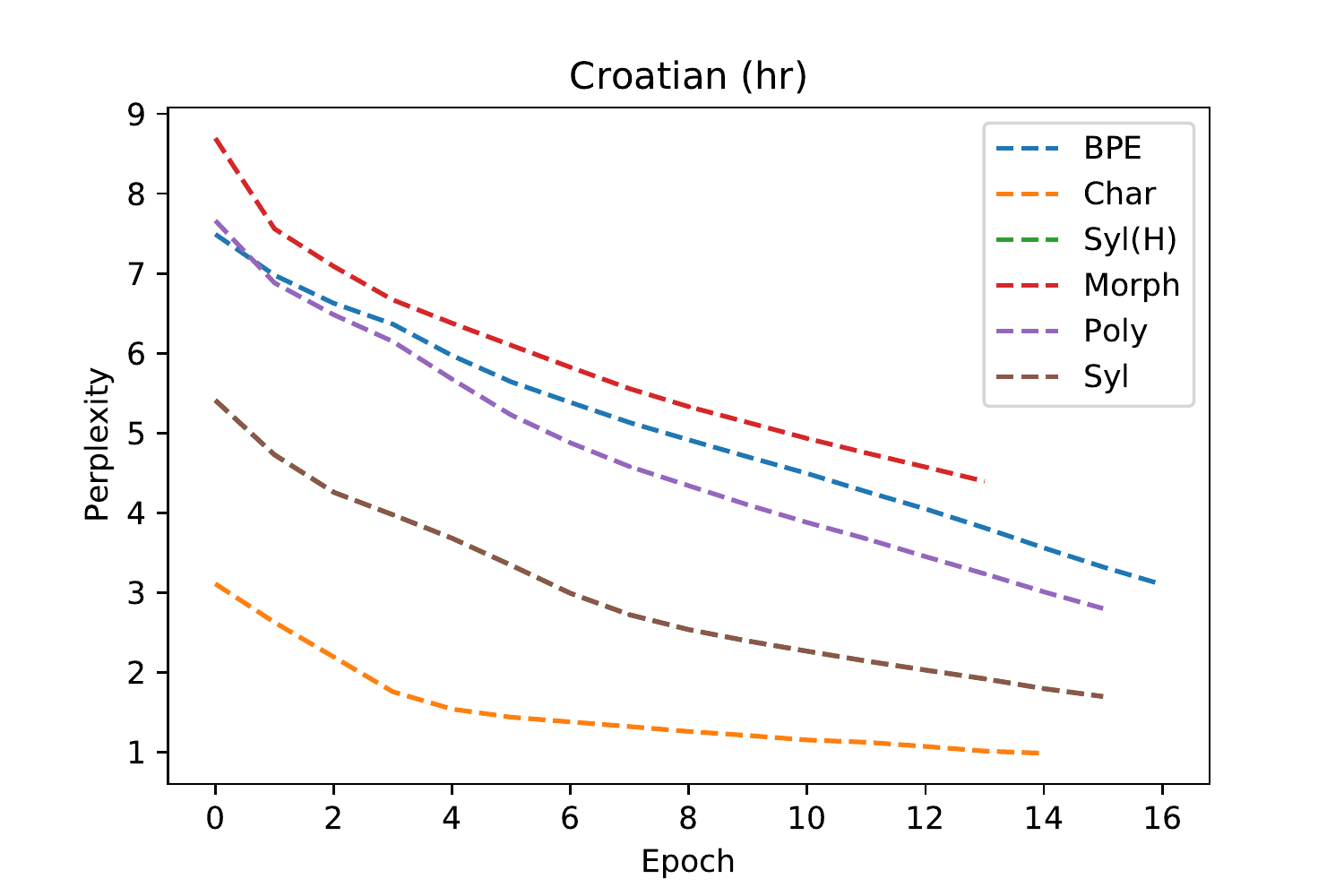}
\end{subfigure}
\begin{subfigure}[t]{0.24\linewidth}
\includegraphics[width=\linewidth,clip]{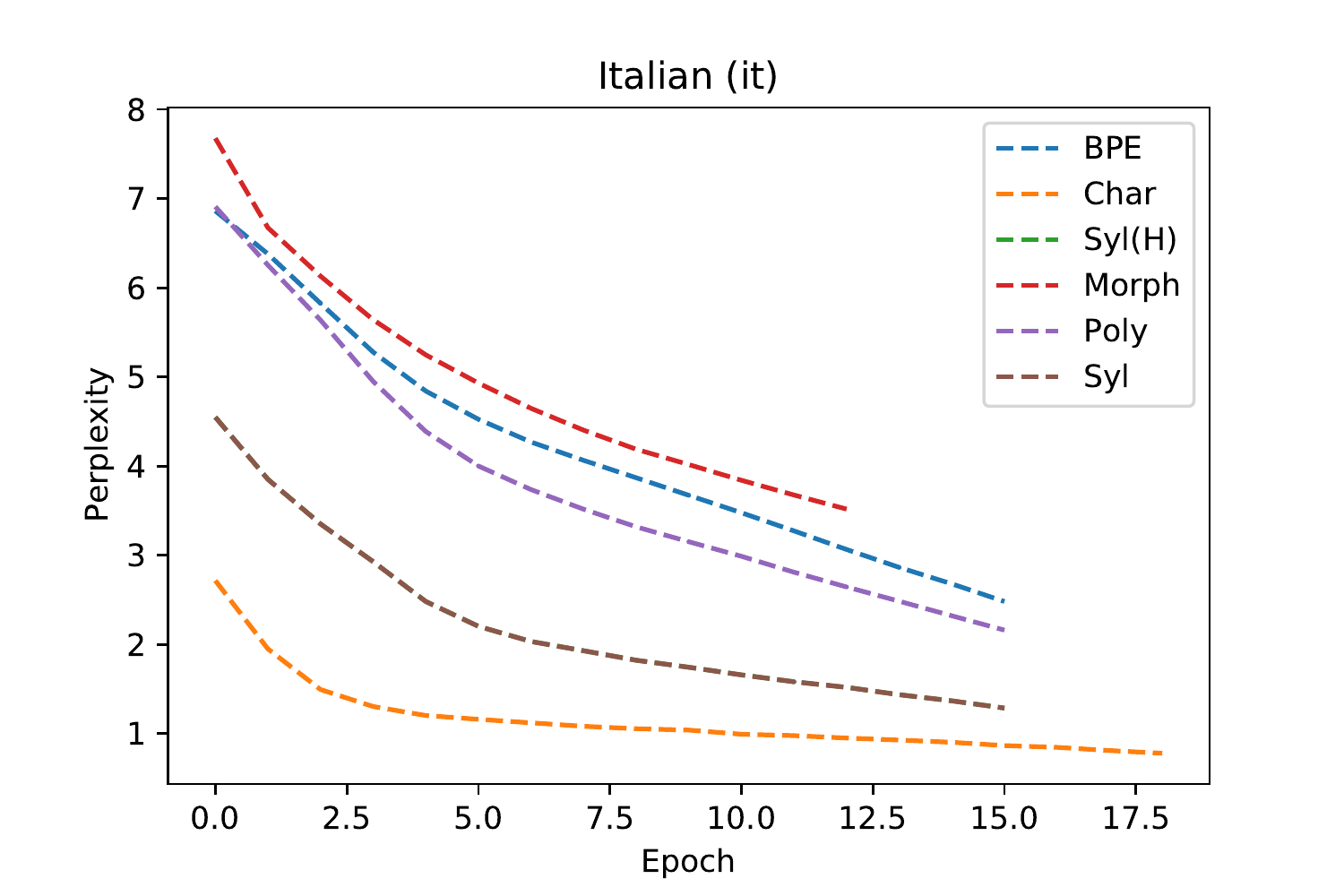}
\end{subfigure}
\begin{subfigure}[t]{0.24\linewidth}
\includegraphics[width=\linewidth,clip]{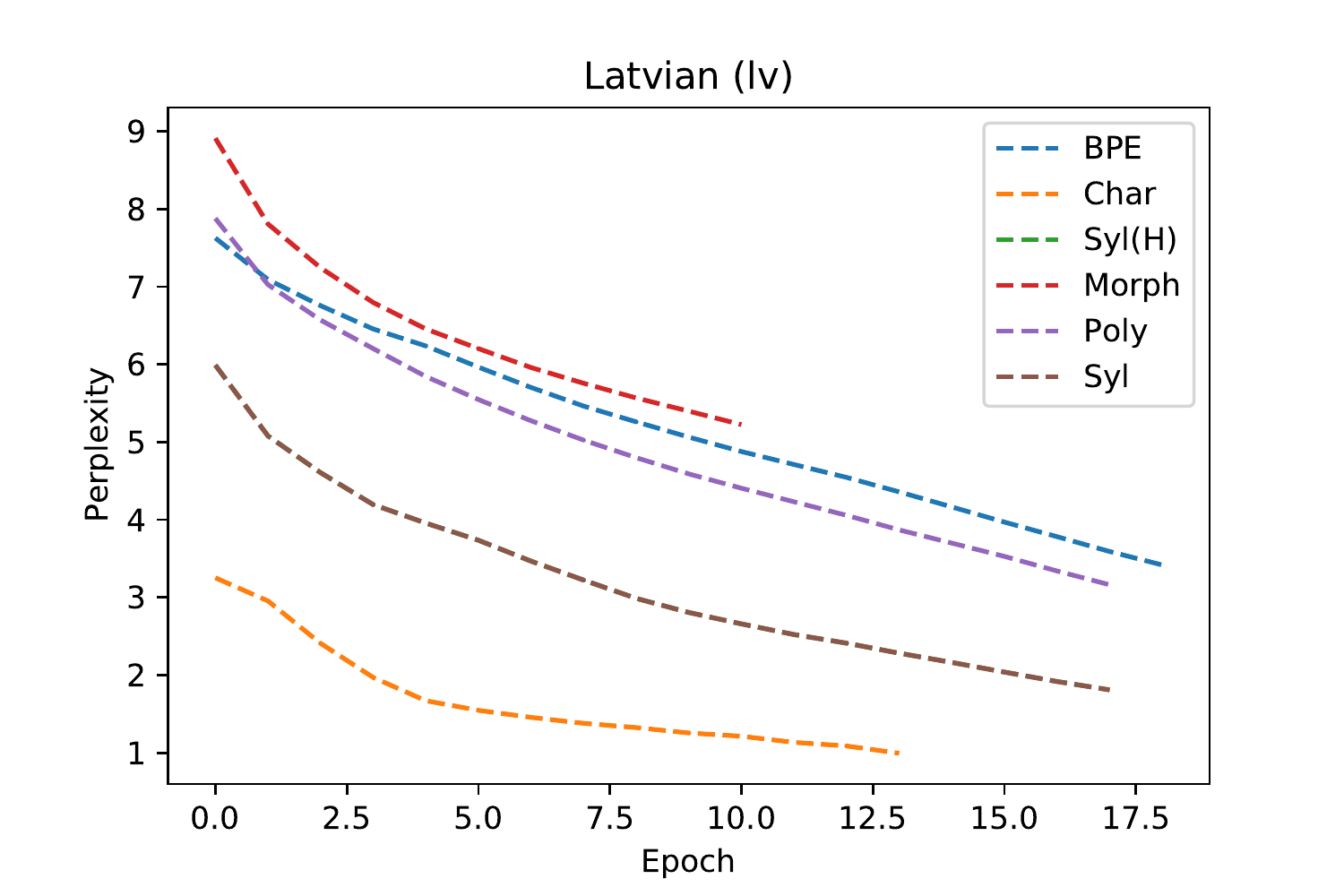}
\end{subfigure}

\begin{subfigure}[t]{0.24\linewidth}
\includegraphics[width=\linewidth,clip]{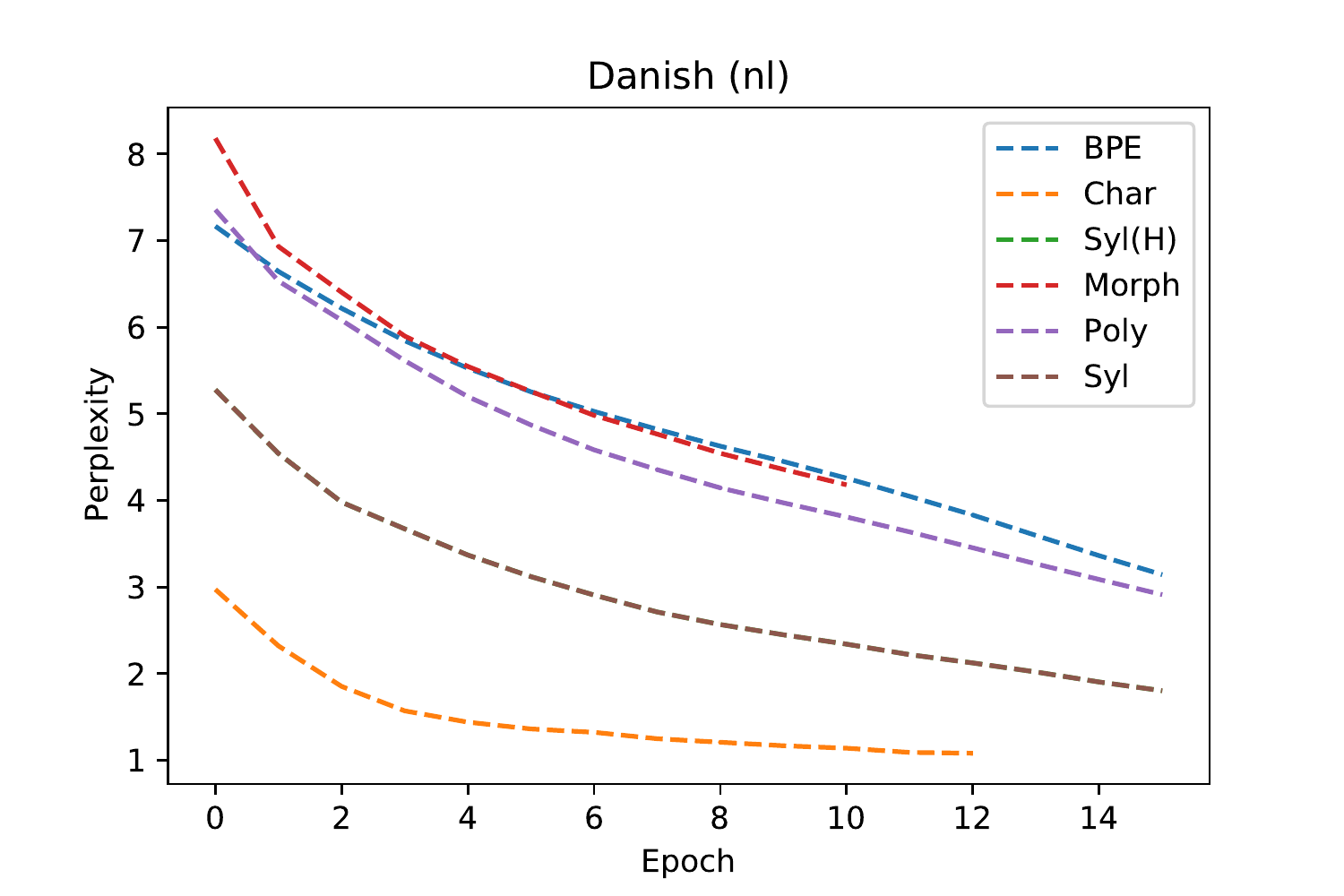}
\end{subfigure}
\begin{subfigure}[t]{0.24\linewidth}
\includegraphics[width=\linewidth,clip]{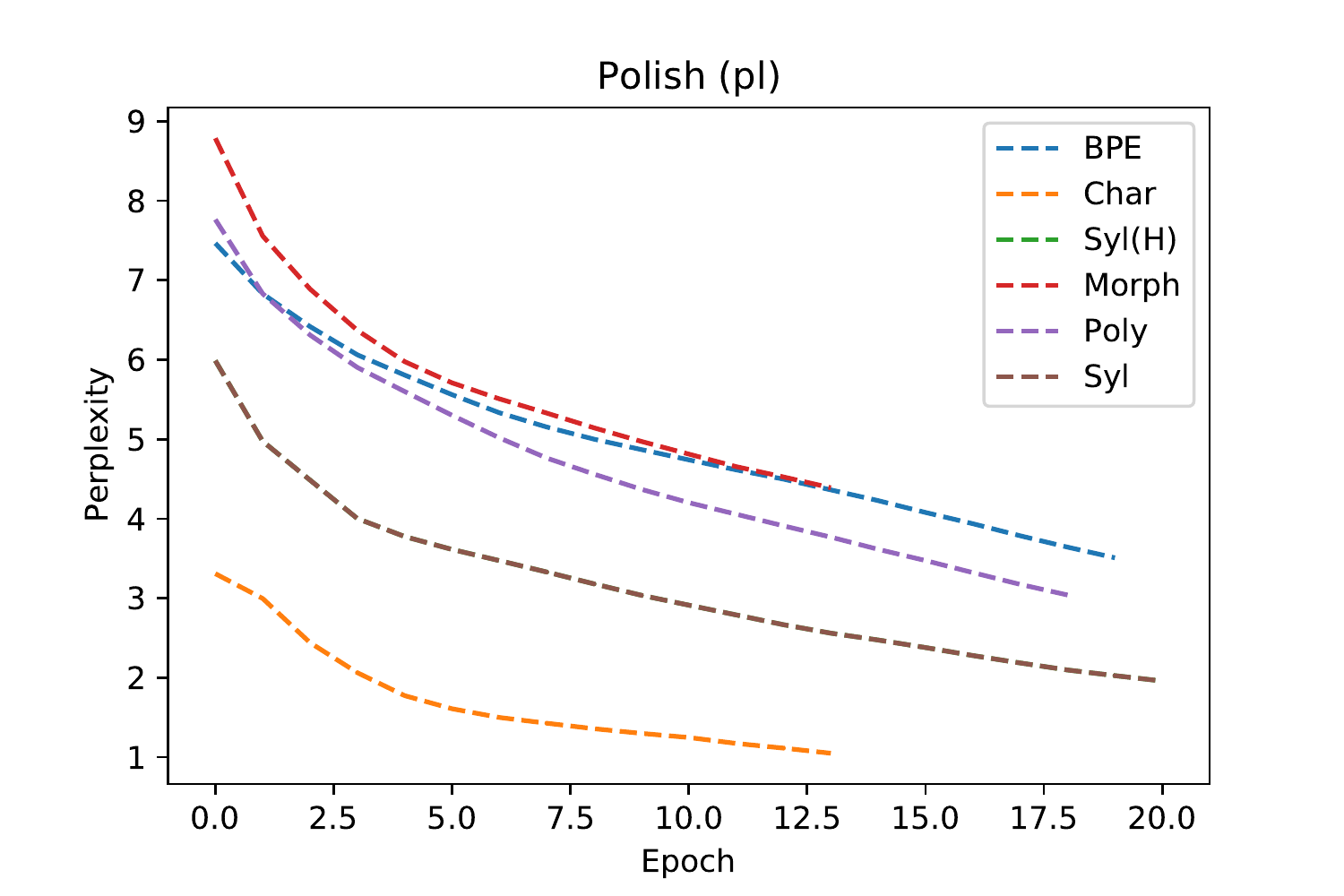}
\end{subfigure}
\begin{subfigure}[t]{0.24\linewidth}
\includegraphics[width=\linewidth,clip]{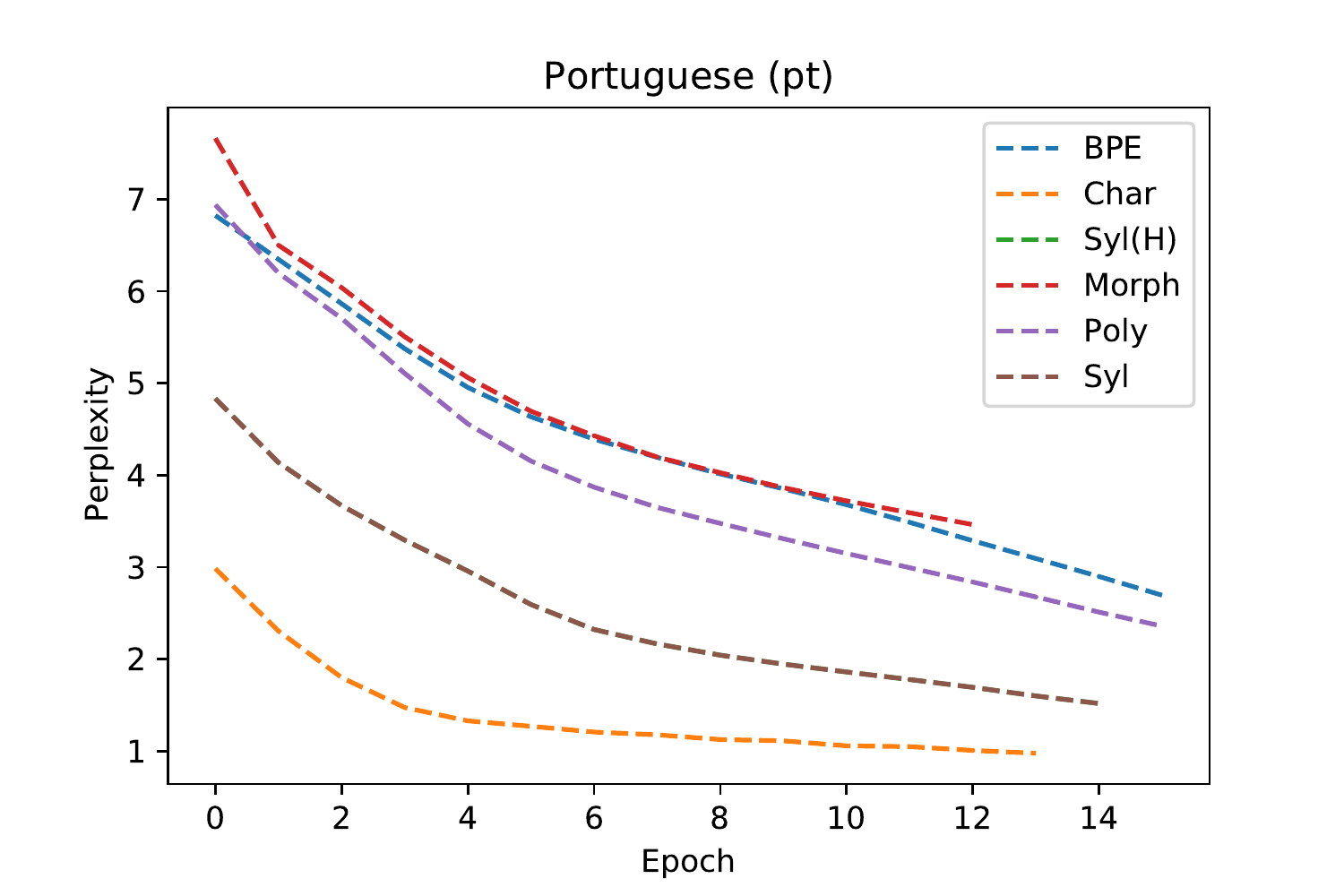}
\end{subfigure}
\begin{subfigure}[t]{0.24\linewidth}
\includegraphics[width=\linewidth,clip]{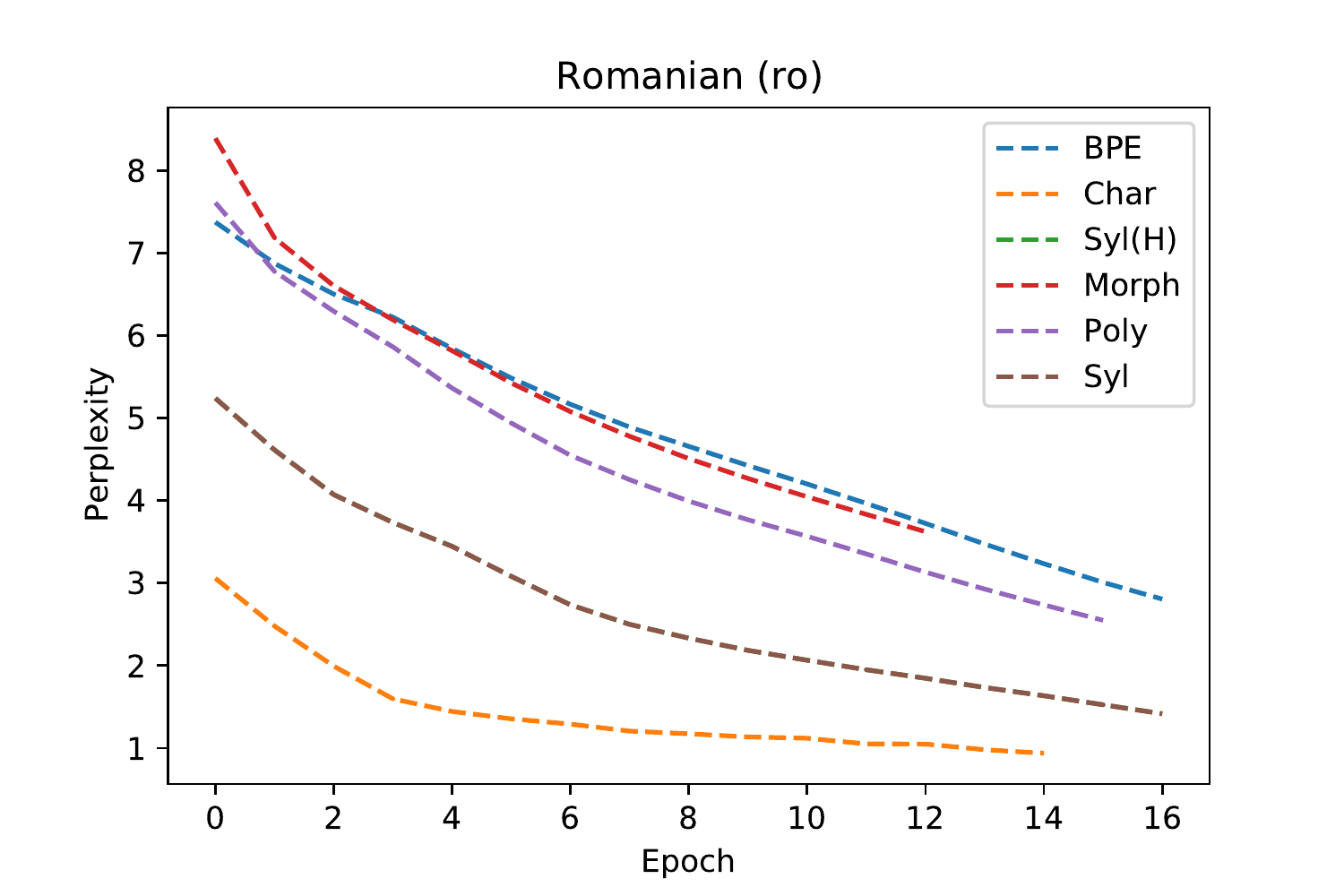}
\end{subfigure}

\begin{subfigure}[t]{0.24\linewidth}
\includegraphics[width=\linewidth,clip]{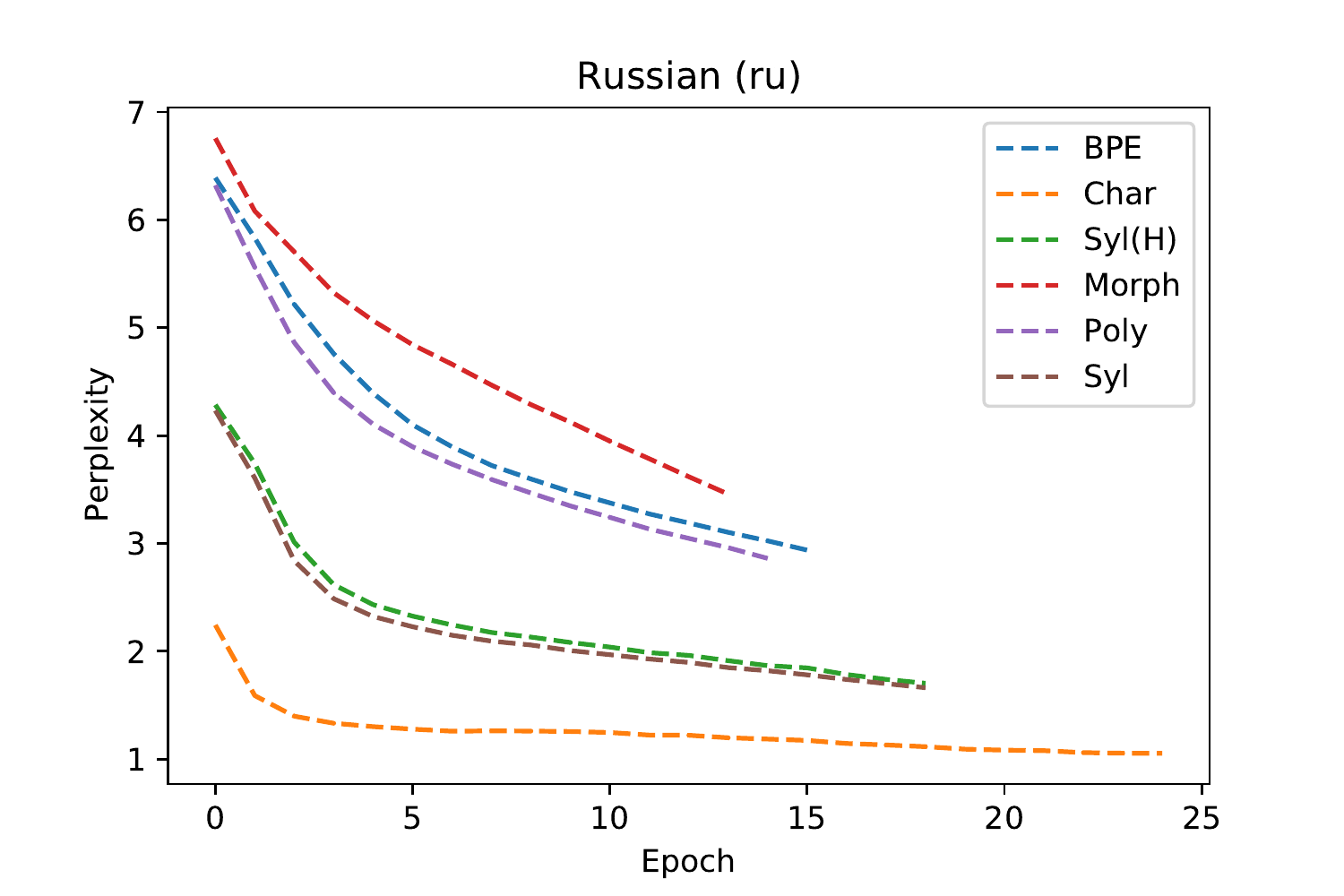}
\end{subfigure}
\begin{subfigure}[t]{0.24\linewidth}
\includegraphics[width=\linewidth,clip]{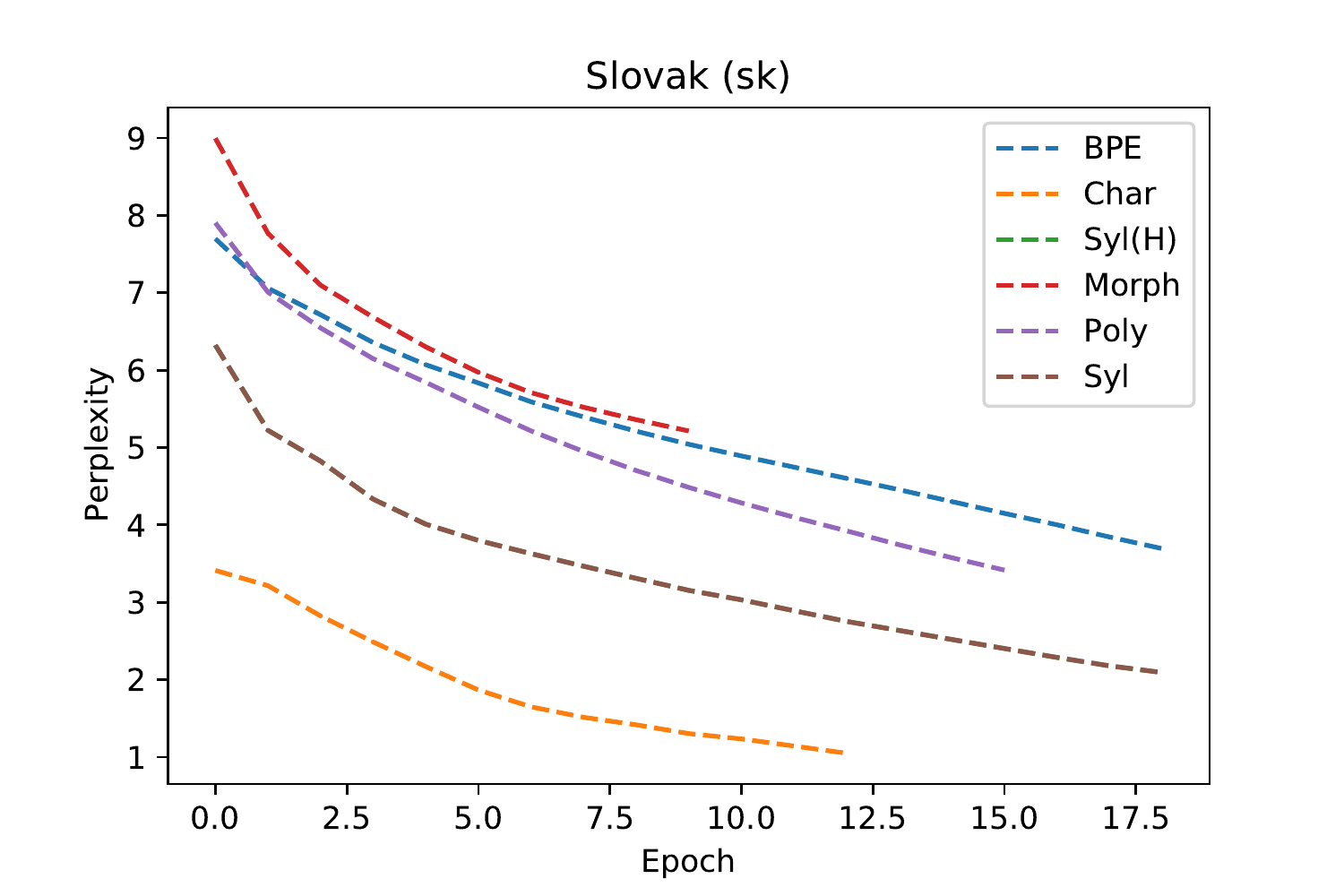}
\end{subfigure}
\begin{subfigure}[t]{0.24\linewidth}
\includegraphics[width=\linewidth,clip]{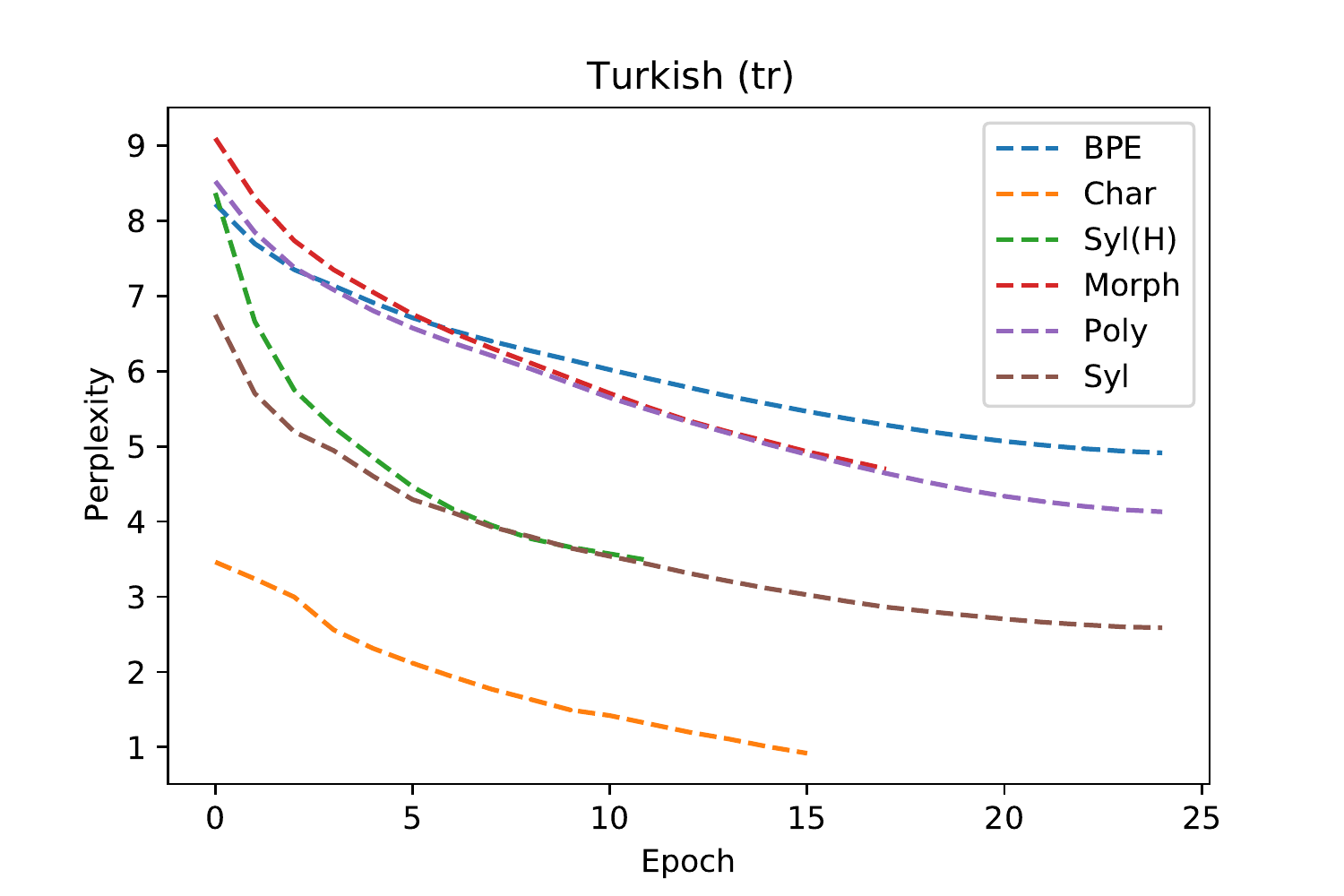}
\end{subfigure}
\begin{subfigure}[t]{0.24\linewidth}
\includegraphics[width=\linewidth,clip]{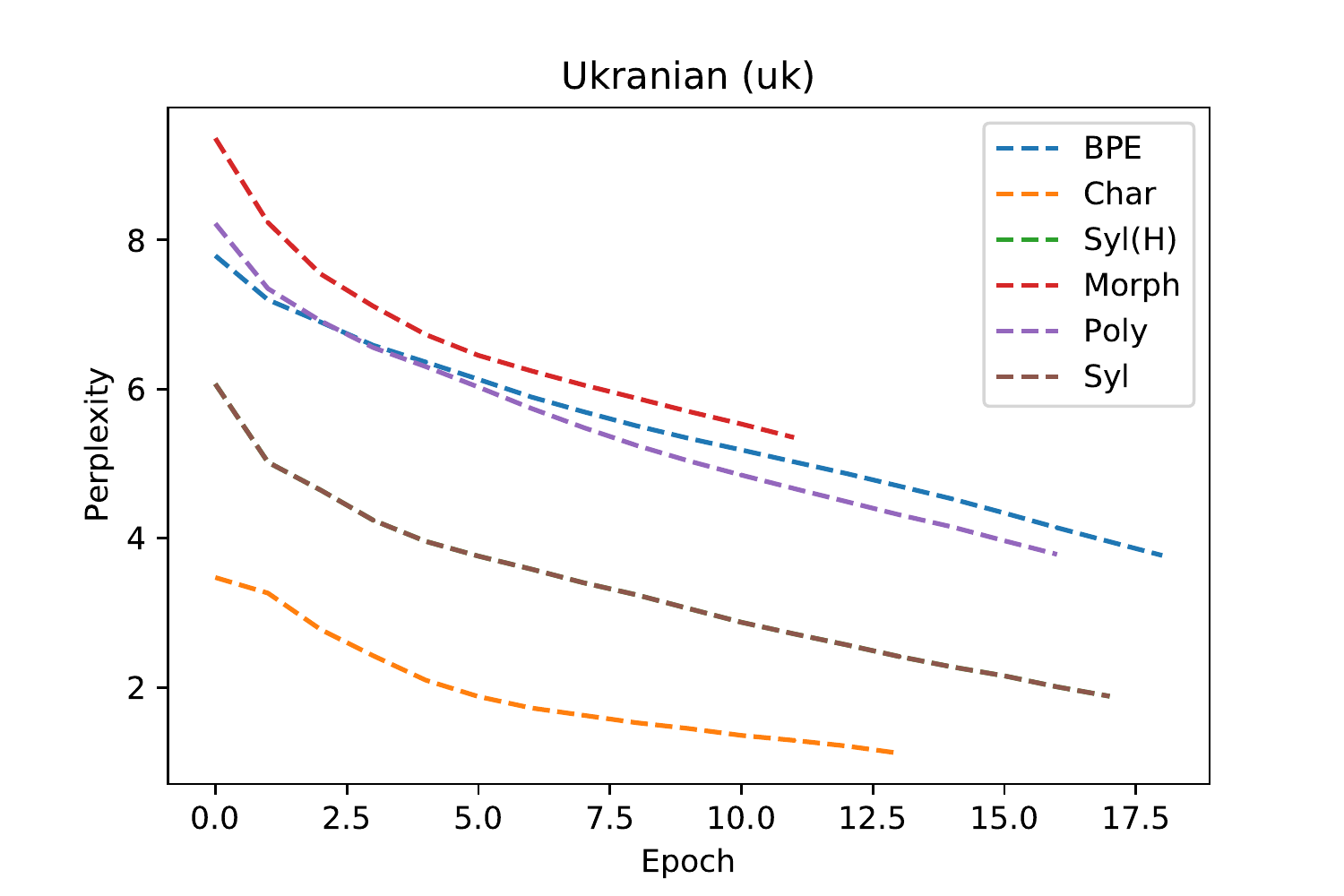}
\end{subfigure}

\caption{Validation perplexity for all the UD treebanks.} 
\label{fig:convergence-all}
\end{center}
\end{figure*}

\begin{figure*}
\begin{center}
\centering

\begin{subfigure}[t]{0.24\linewidth}
\includegraphics[width=\linewidth,clip]{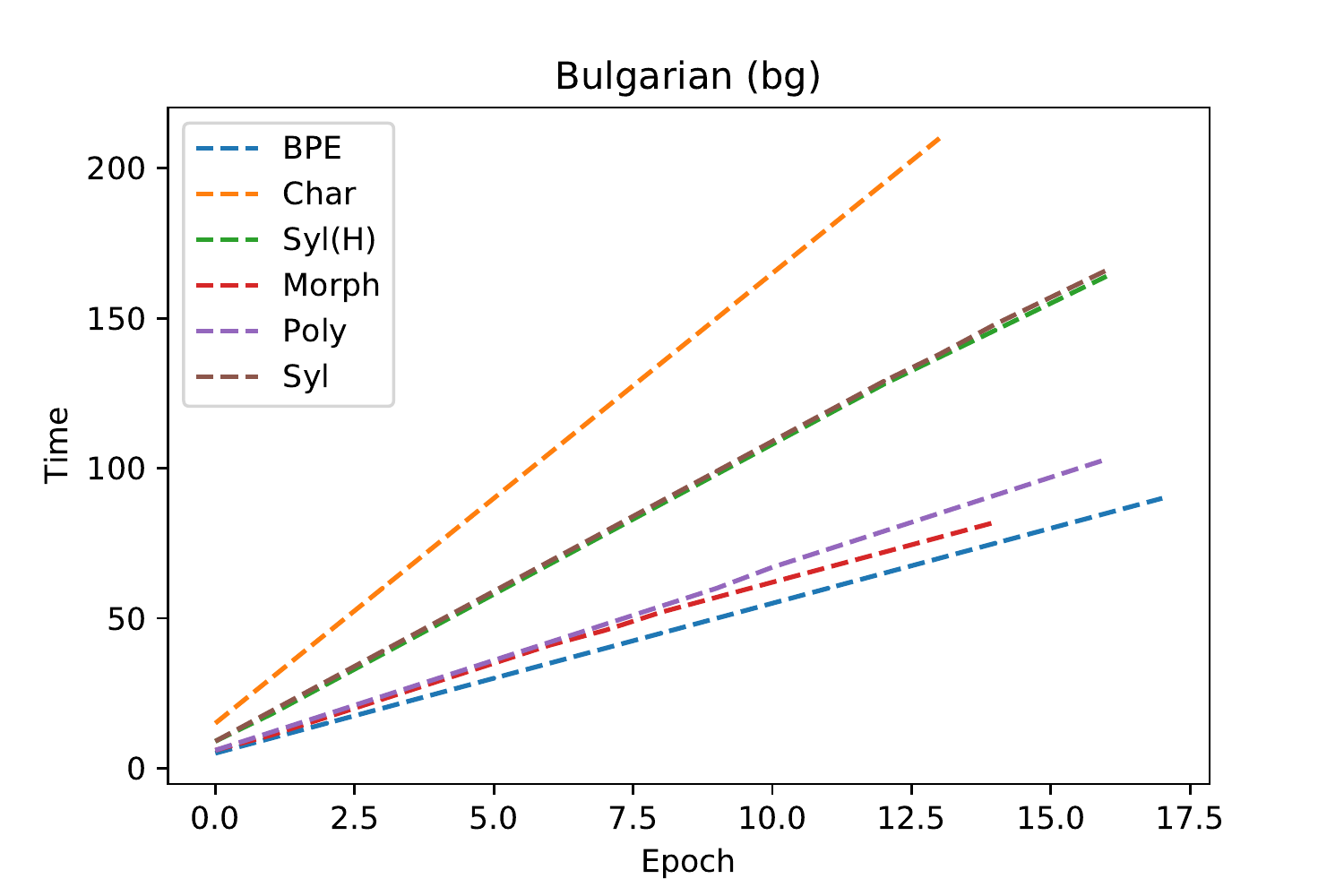}
\end{subfigure}
\begin{subfigure}[t]{0.24\linewidth}
\includegraphics[width=\linewidth,clip]{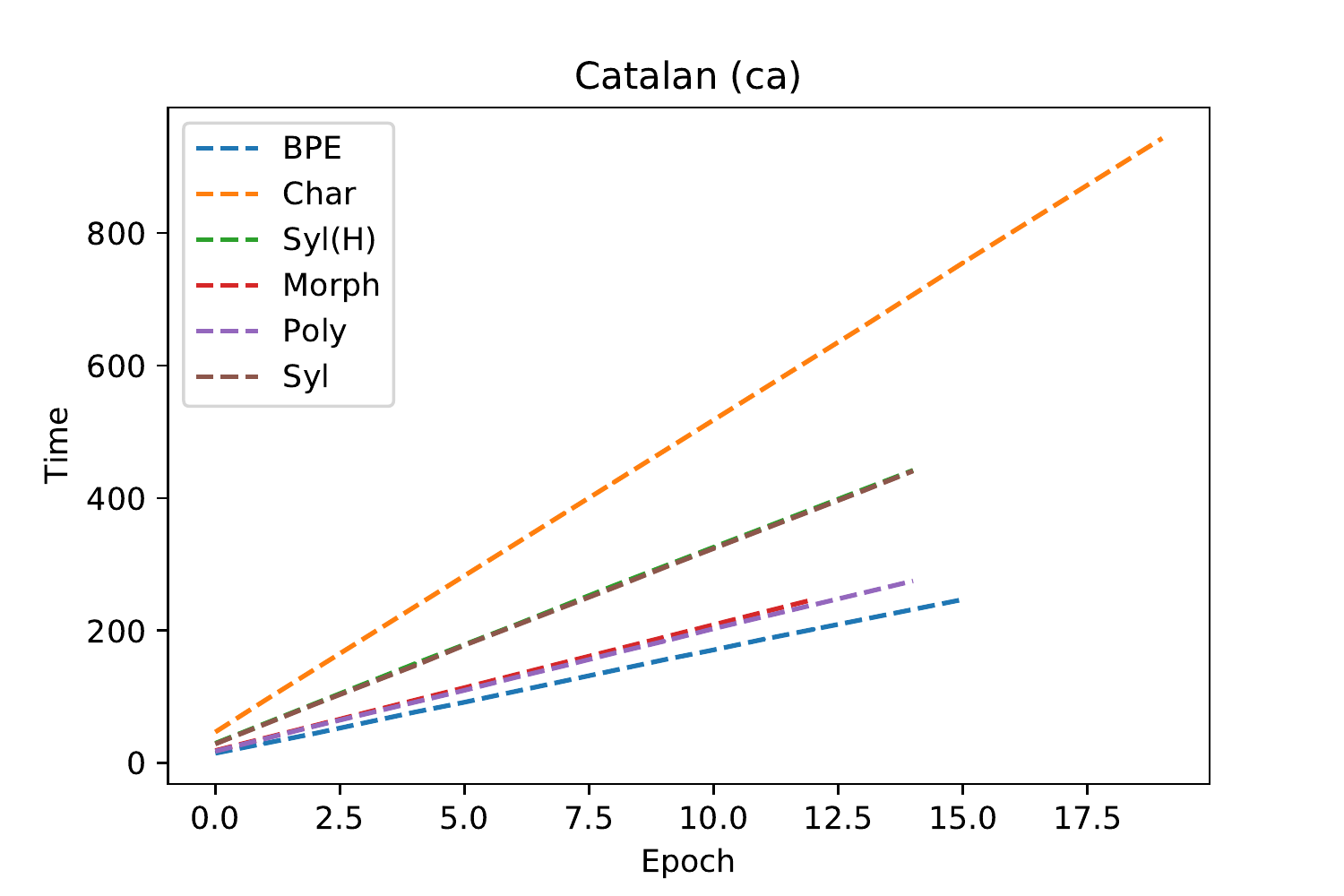}
\end{subfigure}
\begin{subfigure}[t]{0.24\linewidth}
\includegraphics[width=\linewidth,clip]{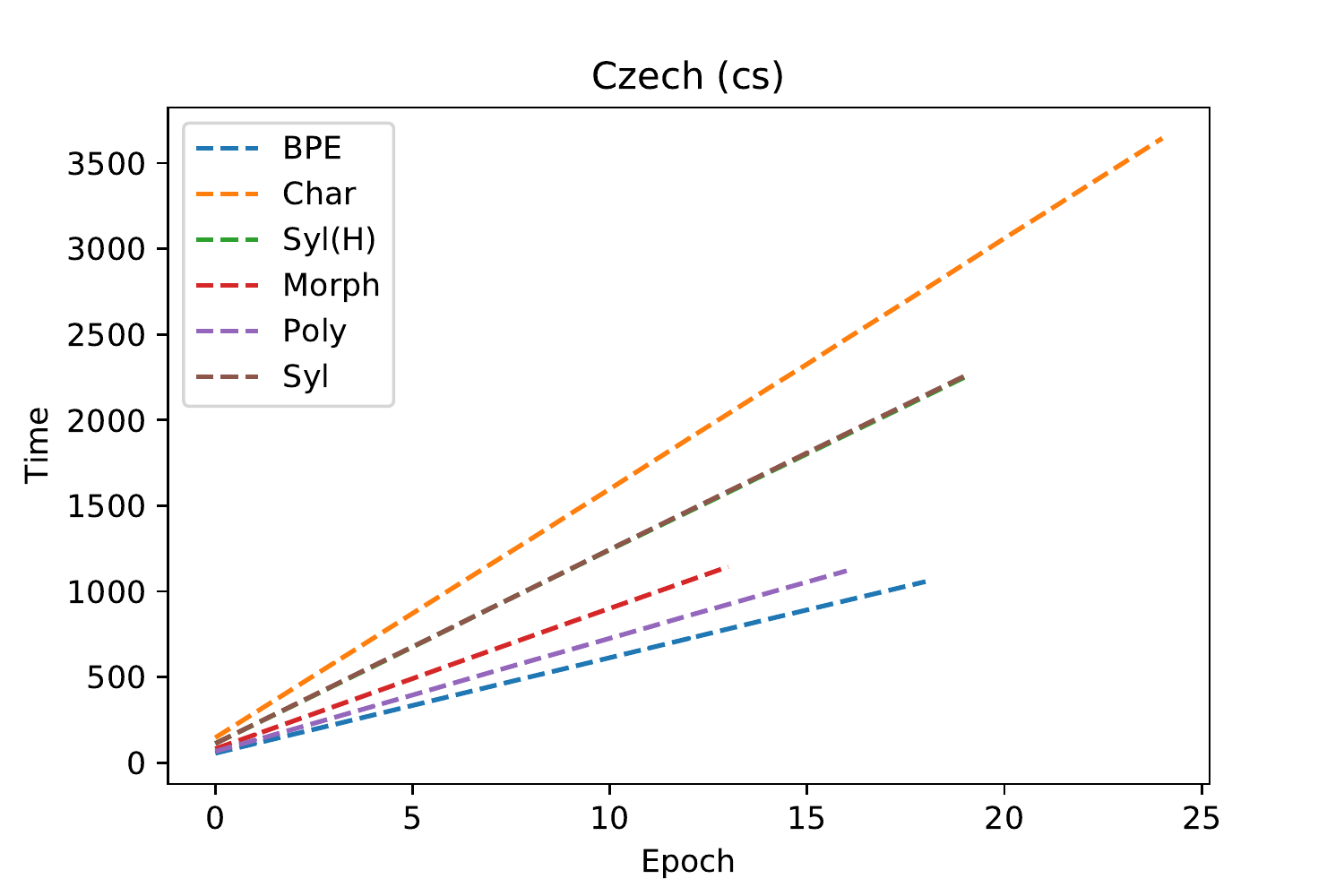}
\end{subfigure}
\begin{subfigure}[t]{0.24\linewidth}
\includegraphics[width=\linewidth,clip]{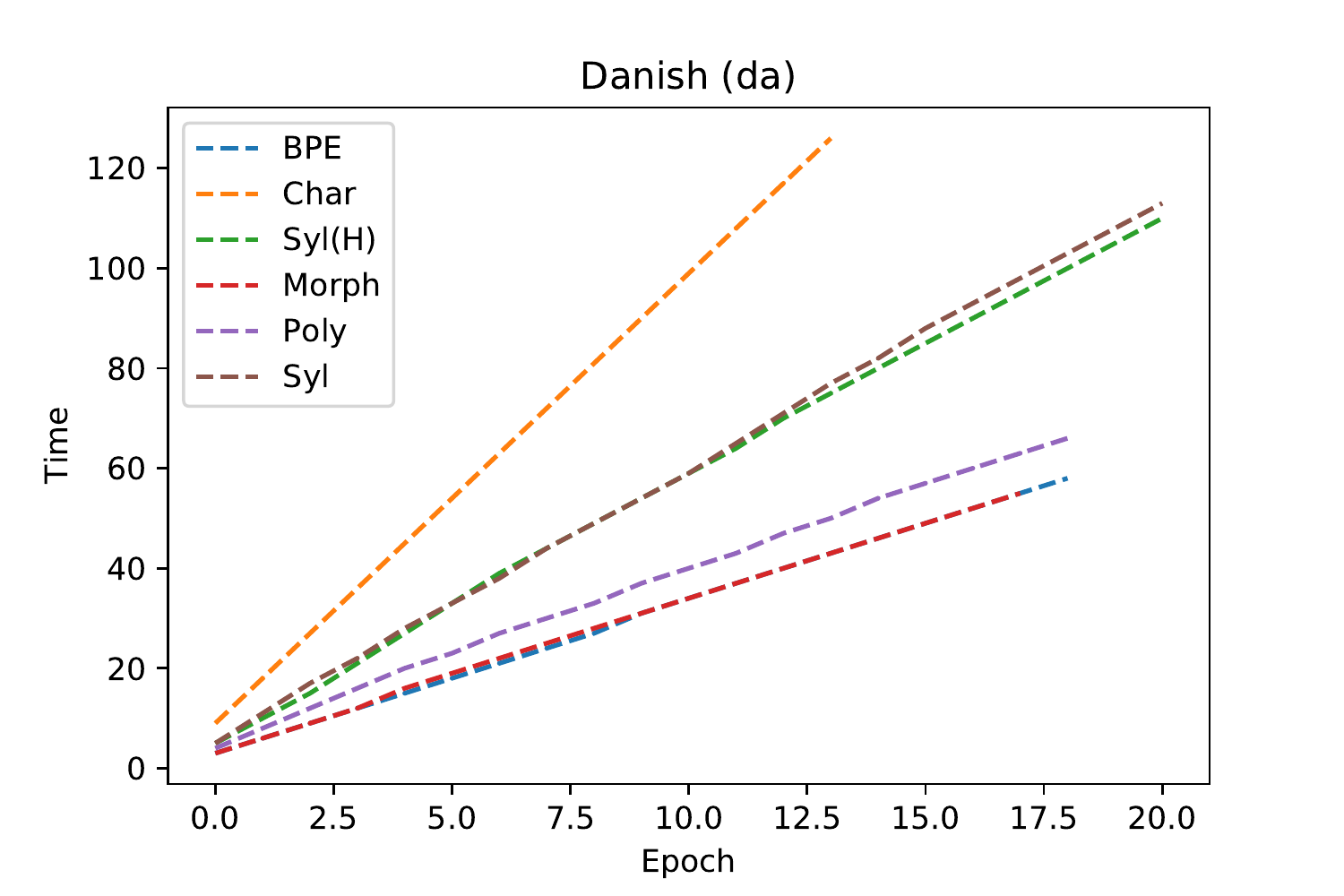}
\end{subfigure}

\begin{subfigure}[t]{0.24\linewidth}
\includegraphics[width=\linewidth,clip]{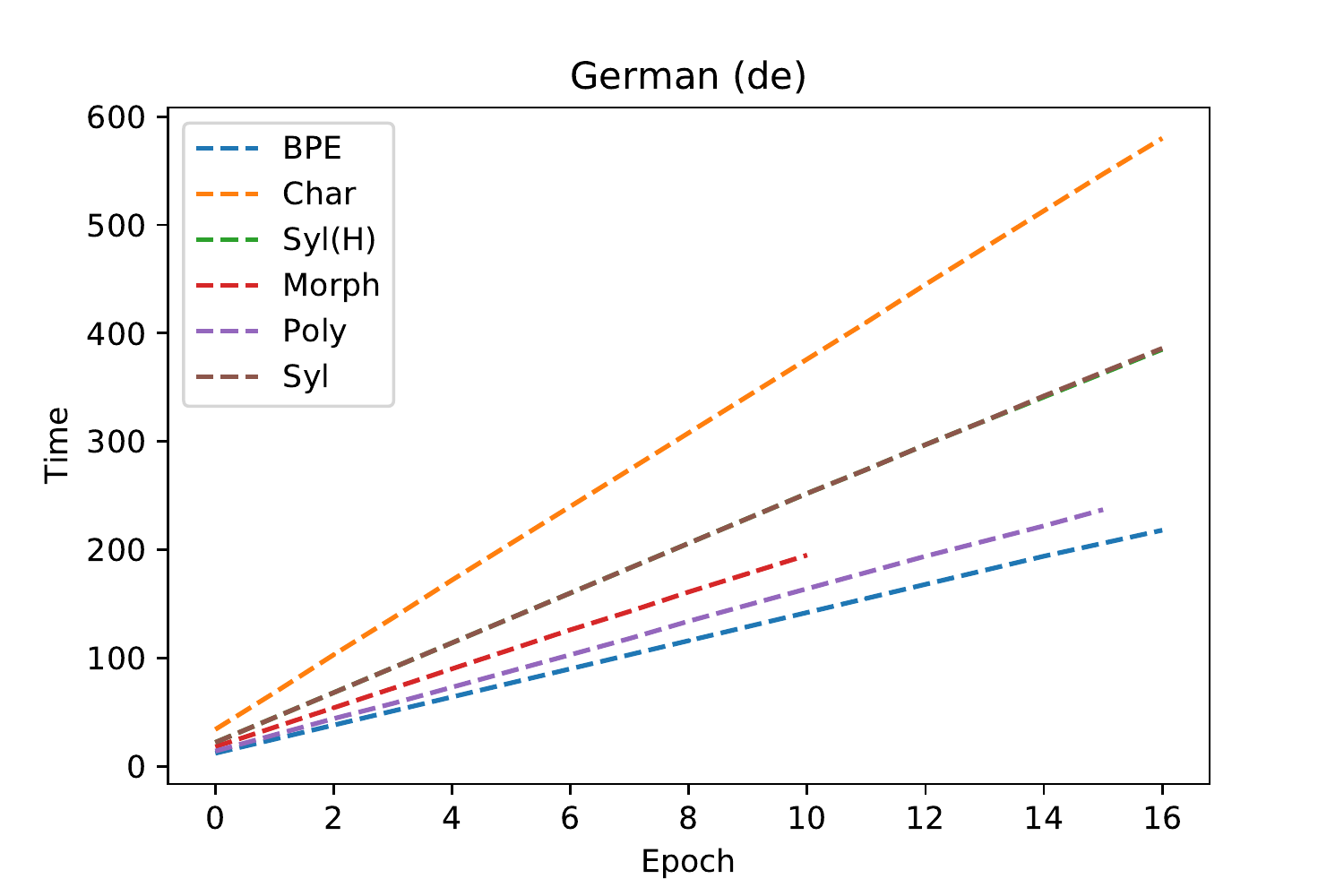}
\end{subfigure}
\begin{subfigure}[t]{0.24\linewidth}
\includegraphics[width=\linewidth,clip]{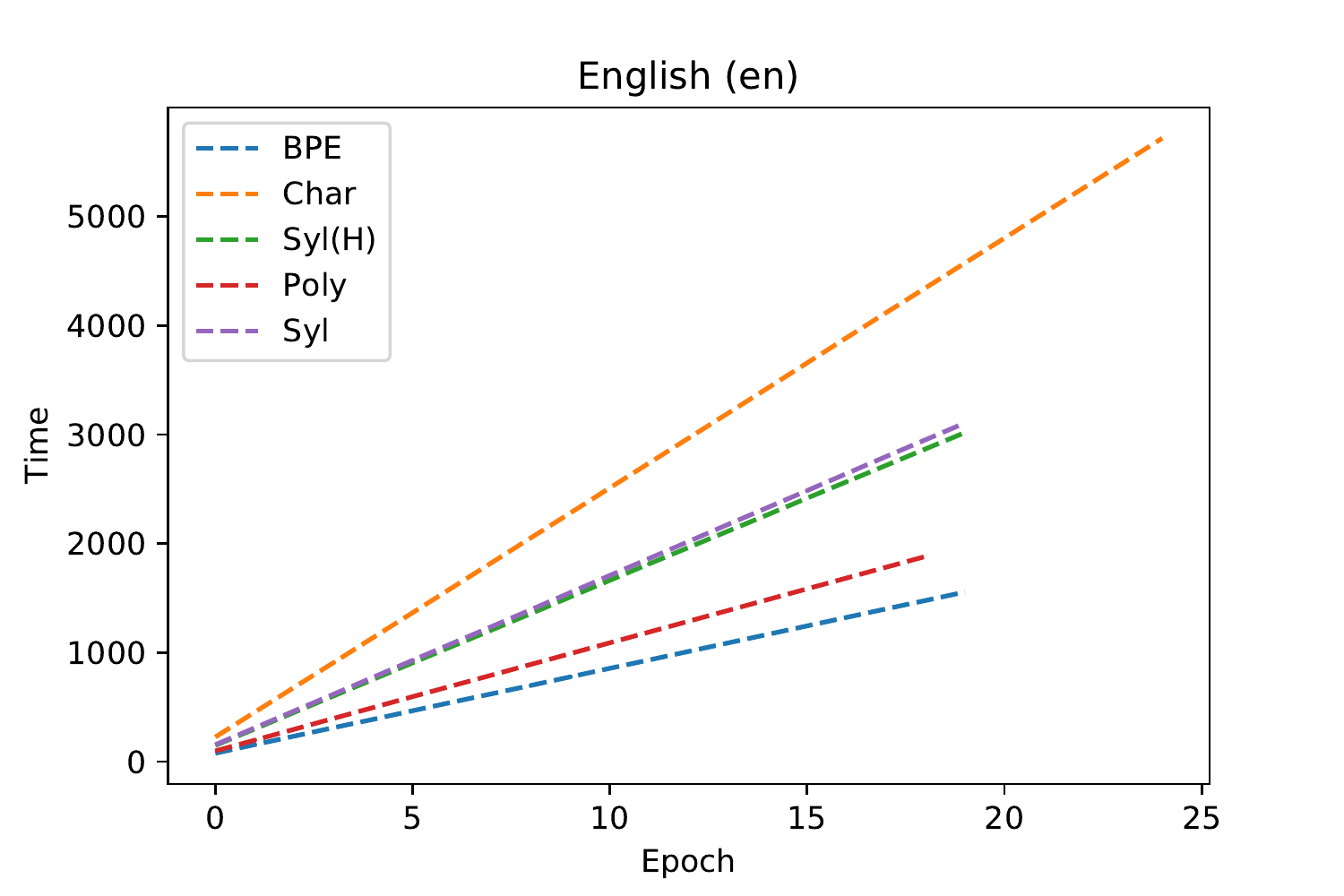}
\end{subfigure}
\begin{subfigure}[t]{0.24\linewidth}
\includegraphics[width=\linewidth,clip]{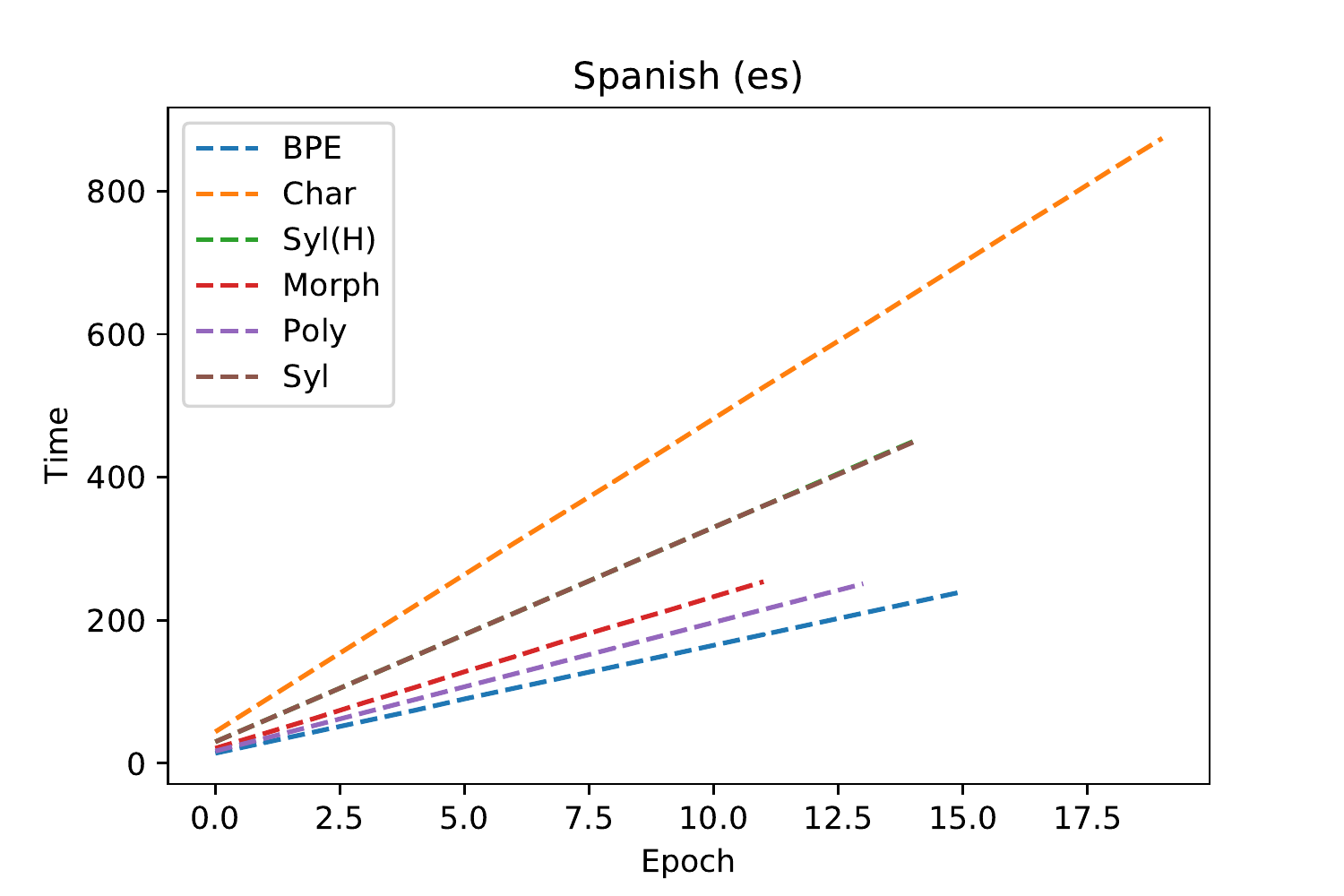}
\end{subfigure}
\begin{subfigure}[t]{0.24\linewidth}
\includegraphics[width=\linewidth,clip]{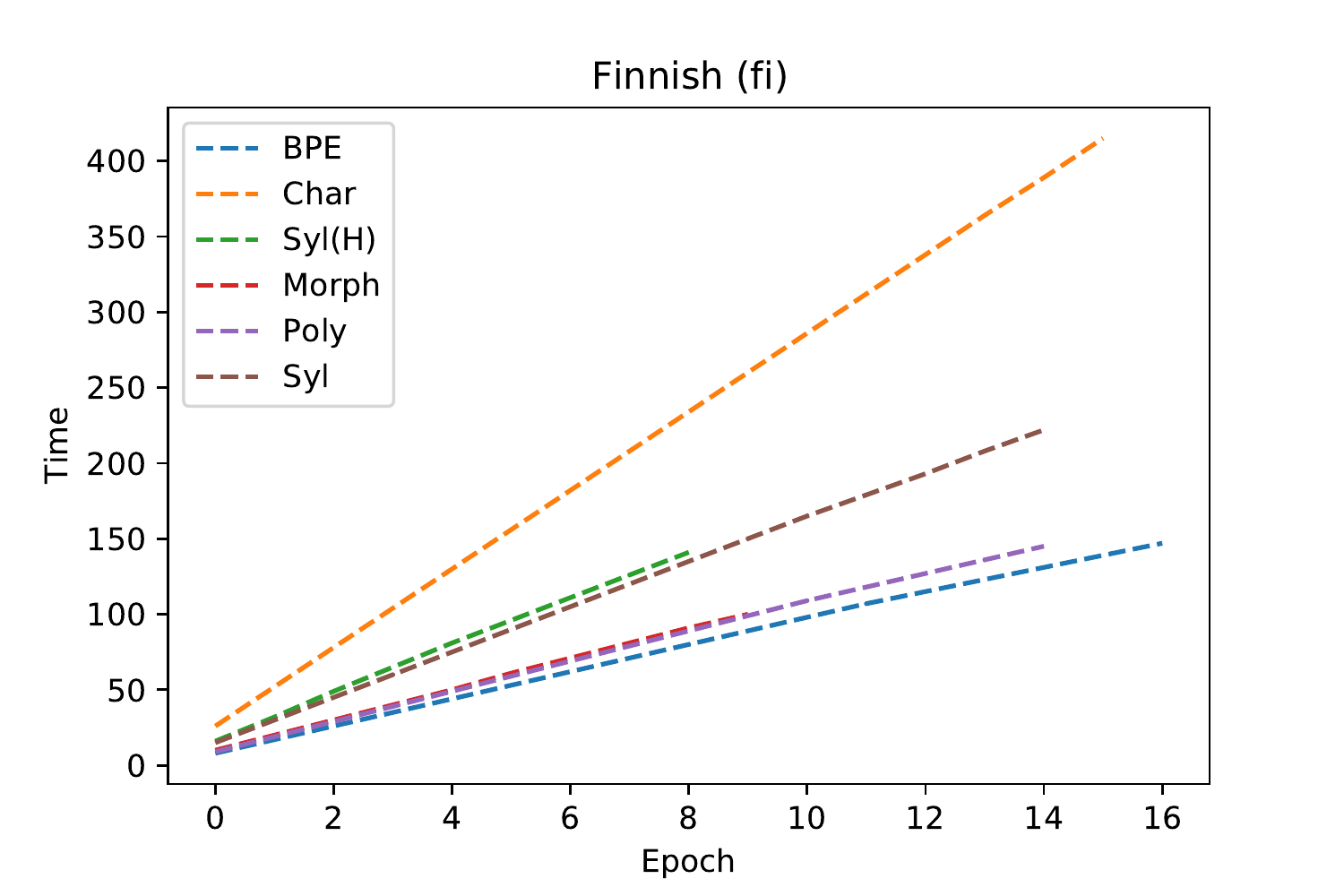}
\end{subfigure}

\begin{subfigure}[t]{0.24\linewidth}
\includegraphics[width=\linewidth,clip]{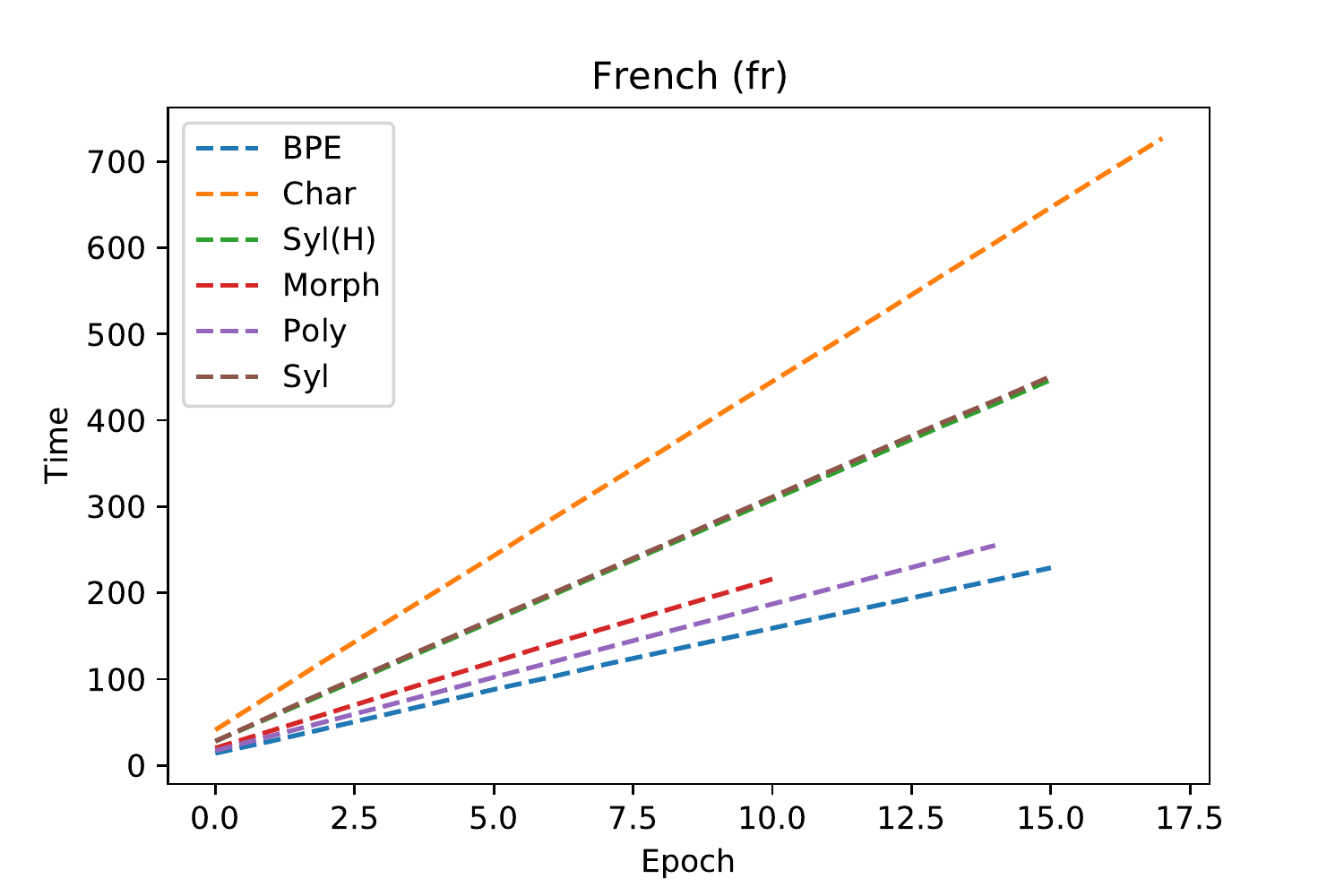}
\end{subfigure}
\begin{subfigure}[t]{0.24\linewidth}
\includegraphics[width=\linewidth,clip]{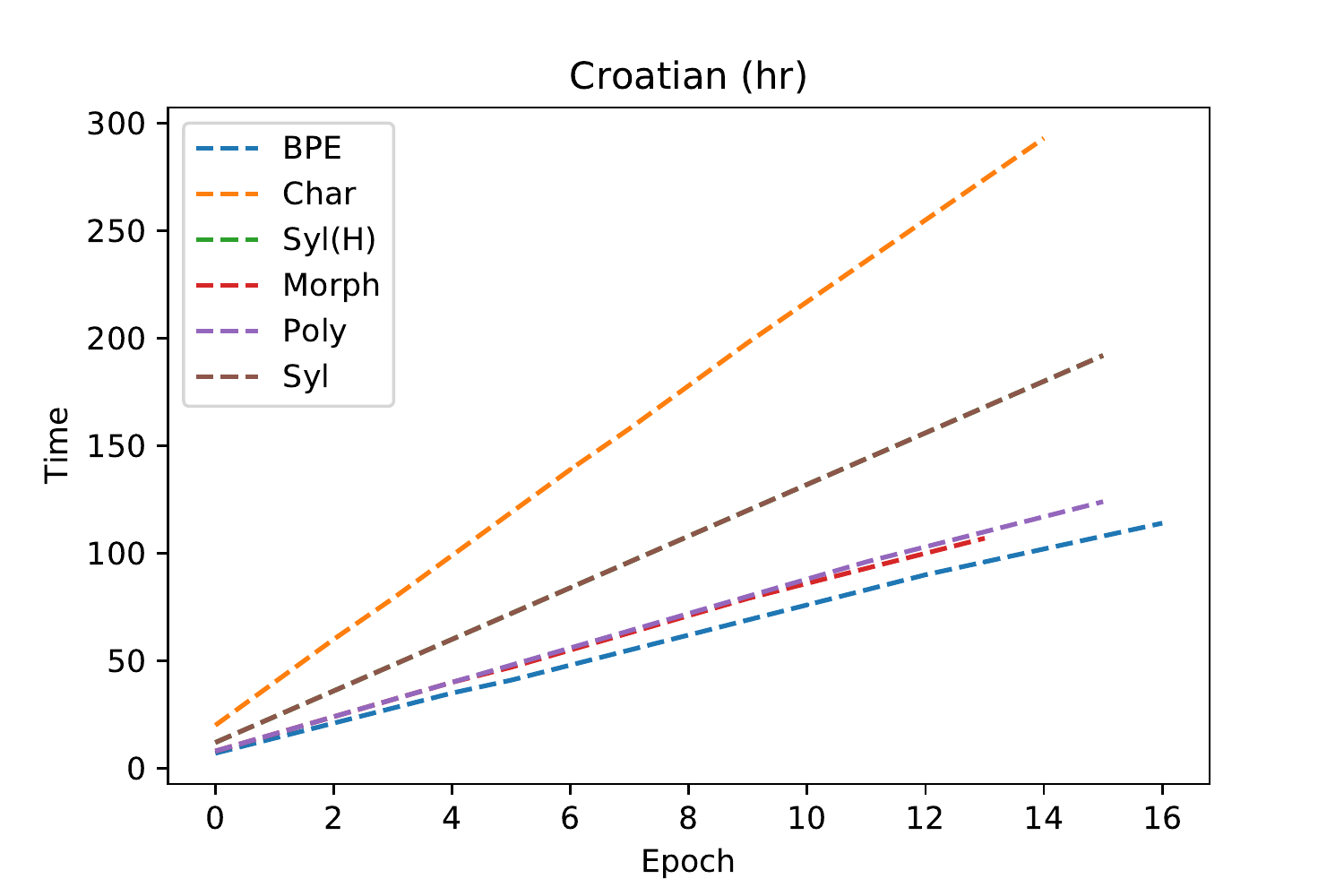}
\end{subfigure}
\begin{subfigure}[t]{0.24\linewidth}
\includegraphics[width=\linewidth,clip]{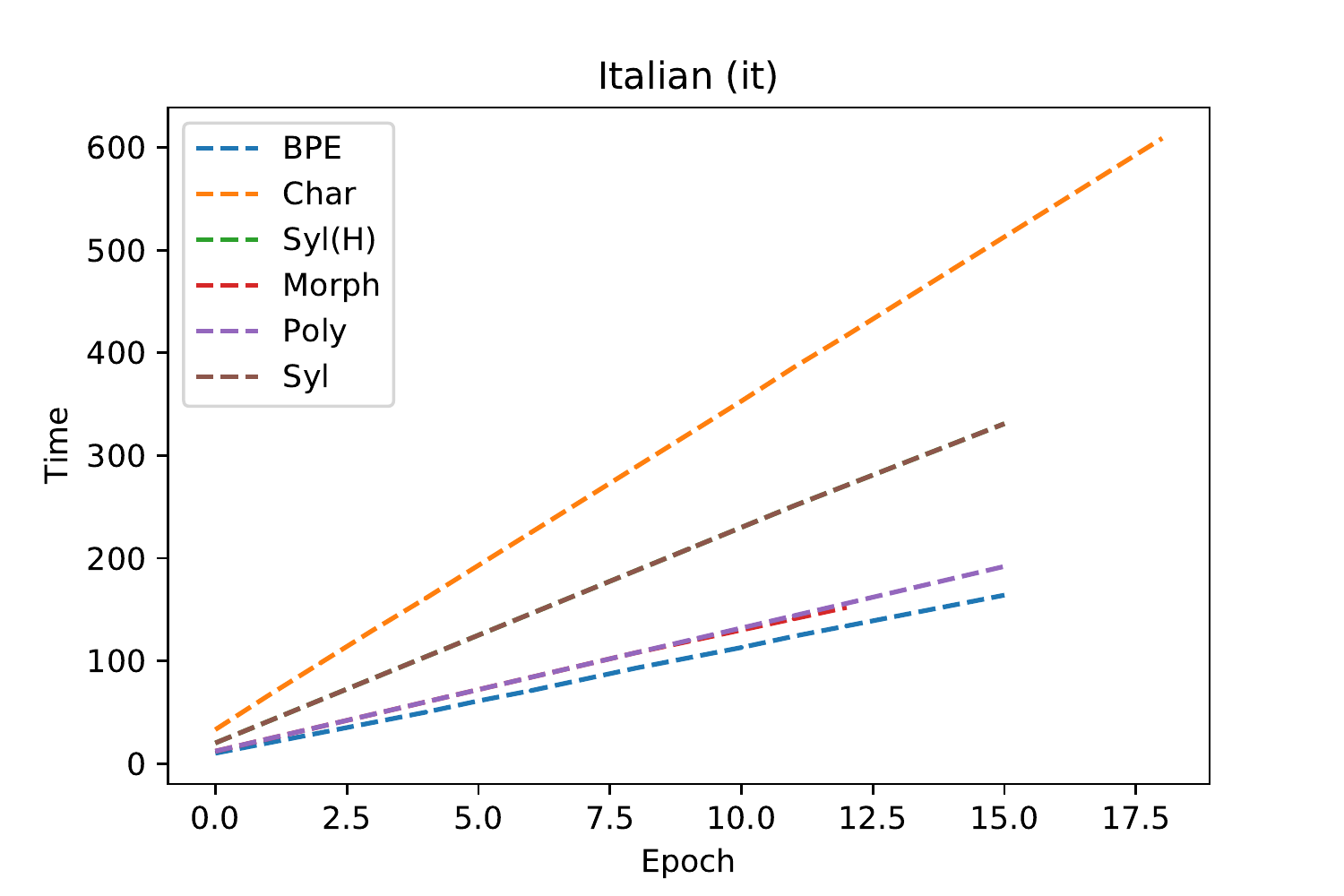}
\end{subfigure}
\begin{subfigure}[t]{0.24\linewidth}
\includegraphics[width=\linewidth,clip]{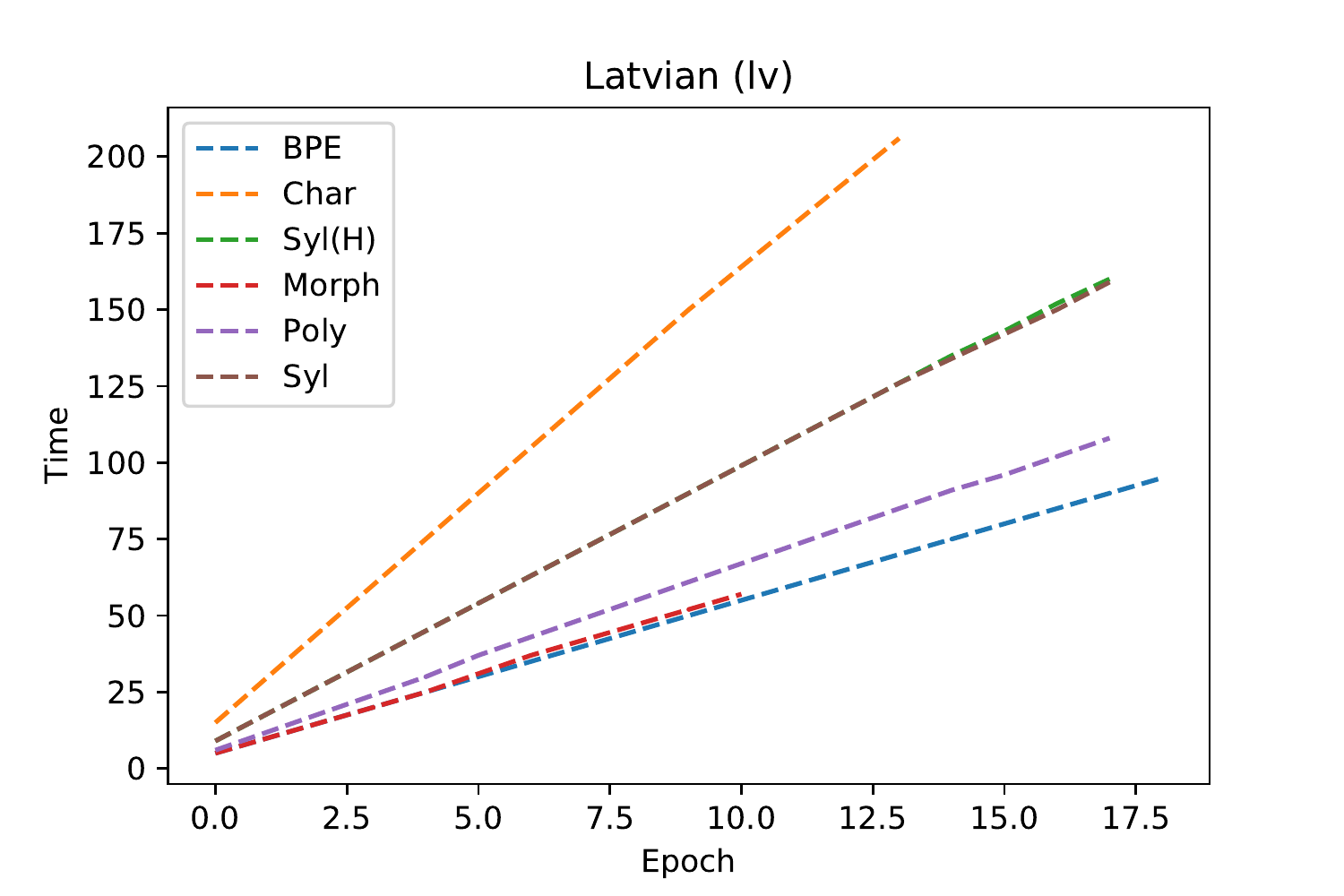}
\end{subfigure}

\begin{subfigure}[t]{0.24\linewidth}
\includegraphics[width=\linewidth,clip]{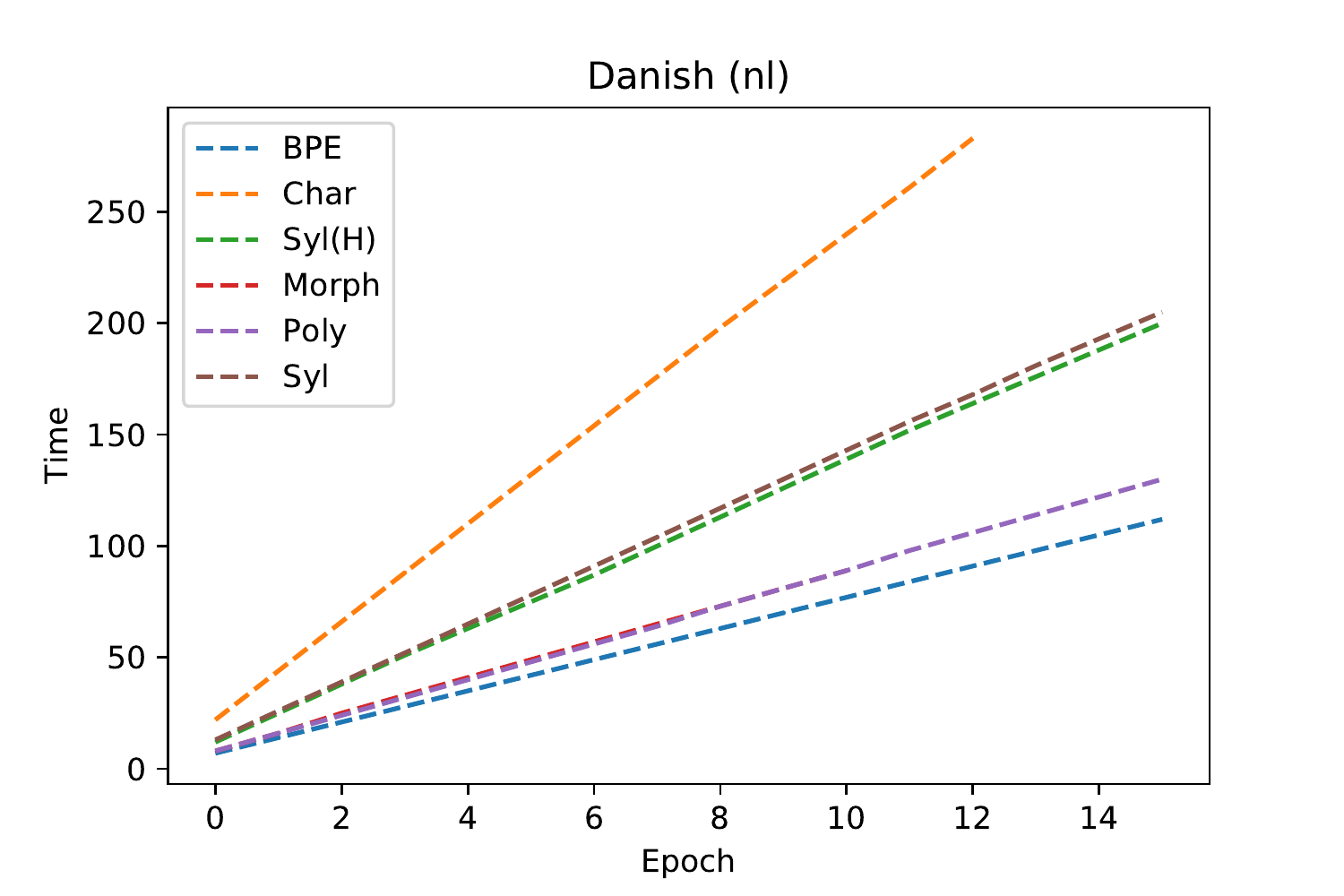}
\end{subfigure}
\begin{subfigure}[t]{0.24\linewidth}
\includegraphics[width=\linewidth,clip]{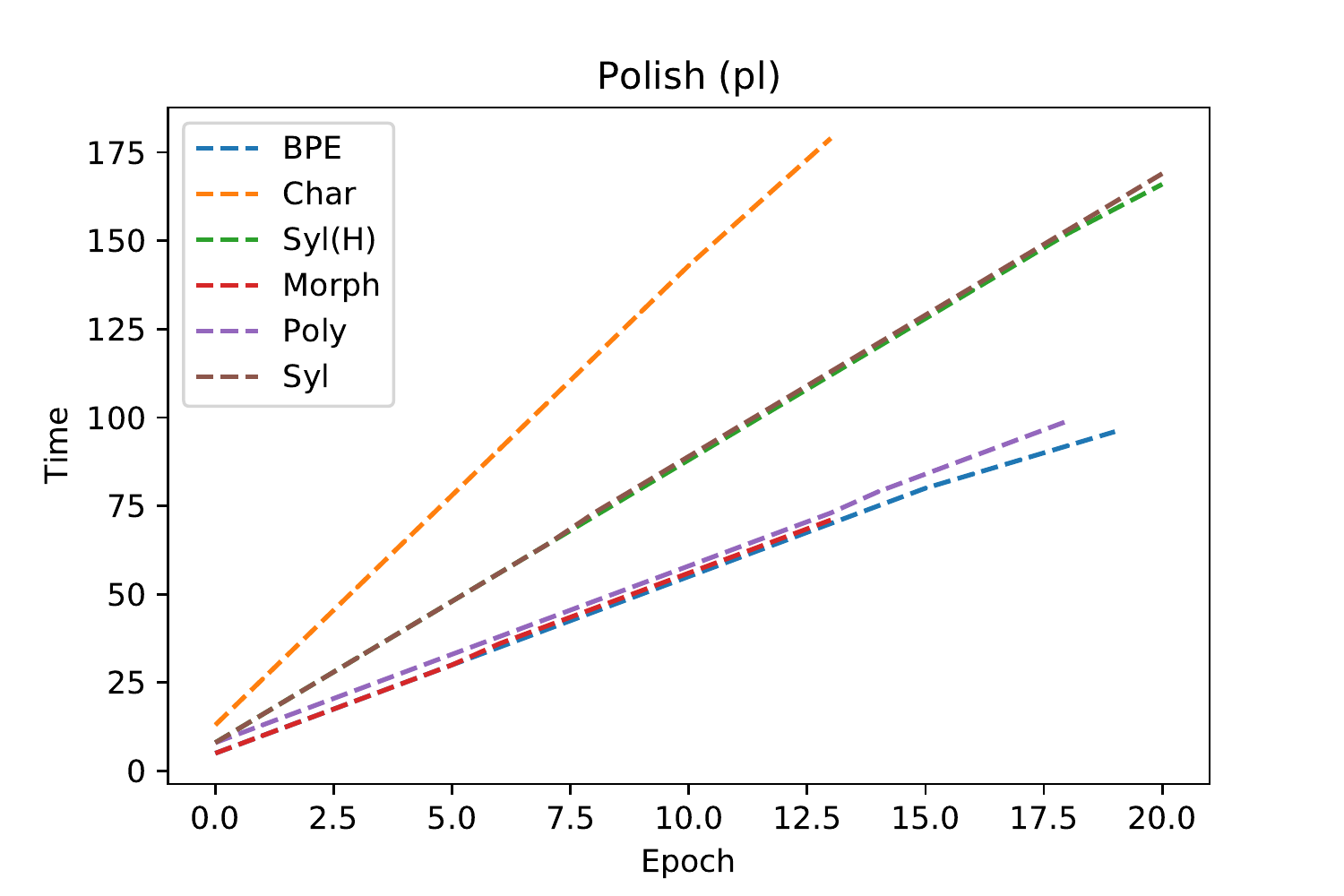}
\end{subfigure}
\begin{subfigure}[t]{0.24\linewidth}
\includegraphics[width=\linewidth,clip]{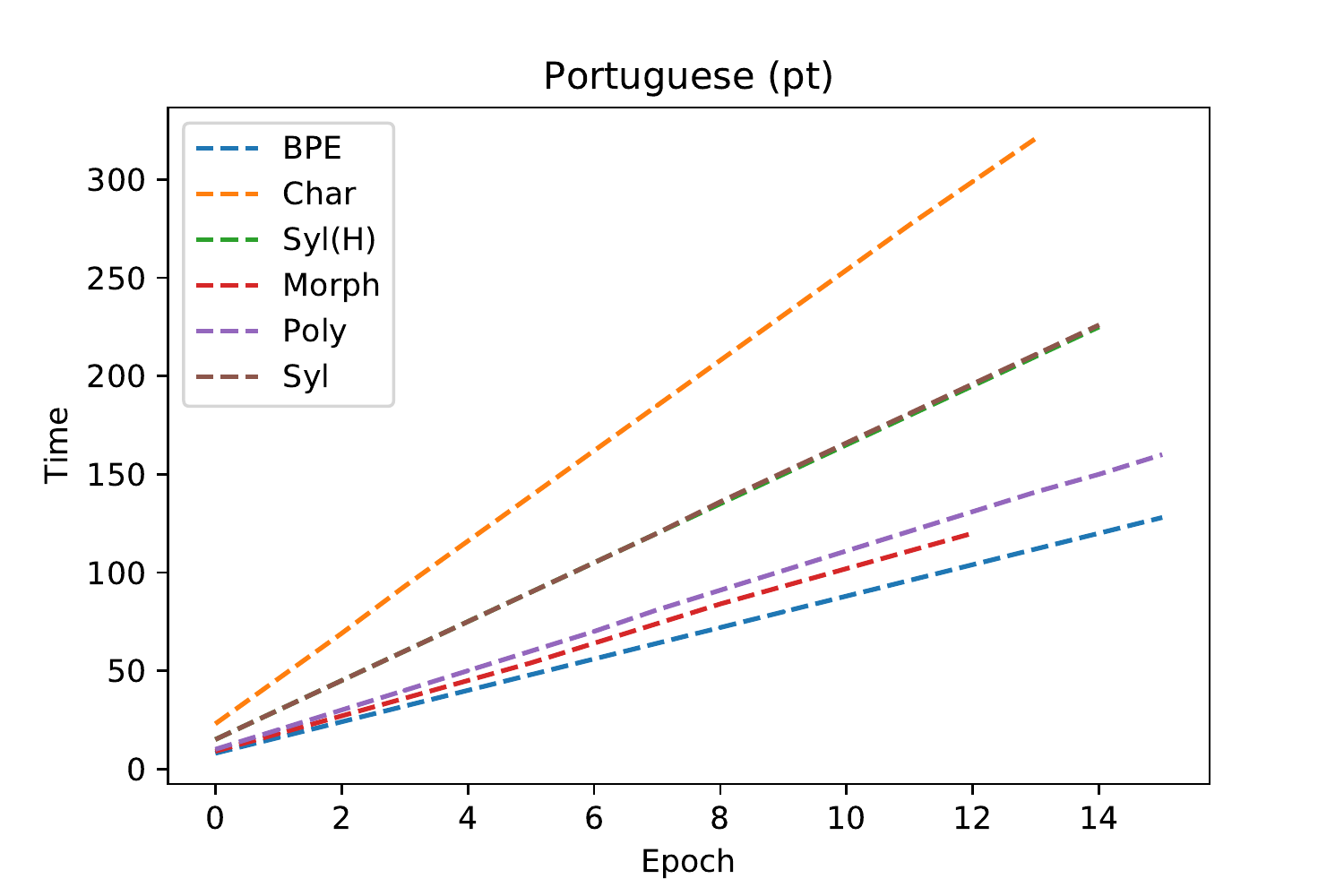}
\end{subfigure}
\begin{subfigure}[t]{0.24\linewidth}
\includegraphics[width=\linewidth,clip]{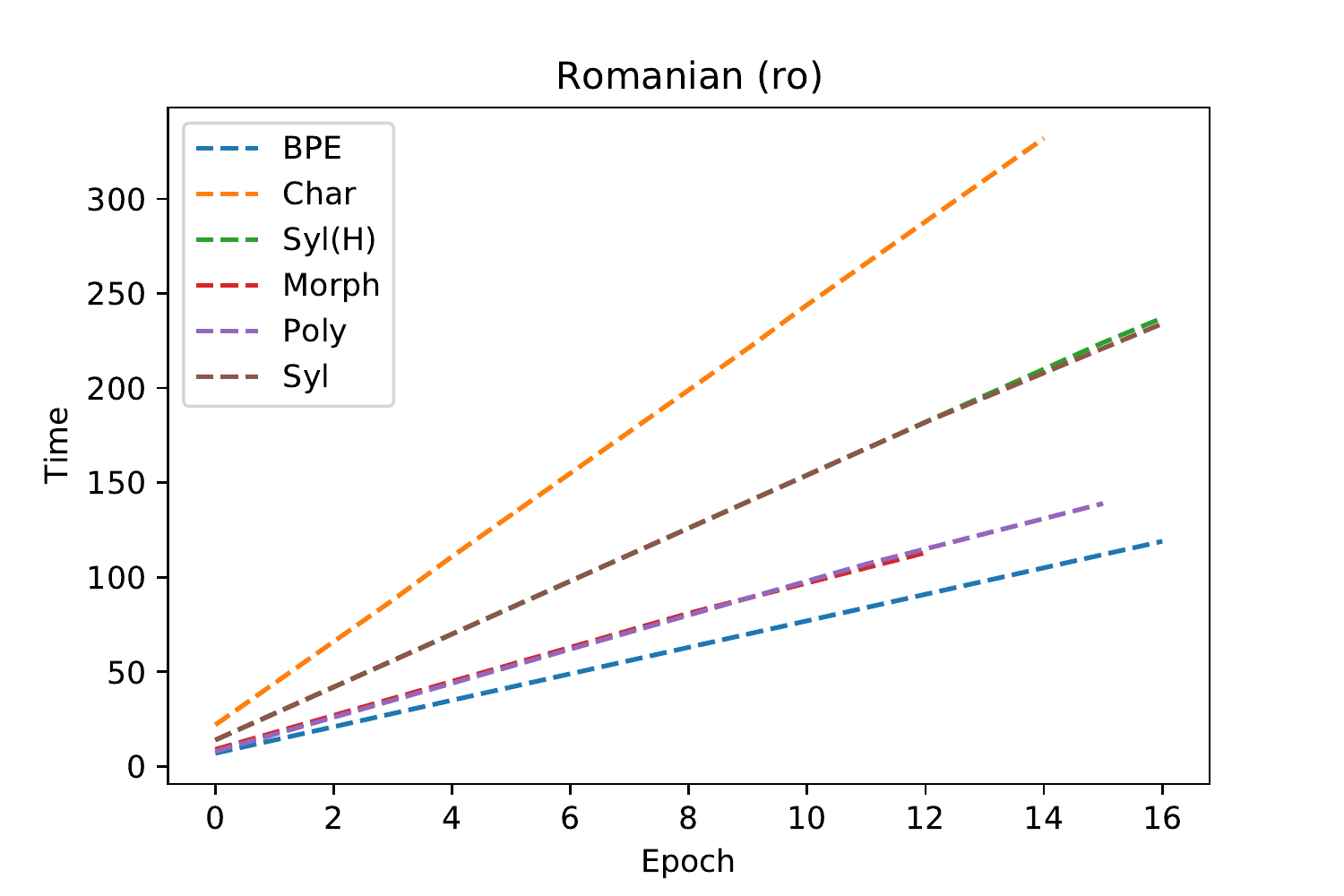}
\end{subfigure}

\begin{subfigure}[t]{0.24\linewidth}
\includegraphics[width=\linewidth,clip]{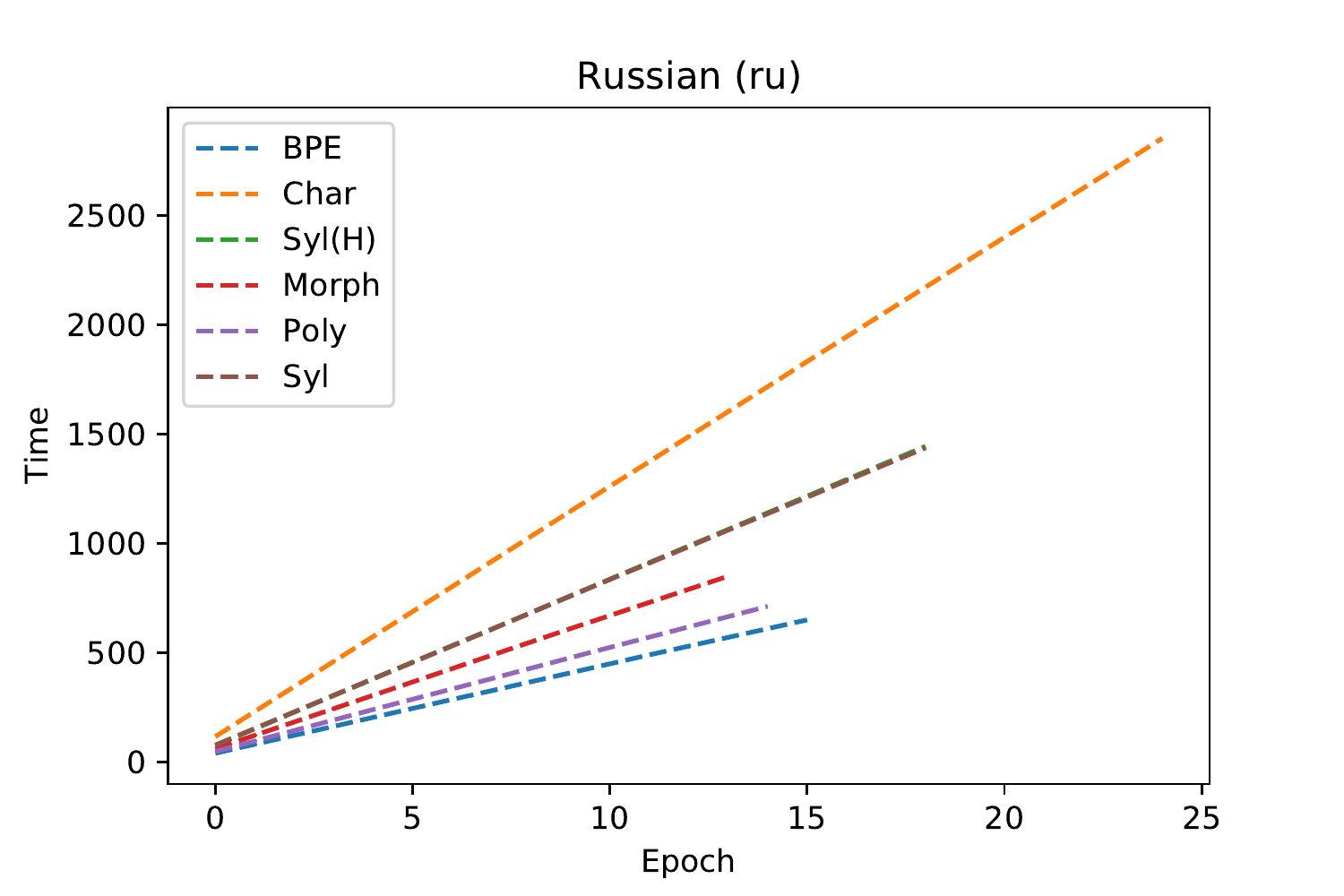}
\end{subfigure}
\begin{subfigure}[t]{0.24\linewidth}
\includegraphics[width=\linewidth,clip]{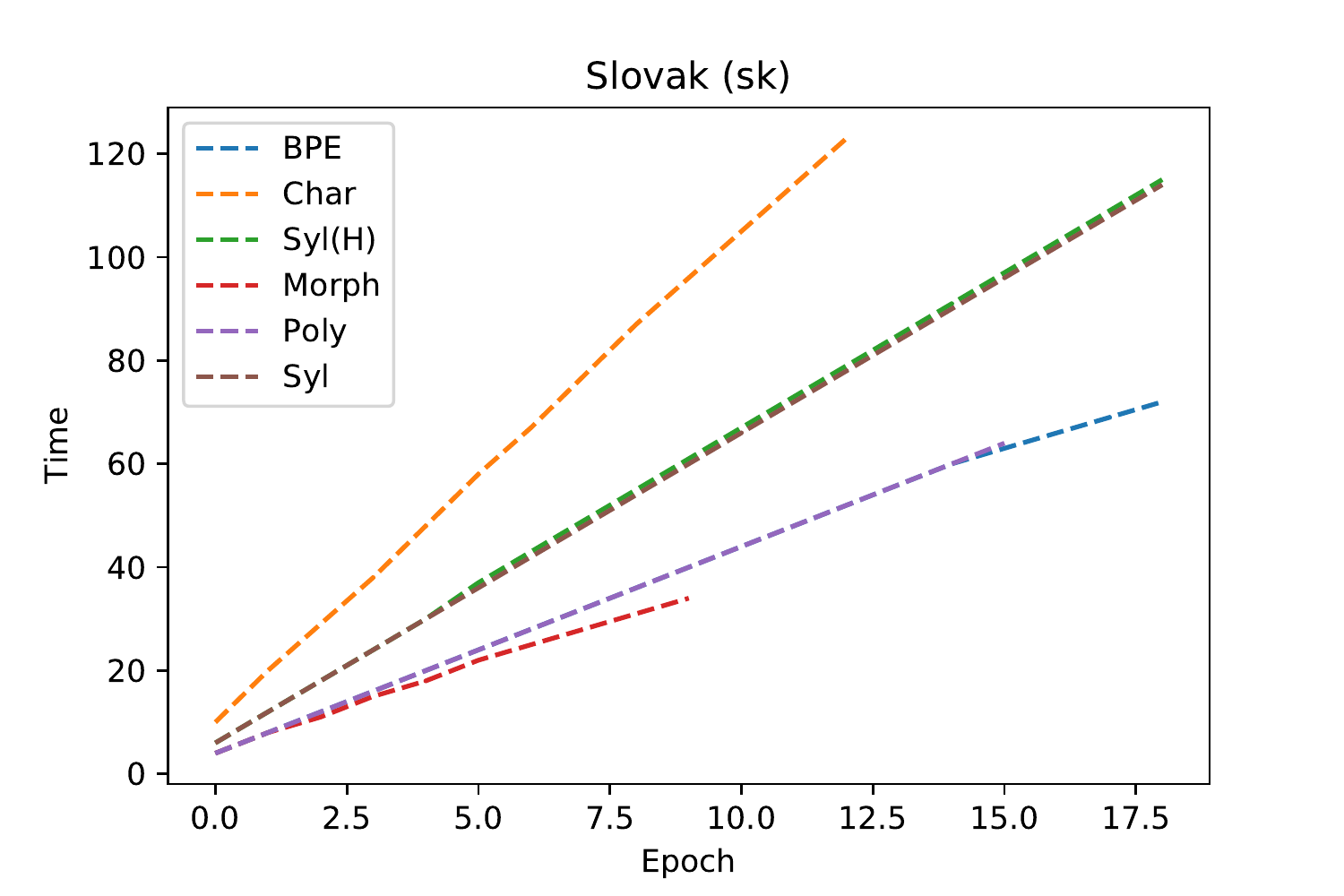}
\end{subfigure}
\begin{subfigure}[t]{0.24\linewidth}
\includegraphics[width=\linewidth,clip]{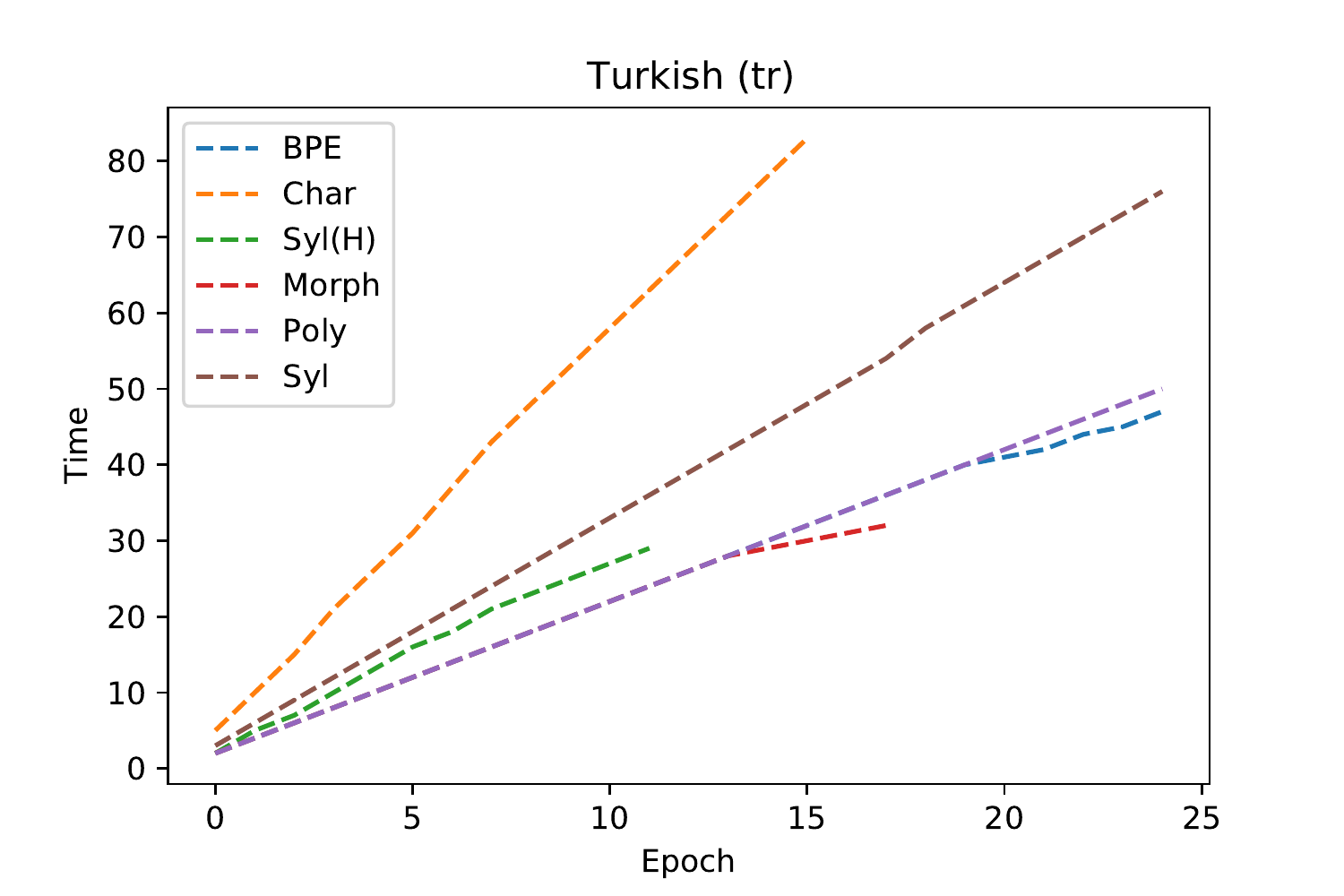}
\end{subfigure}
\begin{subfigure}[t]{0.24\linewidth}
\includegraphics[width=\linewidth,clip]{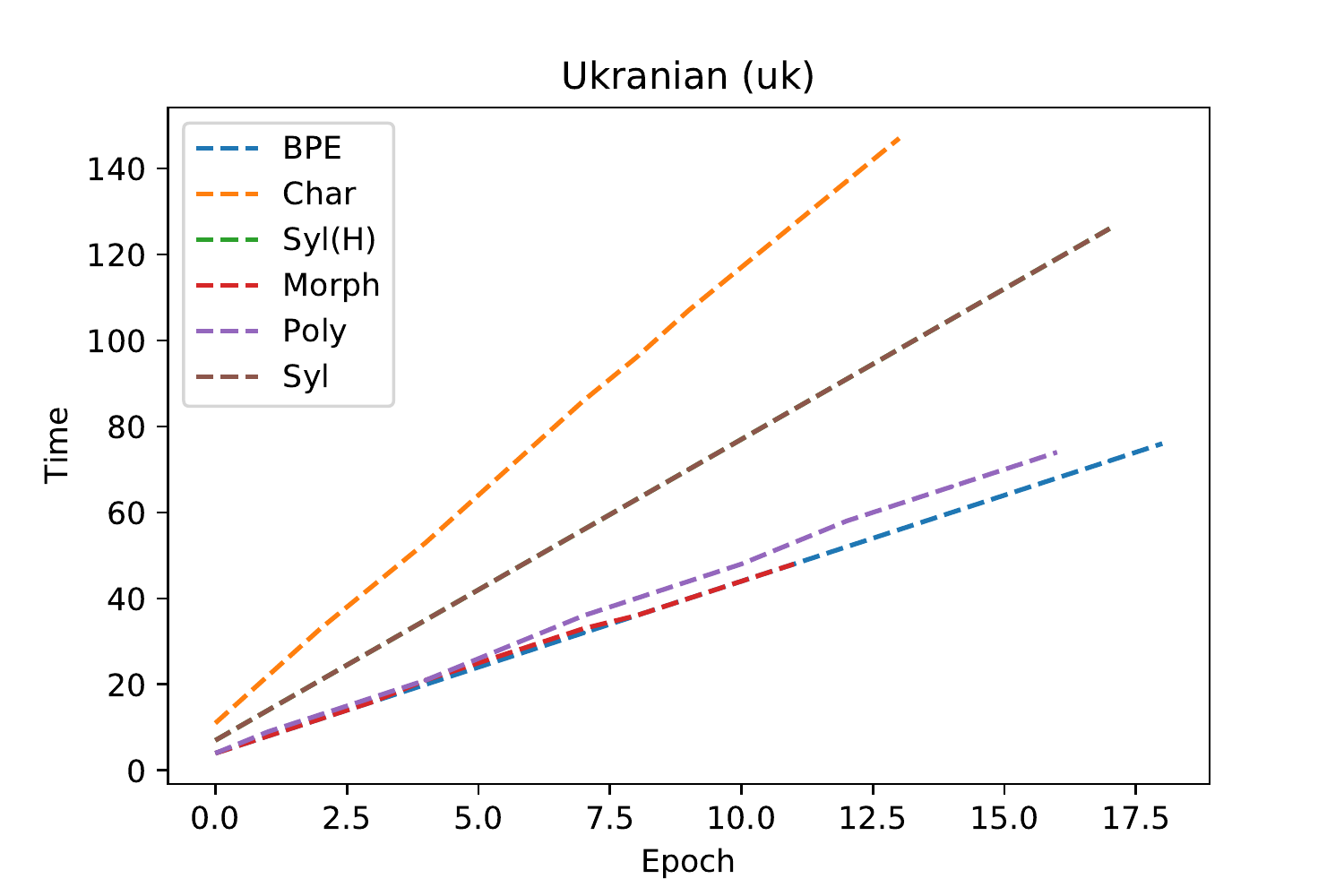}
\end{subfigure}

\caption{Training time (in seconds) until convergence for the UD treebanks} 
\label{fig:time-all}
\end{center}
\end{figure*}

Figures \ref{fig:convergence-all} and \ref{fig:time-all} shows the validation perplexity ($\operatorname{ppl}^c$) and the training time until convergence, respectively, in all the Universal Dependency treebanks. 

\begin{figure}[t!]
\begin{center}
\centering

\begin{subfigure}[t]{0.49\linewidth}
\includegraphics[width=\linewidth,clip]{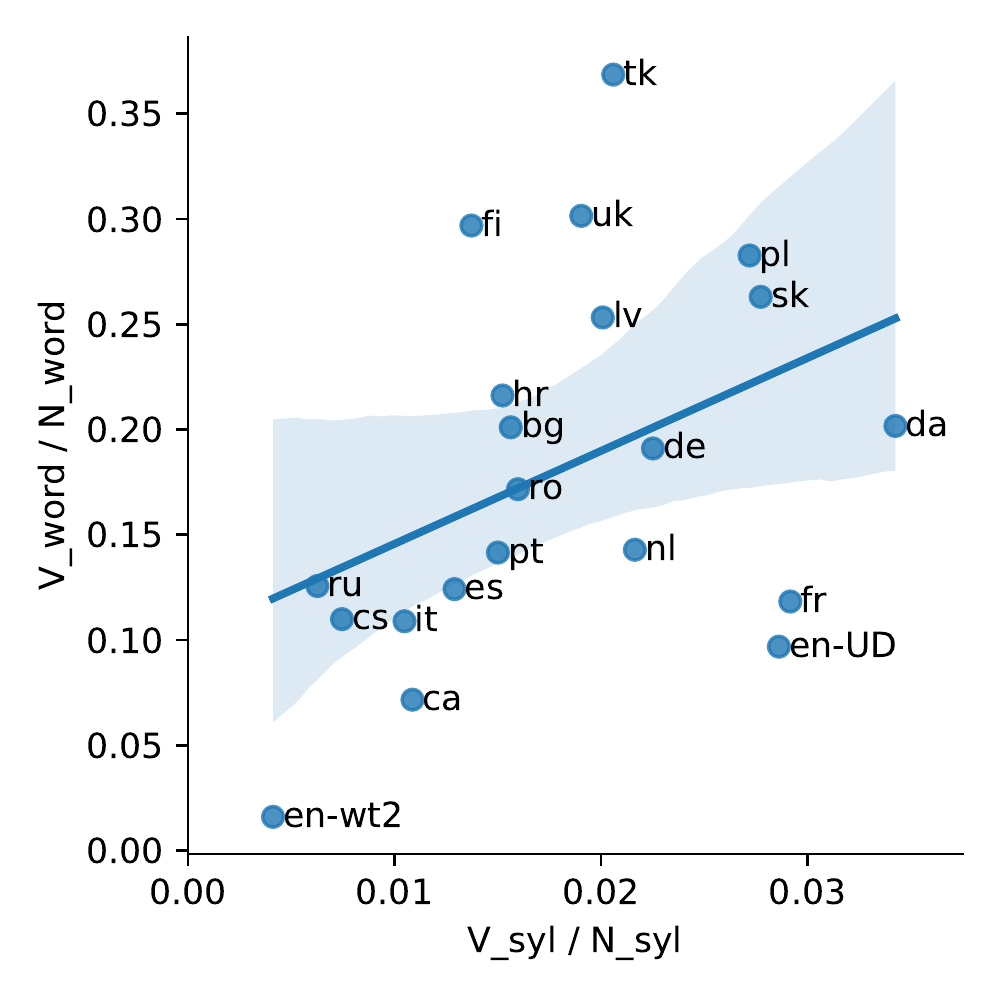}
\caption{V\textsubscript{syl}/N\textsubscript{syl} vs. V\textsubscript{word}/N\textsubscript{word}}
\label{fig:vocab-rate}
\end{subfigure}
\begin{subfigure}[t]{0.49\linewidth}
\includegraphics[width=\linewidth,clip]{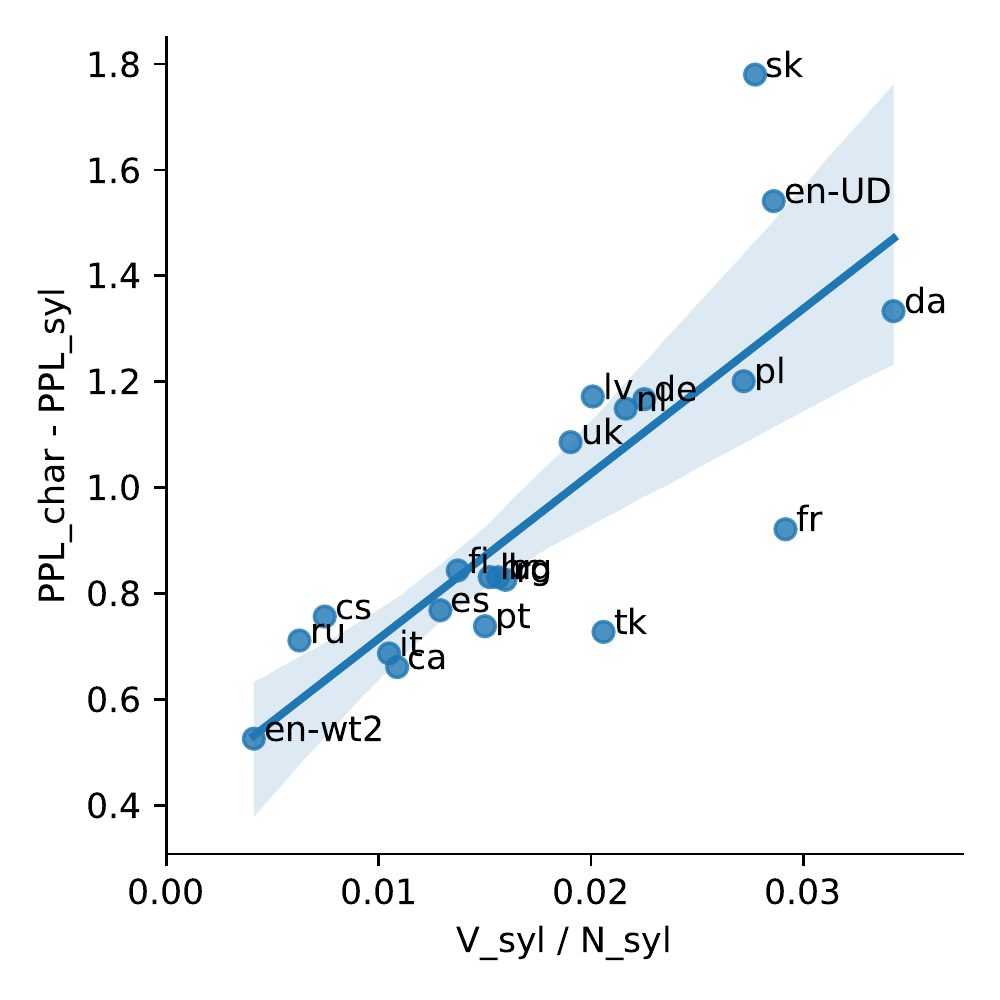}
\caption{V\textsubscript{syl}/N\textsubscript{syl} vs. $\Delta\operatorname{ppl}^c$\textsubscript{char-syl}}
\label{fig:ppl-diff}
\end{subfigure}

\caption{Left (a): Vocabulary growth rate of syllables (x-axis) versus words (y-axis). Right (b): Vocabulary growth rate of syllables (x-axis) versus the difference of $\operatorname{ppl}^c$ obtained by characters and syllables (y-axis).} 
\label{fig:regression-analysis}
\end{center}
\end{figure}

\section{Complementary discussion}
\label{app:discusssion}

\paragraph{Type/token ratio of syllables} 
In Figure \ref{fig:vocab-rate}, we show a scatter plot of the token/type growth rate of syllables versus words for all languages and corpora. In other words, the ratio of syllable-types (syllabary or V\textsubscript{syl}) per total number of syllable-tokens (N\textsubscript{syl}) versus the type/token ratio of words (V\textsubscript{word}/N\textsubscript{word}) in the train set. 
The figure suggests at least a weak relationship, which agrees with the notion that a low word-vocabulary richness only requires a low syllabary richness for expressivity. Also, a richer vocabulary can use a richer syllabary or just longer words, so the distribution of the vocabulary richness could be larger.

We expected that the syllabary growth rate (V\textsubscript{syl}/N\textsubscript{syl}) for a low phonemic language like English would be relatively high, but wikitext-2 (en-wt2) is located in the bottom-left corner of the plot, probably caused by its large amount of word-tokens. However, we observe a large V\textsubscript{syl}/N\textsubscript{syl} for the English (en-UD) and French (fr) treebanks, despite their low V\textsubscript{word}/N\textsubscript{word} ratio, which is an expected pattern for languages with deep orthographies. 

We also observe that languages with a more transparent orthography, like Czech (cs) or Finnish (fi), are located in the left side of the figure, whereas Turkish (tr) is around the middle section. 
Nevertheless, our study does not aim to analyse the relationship between the level of phonemic orthography with the V\textsubscript{syl}/N\textsubscript{syl} ratio. For that purpose, we might need an instrument to measure how deep or shallow a language orthography is \cite{borgwaldt-etal-2005-onset, borleffs-etal-2017-measuring}, and a multi-parallel corpus for a more fair comparison.

\end{document}